%% file: main.tex
\documentclass{article}

\usepackage[main, final]{neurips_2026}

\usepackage[utf8]{inputenc} 
\usepackage[T1]{fontenc}    
\usepackage{hyperref}       
\usepackage{url}            
\usepackage{booktabs}       
\usepackage{amsfonts}       
\usepackage{nicefrac}       
\usepackage{microtype}      
\usepackage{xcolor}         

\input{math_commands}

\usepackage[utf8]{inputenc} 
\usepackage[T1]{fontenc}    
\usepackage{hyperref}       
\usepackage{url}            
\usepackage{booktabs}       
\usepackage{amsfonts}       

\usepackage{amsmath}
\newcommand{\sci}[2]{\ensuremath{#1{\times}10^{#2}}}

\usepackage{amssymb}
\usepackage{nicefrac}       
\usepackage{microtype}      
\usepackage{xcolor}         
\usepackage{graphicx}
\usepackage{wrapfig}
\usepackage{caption}
\usepackage{subcaption}
\usepackage{adjustbox}
\usepackage{multirow}
\usepackage{xcolor}
\usepackage{pifont}
\usepackage{enumitem}
\usepackage{comment}
\usepackage{cleveref}
\usepackage[table]{xcolor}
\usepackage[dvipsnames]{xcolor}
\usepackage{amsthm}
\usepackage{algorithm}
\usepackage{algpseudocode}
\definecolor{darkpastelblue}{rgb}{0.47, 0.62, 0.8}
\hypersetup{
    colorlinks=true,          
    linkcolor=teal,     
    citecolor=Tan,    
    urlcolor=teal,      
    pdfborder={0 0 0}         
}

\definecolor{softgreen}{RGB}{180, 220, 180}   
\definecolor{softorange}{RGB}{255, 240, 150}   
\definecolor{softred}{RGB}{245, 180, 180}     

\usepackage{pifont}
\newcommand{\cmark}{\ding{51}}%
\newcommand{\xmark}{\ding{55}}%

\newcommand{\greencheck}{{\color{green}\checkmark}}
\newcommand{\redcross}{{\color{red}\ding{55}}}

\newcommand{\sn}[2]{#1e#2}                    
\newcommand{\bsn}[2]{\textbf{#1e#2}}          
\newcommand{\usn}[2]{\underline{#1e#2}}       

\usepackage{float}

\usepackage{stfloats}

\usepackage{enumitem}
\usepackage{array} 
\newcolumntype{C}[1]{>{\centering\arraybackslash}p{#1}}

\title{Efficient Generative Transformer Operators for Million-Point PDEs}

\author{
Armand Kassaï Koupaï\textsuperscript{1}\thanks{Equal contribution. Correspondence: \href{mailto:armand.kassai@isir.upmc.fr}{armand.kassai[at]isir.upmc.fr}, \href{mailto:lise.leboudec@isir.upmc.fr}{lise.leboudec[at]isir.upmc.fr}} \quad
Lise Le Boudec\textsuperscript{1}\footnotemark[1] \quad
Patrick Gallinari\textsuperscript{1,2} \\
\textsuperscript{1} Sorbonne Université, CNRS, ISIR, 75005 Paris, France \\
\textsuperscript{2} Criteo AI Lab, Paris, France
}

\begin{document}

\maketitle

\begin{abstract}

We introduce \textbf{ECHO}, a transformer-based neural operator for large-scale PDE modeling that learns to \emph{generate full trajectories in compressed spatio-temporal latent spaces}. 
Scaling neural operators to high-resolution systems remains challenging: dense grids are computationally prohibitive, and autoregressive solvers accumulate errors over long horizons. ECHO addresses both issues by jointly compressing space and time and replacing step-by-step prediction with trajectory-level generation.
ECHO combines a hierarchical encoder–decoder achieving up to 100× compression, a staged training strategy for high-resolution fidelity, and a latent generative process that models distributions over complete trajectories, improving long-term consistency.
This formulation enables a single model to handle forward prediction, backward prediction, and interpolation. Across large-scale 2D and 3D benchmarks, ECHO achieves state-of-the-art performance and scales to million-point simulations with complex dynamics.
\end{abstract}

\section{Introduction}

Neural networks have emerged as a promising alternative to classical numerical solvers for modeling physical dynamics and solving partial differential equations (PDEs) \citep{Long2018,Raissi2019,bezenac2019}. Neural operators (NOs) extend these approaches by learning mesh-independent mappings between function spaces, enabling generalization across domains, resolutions, and discretizations \citep{Lu2019,Li2020fno,li2023transformer,serrano2023}, and supporting parametric \citep{approximationpdes,koupai2024geps} and multi-physics problems \citep{mccabe2023multiple,herde2024poseidon}.
Despite this progress, scaling NOs to high-resolution, large-scale systems remains a fundamental challenge. Interactions over millions of points quickly exhaust computational resources, while autoregressive time-stepping accumulates errors over long horizons \citep{Lippe2023,pedersen2025thermalizer}. As a result, existing approaches either focus on simplified settings, such as regular grids \citep{holzschuh2025pdetransformer,rozet2025lostlatentspaceempirical}, or limit problem complexity when addressing irregular geometries and high-frequency dynamics. Latent-space methods alleviate some of these costs \citep{alkin2024upt, serrano2024aroma}, but still struggle to jointly compress space and time without degrading reconstruction quality, especially at scale \citep{enma,wen2025goat,DrivAerNet}.

\begin{figure}[!t]
    \centering
    \includegraphics[width=\linewidth]{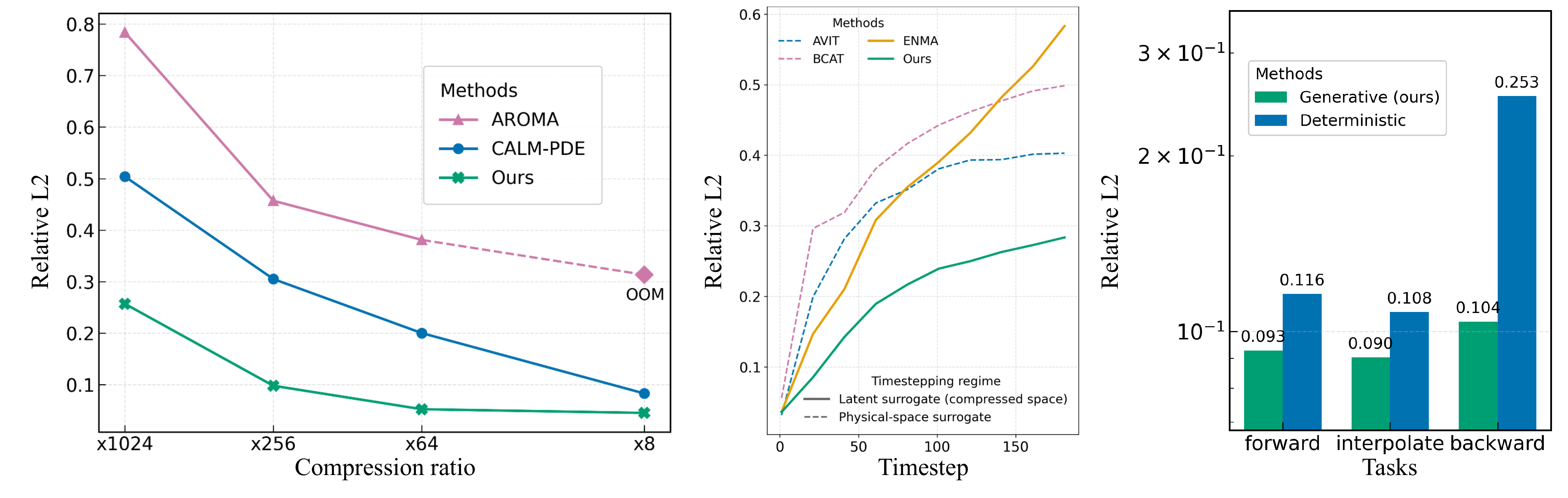}
    \caption{\textbf{Experimental analysis of ECHO.}
    Left: reconstruction error versus compression ratio on \emph{Vorticity}. Hierarchical, iterative compression improves accuracy, especially at high compression ratios.
    Middle: long-range rollouts on \emph{Gray-Scott}. Given the first $3$ frames, each model predicts the next $197$ frames; frames beyond $t=40$ lie outside the training horizon. Errors are averaged over test trajectories. Full-trajectory generation accumulates less error than latent autoregressive and physical-space baselines.
    Right: generative ECHO versus its deterministic variant on forward prediction, interpolation and backward prediction on \emph{Rayleigh-Bénard}.
    All panels report the relative $L^2$ error (lower is better).}
    \label{fig:experimental_analysis}
\end{figure}

This motivates a central question: \emph{What is the correct paradigm to scale surrogate models accurately and robustly to large-scale, arbitrary domains?}
We explore a formulation that couples strong spatio-temporal compression with trajectory-level generation, learning to represent dynamics in compact latent spaces while preserving relevant physical information.
This enables models to generate complete trajectory segments, enforcing global temporal coherence, reducing error accumulation, and improving scalability to large-scale systems.
We operationalize this idea through three complementary principles, illustrated in \cref{fig:experimental_analysis}.

\textbf{(i) Progressive hierarchical spatio-temporal compression.}
Most encoders downsample only spatial dimensions and perform compression in a single step, leading to poor scalability and information loss \citep{rigno,alkin2024upt,serrano2024aroma,wang2024LNO}. We introduce a hierarchical encoder–decoder that progressively compresses both space and time, yielding compact yet faithful spatio-temporal representations (\cref{fig:experimental_analysis} -- \textit{Left}).
\textbf{(ii) Trajectory-level generation.}
Autoregressive prediction remains the dominant paradigm \citep{mccabe2023multiple,hao2024dpot,wang2024cvit,liu2025bcat}, but suffers from error drift. We instead generate complete trajectory segments, enforcing horizon-wide consistency, mitigating error drift, and improving long-term stability (\cref{fig:experimental_analysis} -- \textit{Center}).
\textbf{(iii) Generative modeling in compressed latent space.}
We adopt a stochastic formulation \citep{zhou2025} that models distributions over full trajectories rather than deterministic predictions \citep{diffusionpde,serrano2024zebra,enma}, enabling robustness to partial observations and uncertainty (\cref{fig:experimental_analysis} -- \textit{Right}).

We instantiate these principles in \textbf{ECHO}, a transformer-based neural operator for large-scale PDE modeling. By combining hierarchical compression with trajectory-level generative modeling, ECHO enables efficient learning and inference on million-point systems defined on arbitrary domains, where most SOTA neural operators fail. Critical to this capability is the efficient spatio-temporal representation. Figure~\ref{fig:experimental_analysis} demonstrates the impact of each component: (left) significantly improved accuracy at high compression ratios as an enabler of large-scale applications, (center) a trajectory generation mechanism that reduces long-horizon error accumulation, and (right) a generative modeling paradigm that offers consistent gains over deterministic baselines. ECHO achieves state-of-the-art performance on large-scale 2D and 3D PDE benchmarks, scaling to simulations with up to 20 million points, where existing methods fail.

In addition, we introduce a staged training strategy that enables high-fidelity learning under memory constraints and a unified formulation that supports forward prediction, backward prediction, interpolation, and long-horizon forecasting within a single model. This enables us to address a variety of problems in a zero-shot setting.

\section{Problem Setting}
\label{sec:problem_setting}

We consider time-dependent partial differential equations (PDEs) defined on a spatial domain $\Omega \subset \mathbb{R}^d$ over a time interval $[0, T]$. Each instance is specified by an initial condition $\vu^0 \in L^2(\Omega, \mathbb{R}^{d_u})$ and a set of parameters $\gamma = (\vb, \vf, \vc)$, which include boundary conditions $\vb \in L^2(\partial \Omega \times [0, T], \mathbb{R}^{d_b})$, a forcing term $\vf \in L^2(\Omega \times [0, T], \mathbb{R}^{d_f})$, and PDE coefficients $\vc$. The governing system is
\begin{align}
    \mathcal{N}\!\left[ \vu; \vc, \vf \right](x,t) &= 0, && (x, t) \in \Omega \times (0, T], \\
    \mathcal{B}\!\left[ \vu; \vb \right](x,t) &= 0, && (x, t) \in \partial \Omega \times [0, T], \\
    \vu(x, 0) &= \vu^0(x), && x \in \Omega,
\end{align}
where $\mathcal{N}$ is a (possibly nonlinear) differential operator, $\mathcal{B}$ encodes the boundary conditions, and $\vu : [0, T] \times \Omega \to \mathbb{R}^{d_u}$ denotes the solution field.

In contrast to classical numerical solvers, fully data-driven neural solvers are not provided with an explicit PDE and must therefore be conditioned either on explicit PDE parameters $\gamma$ or on sequences of observed states.
For training, we assume access to a finite dataset $\mathcal{D}{\texttt{tr}}$ of $N$ trajectories, observed on a free-form spatial grid $\mathcal{X}{\texttt{tr}}$ at discrete times $\mathcal{T} \subset [0, T]$. Each trajectory consists of $|\mathcal{X}|$ mesh points, with $|\mathcal{X}|$ potentially very large in real-world problems. At test time, trajectories are observed on a spatial grid $\mathcal{X}{\texttt{te}}$, which may differ from $\mathcal{X}{\texttt{tr}}$.

To our knowledge, \textbf{ECHO} is the first generative operator to address forward prediction, backward prediction and interpolation under a unified formalism while operating in a compressed spatio-temporal latent space on arbitrary domains, enabling scalable modeling of high-resolution PDEs. Details of the inference setup, architecture, and training strategy are provided in the following sections.

\subsection{Inference Model}
\label{ssec:inference_model}

\begin{figure*}[htbp]
    \centering
    \includegraphics[width=0.9\linewidth]{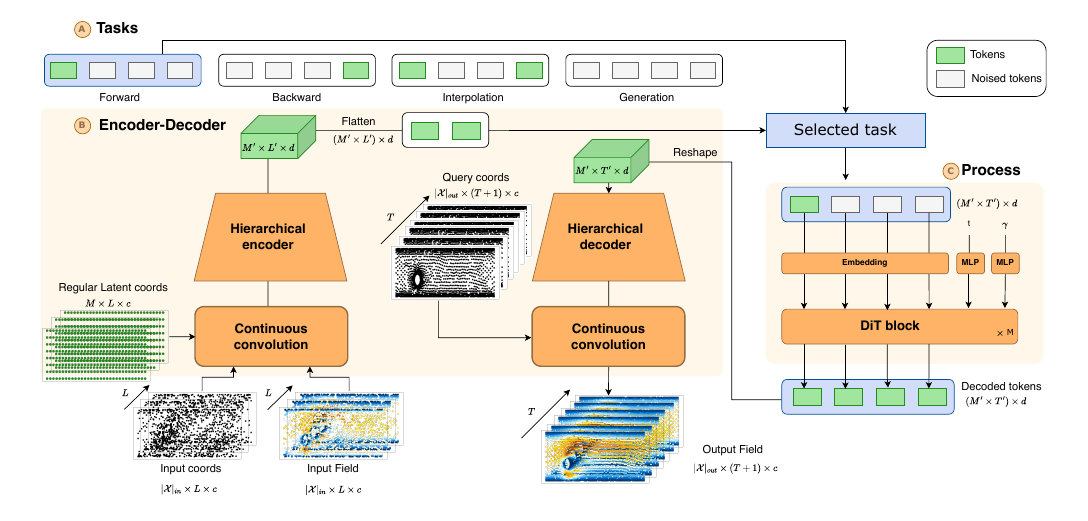}
    \caption{Architecture of the ECHO framework. ECHO comprises two components: (B) a convolutional auto-encoder and (C) a DiT-based generative process. The auto-encoder uses continuous convolutions to ingest irregular input grids of arbitrary size, maps the dynamics to a regular latent grid, and hierarchically compresses it; the decoder mirrors this hierarchy and applies a final continuous convolution, enabling queries at arbitrary output locations. The DiT module is trained with a flow-matching objective to denoise latent tokens, optionally conditioned on PDE parameters. This design allows ECHO to handle irregular grids and support multiple inference tasks (A).}
    \label{fig:archi-generale}
\end{figure*}

Since directly modeling PDE dynamics in physical space becomes computationally prohibitive at large scale, ECHO adopts an \emph{encode–generate–decode} paradigm (see \cref{fig:archi-generale}). This formulation allows us to handle partial observations on regular or irregular meshes at arbitrary resolutions.

\paragraph{Inference setting.}
At inference time, the objective is to reconstruct a full spatio-temporal trajectory from a limited set of observations. Let $\vu^{0:T} \in \mathbb{R}^{|\mathcal{X}| \times (T+1) \times c}$ denote the complete trajectory defined over $|\mathcal{X}|$ spatial locations and $T{+}1$ time steps with $c$ physical channels. We observe only $L$ states at times $\{t_0,\dots,t_{L-1}\} \subset [0,T]$ (see \cref{fig:archi-generale} - A).
We denote the observed and missing subsets as
\[
\mathcal{O} = \{\vu^{t_\ell}: t_\ell \in \{t_0, \dots, t_{L-1}\}\}, 
\quad
\mathcal{M} = \{\vu^t : t \notin \{t_0,\dots,t_{L-1}\}\},
\]
with $\mathcal{O} \cap \mathcal{M} = \emptyset$. Using a binary mask $m \in \{0,1\}^{T+1}$, we write
\[
\vu^{\mathcal{O}} = \vu^{0:T} \odot m,
\quad
\vu^{\mathcal{M}} = \vu^{0:T} \odot \bar{m},
\quad
\vu^{\mathcal{O}} + \vu^{\mathcal{M}} = \vu^{0:T}.
\]

\paragraph{Latent formulation.}
Rather than operating in physical space, inference is performed in a compressed spatio-temporal latent space. Observations are encoded by a hierarchical encoder $E_\phi$ into latent tokens $\vz^{\mathcal{O}'} = E_\phi(\vu^{\mathcal{O}}) \in \mathbb{R}^{M' \times L' \times d},$
defined on a regular latent grid. Here, $M'$ and $L'$ denote the number of spatial and temporal tokens, and $d$ is the latent dimension (see \cref{fig:archi-generale} - B). The temporal horizon is also compressed, with $T'$ corresponding to the latent counterpart of $T$.
The inference problem is thus recast as \emph{latent trajectory completion}: given observed tokens $\vz^{\mathcal{O}'}$, we aim to predict the missing tokens $\vz^{\mathcal{M}'}$. At the decoding stage, the latter will be mapped back to the physical space to get the reconstruction of $\vu^{\mathcal{M}}$.

\paragraph{Latent trajectory generation.}
The generative model $\mathcal{G}_\theta$ predicts the full latent trajectory $\vz^{0:T'}$ using a transformer trained with a flow-matching objective, made tractable thanks to the reduced number of tokens. During generation, missing tokens $\vz^{\mathcal{M}'}$ are initialized from Gaussian noise, while observed tokens $\vz^{\mathcal{O}'}$ and (optionally) PDE parameters $\bm{\gamma}$ provide conditioning (see \cref{fig:archi-generale} - C).

Trajectory generation is performed by solving the ODE
\begin{equation}
\vz^{0:T'} = \texttt{ODESOLVE}\!\left( \mathcal{G}_\theta(\vz_r^{\mathcal{M}'}, \vz^{\mathcal{O}'}, \bm{\gamma}, r) \right),
\quad
\vz^{\mathcal{M}'}_0 \sim \mathcal{N}(0, I),
\end{equation}
where $r \in [0,1]$ is the denoising index. We use the midpoint method as the ODE solver. 
Because the entire temporal sequence is resolved simultaneously, temporal drift is explicitly decoupled from the sequence length. 
The generation error is bounded by the flow-matching transport error and the autoencoder's information bottleneck, allowing ECHO to maintain stability over significantly longer physical horizons than iterative methods.

\paragraph{Decoding.}
The predicted latent trajectory $\hat{\vz}^{0:T'}$ is finally mapped back to the physical space using a decoder $D_\psi$, yielding the reconstructed trajectory $\hat{\vu}^{0:T}$.

\paragraph{Example: forward forecasting.}
ECHO supports multiple inference tasks by varying the observed set $\mathcal{O}$. For forward forecasting, $\mathcal{O}$ contains the first $L$ frames. These are encoded as $\vz^{0:L'-1} = E_\phi(\vu^{0:L-1})$, and the model predicts the remaining tokens $\{\vz^{L'}, \dots, \vz^{T'}\}$ to complete the trajectory.

The full pipeline can be summarized as:


\resizebox{1\linewidth}{!}{%
$\vu^{\mathcal{O}} \in \mathbb{R}^{|\mathcal{X}| \times L \times c}
\xrightarrow{\text{encode}}
\vz^{\mathcal{O'}} \in \mathbb{R}^{M' \times L' \times d}
\xrightarrow{\text{mask}}
\vz^{\mathcal{O'} \cup \mathcal{M'}} \in \mathbb{R}^{M' \times T' \times d}
\xrightarrow{\text{generate}}
\hat{\vz}^{0:T'} \in \mathbb{R}^{M' \times T' \times d}
\xrightarrow{\text{decode}}
\hat{\vu}^{0:T} \in \mathbb{R}^{|\mathcal{X}| \times (T+1) \times c}$%
}

\paragraph{Extensions.}
By varying $\mathcal{O}$, the same framework supports interpolation, backward prediction, and conditional or unconditional generation (see Appendix~\ref{app:Inference_instances}). For long-horizon prediction, we further apply segment-wise autoregressive generation. As in neural operator frameworks, outputs can be queried at arbitrary spatial locations, enabling tasks such as super-resolution and inpainting.

\section{Method}
\subsection{Architecture}
\label{ssec:architecture}
\label{sssec:encoder_decoder}

\textbf{Encoding.}
A core component of ECHO is a \emph{hierarchical spatio-temporal encoder} that progressively compresses inputs. High-quality reconstruction at strong compression ratios is difficult to achieve with a single-step projection to a small latent space. Instead, we adopt a sequential compression strategy, where information is gradually aggregated across space and time, allowing early layers to preserve fine-scale structure before further compression.

We first embed irregular spatio-temporal data $\vu^{\mathcal{O}}$ onto a regular latent grid, and then apply convolutional stages that reduce spatial and temporal resolution in a structured, multi-scale manner. This enables high compression while preserving fine-scale dynamics, which is critical for stable generation at scale. Encoding proceeds in two steps:

\begin{itemize}[leftmargin=*,align=parleft]

\item \textbf{Latent grid mapping.}
We map irregular spatio-temporal inputs onto a fixed regular grid $\bm{\Xi} = \{\xi_j\}_{j=1}^{S}$ using a continuous convolution~\citep{calmpde}. Let $f : \mathcal{X} \rightarrow \mathbb{R}^{C_{\mathrm{in}}}$ denote the input field defined at positions $\{x_p\}_{p=1}^N$, and $k$ a learned kernel. The output at latent query $\xi_j$ and channel $o$ is:
\begin{equation}
(f \ast k)_o(\xi_j) =
\sum_{i=1}^{C_{\mathrm{in}}} \sum_{p \in \mathrm{RF}(\xi_j)}
f_i(x_p)\, k_{i,o}(\xi_j - x_p).
\end{equation}
Here, $\mathrm{RF}(\xi_j)$ denotes the set of input points contributing to $\xi_j$, and the kernel $k_{i,o}$ is parameterized by an MLP over relative displacements $\xi_j - x_p$.  This continuous-convolution layer is the only operation that directly couples the irregular physical mesh to the latent grid.

A naive implementation evaluates all pairs in $\mathcal{X} \times \bm{\Xi}$, requiring $\mathcal{O}(NS)$ operations. To ensure scalability, we restrict each query $\xi_j$ to a local neighborhood $\mathcal{N}(\xi_j)$ of size $K \ll N$, yielding
\begin{equation}
(f \ast k)_o(\xi_j) =
\sum_{i=1}^{C_{\mathrm{in}}} \sum_{p \in \mathcal{N}(\xi_j)}
f_i(x_p)\, k_{i,o}(\xi_j - x_p).
\end{equation}
The mapping $j \mapsto \mathcal{N}(\xi_j)$ is computed using a spatial index, reducing complexity to approximately $\mathcal{O}(S \log N + SK)$, and is cached when meshes are reused.

We further reduce memory by factorizing the last layer of the kernel MLP: hidden features are first aggregated over the $K$ neighbors, followed by channel mixing. This avoids materializing tensors of size $S \times K \times C_{\mathrm{in}} \times C_{\mathrm{out}}$, reducing memory by a factor $\frac{K C_{\mathrm{out}}}{H+1}$, per latent query where $H$ is the hidden width.


\item \textbf{Spatio-temporal compression.}
Once the irregular field is mapped to $\vu^{\mathcal{O}}(\bm{\Xi})$, the remainder of the encoder operates on a regular grid. We can therefore use standard spatial and temporal convolutions. In this way, ECHO incurs the irregular-mesh cost only at the input and output layers. We stack \texttt{compress} blocks, which reduce spatial or temporal resolution, and \texttt{residual} blocks, which preserve resolution while enriching the representation. All temporal convolutions are causal, so a state at time $j$ depends only on states at times $i \leq j$. This hierarchical design progressively reduces the number of spatio-temporal tokens before the transformer stage. As a result, the generative model operates on compact latent trajectories rather than on the original physical mesh, which is essential for efficient large-scale trajectory generation.

\end{itemize}

\paragraph{Decoding.}
The decoder mirrors the encoder. Transposed convolutions reverse the spatio-temporal compression, and a final continuous convolution interpolates latent tokens to arbitrary physical query locations. 
This design enables querying at arbitrary spatial locations and supports tasks such as super-resolution and inpainting (see \cref{app:Inference_instances}).

\label{sssec:process}

\paragraph{Transformer Process} The ECHO process component is a transformer architecture trained with a flow-matching loss. It builds on the MM-DiT block  \citep{esser2024}. Given observed latent tokens $\vz^{\mathcal{O'}}$, unobserved $\bm{z}^{\mathcal{M'}}$ are initialized with a noise distribution and the model is trained to reconstruct the full trajectory $\vz^{0:T'}$ in the compact latent space. Thanks to the high compression ratio, global attention remains tractable—unlike recent PDE transformers that resort to neighborhood attention on raw fields \citep{holzschuh2025pdetransformer}.
A key feature of ECHO is its latent conditioning strategy, which is designed to support generative modeling of PDEs defined on arbitrary geometries. Instead of conditioning via adaptive normalization alone, as usually done, we concatenate the observed regularized latent tokens with noise:
\[
\texttt{concat}(\vz^{\mathcal{O'}}, \vz_{\texttt{noise}}^{\mathcal{M'}}), 
\quad \text{where} \quad \mathcal{O'} \cup \mathcal{M'} = [0, T'].
\]
This provides an in-context mechanism where observed frames guide noised ones. Denoising index $r$ is injected explicitly; it is embedded and added to normalization layers via AdaLN. The model thus processes the entire spatio-temporal trajectory as a unified structure. When PDE parameters $\bm{\gamma}$ are available, ECHO conditions on both $\bm{\gamma}$ and the observed tokens $\vz^{\mathcal{O'}}$. Parameters are encoded by a lightweight MLP and injected into each transformer block via AdaLN \citep{zhou2025,Li2025DiT, zhou2025unisolver}.

\subsection{Training}
\label{ssec:training}

\begin{figure*}[!htbp]
    \centering
    \includegraphics[width=0.9\linewidth]{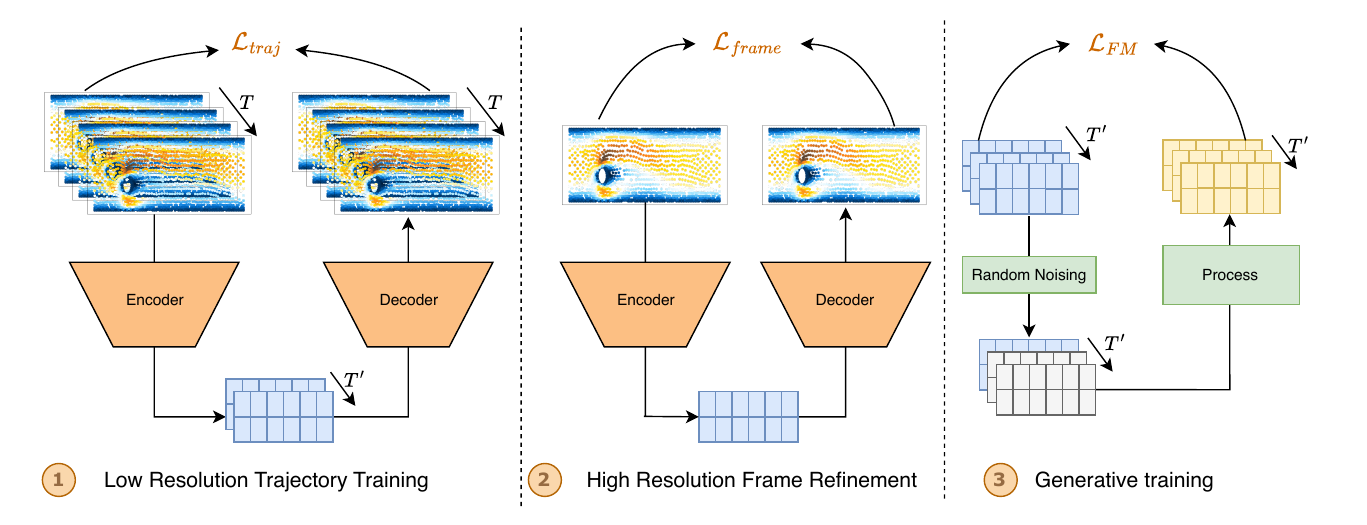}
    \caption{Three-stage training strategy for million-point trajectory generation. ECHO's auto-encoder is first trained for low-resolution trajectory reconstruction, then refined on high-resolution single frames. Both stages use reconstruction objectives. The generative processor is then trained separately in the frozen latent space using flow matching.}
    \label{fig:training}
\end{figure*}

Training generative models on full high-resolution trajectories is computationally and memory intensive, especially on irregular meshes, and existing methods often fail to scale to large problems. ECHO addresses this with a three-stage strategy that reduces memory requirements while preserving high-fidelity reconstruction. First, the auto-encoder learns spatio-temporal compression on low-resolution trajectory segments. Second, it is refined for spatial compression on high-resolution individual frames. Finally, the transformer processor is trained in the frozen latent space with a flow-matching objective. 

\paragraph{Stage 1: Spatio-temporal compression on subsampled trajectories.}
Training on dense, long trajectories is prohibitive. We therefore split trajectories into sub-trajectories using a sliding window and spatially subsample each frame by a factor between $0.2$ and $0.5$. This avoids out-of-memory errors while exposing the model to temporal contexts of $10$--$50$ frames. The auto-encoder is trained with a relative $L^2$ reconstruction loss. Let $B$ denote a batch of $N_B$ trajectories:
\[
\mathcal{L}_{\text{traj}} =
\frac{1}{N_B} \sum_{i=1}^{N_B}
\frac{\lVert \hat{\vu}^{\text{trajectory}}_i - \vu^{\text{trajectory}}_i \rVert}
{\lVert \vu^{\text{trajectory}}_i \rVert},
\]

\paragraph{Stage 2: Spatial refinement on high-resolution frames.}
Although Stage~1 provides strong spatio-temporal compression, it does not fully capture high-resolution information. Since high-resolution trajectory-wise training exceeds GPU memory, we refine the auto-encoder frame-wise on high-resolution frames, spatially subsampled by a factor between $0.5$ and $1.0$. This shorter refinement stage incorporates fine-scale spatial information at substantially lower cost. Let $B$ be a batch of $N_B$ trajectories, the analogous frame-wise loss is
\[
\mathcal{L}_{\text{frame}} =
\frac{1}{N_B} \sum_{i=1}^{N_B}
\frac{\lVert \hat{\vu}^{\text{frame}}_i - \vu^{\text{frame}}_i \rVert}
{\lVert \vu^{\text{frame}}_i \rVert},
\]

\paragraph{Stage 3: Flow-matching in latent space.}
The auto-encoder is then frozen, and the transformer processor is trained in latent space using the flow-matching objective described in \cref{sssec:process}. Given an encoded latent trajectory $\vz$, we randomly noise $80\%$ of its latent frames, denoted $\vz^{\mathcal{M'}}$, and condition on the remaining observed $20\%$, denoted $\vz^{\mathcal{O'}}$; see \cref{fig:training}. The model learns the conditional distribution
$p(\vz^{\mathcal{M'}} \mid \vz^{\mathcal{O'}}, \bm{\gamma})$
through a flow-based transport from the base distribution $p_0=\mathcal{N}(0,I)$ to the target distribution. The transport is parameterized by the ODE
\begin{equation}
    \frac{d \mathbf{z}^r}{d r} = v(\mathbf{z}^r, r),
\end{equation}
where $r \in [0,1]$ is a denoising index and $v$ is a neural-network velocity field. During training, intermediate points are sampled as
\begin{equation}
    \mathbf{z}^r = r \vz^{\mathcal{M'}} + (1-r)\bm{\epsilon},
    \quad \bm{\epsilon} \sim \mathcal{N}(0,I).
\end{equation}
Conditioned on $\vz^{\mathcal{O'}}$ and, when available, PDE parameters $\bm{\gamma}$, the transformer takes
$(\mathbf{z}^r, \vz^{\mathcal{O'}}, \bm{\gamma}, r)$
as input and predicts the transport direction. It is trained with the following loss:
\begin{equation}
    \mathcal{L}_{\texttt{FM}} =
    \mathbb{E}_{\bm{\epsilon}, r}
    \left[
    \left\|
    \mathcal{G}_{\theta}(\mathbf{z}^r, \vz^{\mathcal{O'}}, \bm{\gamma}, r)
    - (\vz^{\mathcal{M'}} - \bm{\epsilon})
    \right\|_2^2
    \right],
\end{equation}
which aligns the predicted velocity with the true transport direction from $\bm{\epsilon}$ to $\vz^{\mathcal{M'}}$. Inference then proceeds as described in \cref{ssec:inference_model}.

\section{Experiments}
\label{sec:experiments}

We extensively evaluate ECHO on a diverse set of dynamical systems from public benchmarks (\cref{ssec:datasets}), with experiments on 2D (\cref{ssec:dyn_forecasting_exp}) and 3D (\cref{ssec:3d-main}) regular grids and irregular grids (\cref{ssec:no_exp}). The regular-grid benchmarks allow comparison with many state-of-the-art models that operate directly in physical space and are restricted to uniform meshes, providing evidence for our processor design (i.e. full-trajectory generation). The irregular-grid benchmarks place ECHO in a more demanding regime with complex settings (including partial observations) and varying numbers of mesh points, and are used to compare the efficiency of operator-based approaches that handle arbitrary resolutions. Finally, \cref{ssec:add_exp} introduces additional experiment in \cref{app:add_exp}. 


\paragraph{Datasets.}
\label{ssec:datasets}

The 2D and 3D PDE benchmarks span dense regular grids, spherical grids, irregular meshes, parametric dynamics and multi-channel physical fields. Dataset statistics are summarized in \Cref{tab:datasets} in the appendix. In particular, we generated high-resolution 2D vorticity trajectories on a dense $1024 \times 1024$ grid, with $20$ frames per trajectory, yielding more than $20$M spatio-temporal values. We also evaluate on large-scale 3D benchmarks from \citet{ohana2024well}, including MHD and Turbulence Gravity Cooling, which contain approximately $29$M and $25$M values per trajectory segment, respectively. Finally, we include Eagle~\citep{janny2023eagle} and Cylinder Flow~\citep{pfaff2020} to evaluate ECHO on irregular meshes and non-uniform geometries.

\subsection{Evaluation on 2D regular grids}
\label{ssec:dyn_forecasting_exp}

\paragraph{Setting.}  
At inference, ECHO addresses multiple tasks in a zero-shot manner. The comparison is performed on regular grids, where state-of-the-art baselines admit particularly efficient implementations. We compare against neural solvers trained for time-stepping, the standard approach for time-dependent PDEs, on forward and backward prediction. We call \emph{backward prediction} the generation of the earlier part of a trajectory from its last observed frames. This temporal task differs from classical PDE inverse problems, which recover unknown coefficients, forcing terms or initial conditions from observations of the solution. Since the baselines handle one task at a time, we train them separately for each direction, while ECHO handles both without retraining. For the forward (resp.\ backward) task, we use the first (resp.\ last) $4$ timesteps as context and generate all remaining frames of the trajectory. Regarding the interpolation task, the results are presented in \cref{app:ssec:addtasks}. All models, including generative ones, are evaluated with a single sample per test trajectory, and we report the relative $L^2$ error. Probabilistic metrics computed on ensembles are reported in \cref{app:ssec:probabilistic}.

\paragraph{Baselines.}  
We benchmark ECHO against both deterministic and generative baselines. Deterministic models include the transformer-based neural solvers BCAT~\citep{liu2025bcat}, AViT~\citep{mccabe2023multiple}, Transolver++~\citep{luotransolver++}, and the classical Fourier Neural Operator (FNO). We also evaluate a deterministic variant of ECHO to isolate the gains brought by the generative formulation; this variant, like ECHO, supports multi-task zero-shot inference. As a generative baseline, we consider ENMA~\citep{enma}, an auto-regressive model trained with a next-token strategy akin to large language models – implementation details are provided in appendix \ref{app:imp_details}.

\begin{table*}[htbp]
\centering
\caption{Comparison of model performance across four dynamical systems.
Errors are relative $L^2$ errors computed from a single sample per test
trajectory, for backward (\emph{Back.}) and forward (\emph{For.}) prediction.
Latency is the time needed to generate a whole trajectory, averaged over the
datasets. Lower is better for all metrics. The ratio under each dataset name is
the compression ratio of the latent models (ECHO, ENMA); the other baselines
operate at full resolution ($\times 1$). Best results in \textbf{bold}, second
best \underline{underlined}; ``--'' indicates non-convergence.}
\label{tab:model-comparison}
\footnotesize
\renewcommand{\arraystretch}{1.15}
\setlength{\tabcolsep}{0pt}
\begin{tabular*}{\textwidth}{@{\extracolsep{\fill}}l c cc cc cc cc@{}}
\toprule
& & \multicolumn{2}{c}{Rayleigh--B\'enard}
  & \multicolumn{2}{c}{Gray--Scott}
  & \multicolumn{2}{c}{Active Matter}
  & \multicolumn{2}{c}{ASM} \\
& & \multicolumn{2}{c}{\footnotesize ($\times 64$)}
  & \multicolumn{2}{c}{\footnotesize ($\times 32$)}
  & \multicolumn{2}{c}{\footnotesize ($\times 176$)}
  & \multicolumn{2}{c}{\footnotesize ($\times 48$)} \\
\cmidrule(lr){3-4}\cmidrule(lr){5-6}\cmidrule(lr){7-8}\cmidrule(l){9-10}
Model & Latency (s) & Back. & For. & Back. & For. & Back. & For. & Back. & For. \\
\midrule
\multicolumn{10}{@{}l}{\textit{Deterministic models}} \\
\addlinespace[2pt]
FNO          & \bsn{4.42}{-2} & \sn{2.47}{3}   & \sn{4.23}{-1}  & --              & --             & --             & --             & 1.87           & 1.52 \\
Transolver++ & \sn{3.52}{-1}  & \sn{6.34}{-1}  & \sn{3.31}{-1}  & \sn{4.43}{-1}   & \sn{2.34}{-1}  & \sn{7.33}{-1}  & \sn{6.91}{-1}  & 1.03           & \sn{9.64}{-1} \\
BCAT         & \sn{1.18}{-1}  & \sn{1.91}{-1}  & \sn{1.06}{-1}  & \sn{2.19}{-1}   & \sn{8.82}{-2}  & \sn{4.98}{-1}  & \sn{4.56}{-1}  & \sn{1.95}{-1}  & \sn{2.18}{-1} \\
AViT         & \sn{1.04}{-1}  & \sn{4.50}{-1}  & \sn{1.01}{-1}  & \sn{1.66}{-1}   & \sn{7.42}{-2}  & \sn{4.50}{-1}  & \sn{4.62}{-1}  & \bsn{1.04}{-1} & \usn{1.52}{-1} \\
ECHO         & \usn{4.89}{-2} & \sn{2.53}{-1}  & \sn{1.32}{-1}  & \usn{8.36}{-2}  & \sn{7.66}{-2}  & \sn{5.74}{-1}  & \sn{3.44}{-1}  & \sn{1.18}{-1}  & \sn{2.01}{-1} \\
\midrule
\multicolumn{10}{@{}l}{\textit{Generative models}} \\
\addlinespace[2pt]
ENMA         & 2.20           & \usn{1.71}{-1} & \usn{9.87}{-2} & \sn{1.08}{-1}   & \usn{5.44}{-2} & \usn{4.27}{-1} & \usn{3.33}{-1} & 1.01           & \sn{4.12}{-1} \\
ECHO         & \bsn{1.10}{-1} & \bsn{1.16}{-1} & \bsn{9.28}{-2} & \bsn{2.53}{-2}  & \bsn{5.12}{-2} & \bsn{1.71}{-1} & \bsn{2.87}{-1} & \usn{1.12}{-1} & \bsn{1.32}{-1} \\
\bottomrule
\end{tabular*}
\end{table*}

\paragraph{Results}
\Cref{tab:model-comparison} reports the relative $L^2$ error across four dynamical systems for both deterministic and generative surrogates.
Despite operating in a highly compressed latent space, whereas all baselines except ENMA work directly in physical space without information loss, \textbf{ECHO} achieves the best overall accuracy, attaining the lowest error on seven of the eight dataset--task pairs. FNO performs poorly and fails to converge on two of the four datasets. The deterministic version of ECHO remains competitive across benchmarks, while the generative formulation outperforms it on all eight dataset--task pairs, highlighting the benefits of stochastic trajectory modeling. We also report ECHO's performance beyond single-sample accuracy (\cref{app:ssec:probabilistic}). Overall, these results show ECHO's ability to combine high compression with generative modeling, delivering both accuracy and scalability for million-point PDE surrogates. We also evaluate the computation time required to generate a whole trajectory (Latency column in \cref{tab:model-comparison}) and show that generative ECHO is on par with the deterministic baselines.

\subsection{Evaluation on 3D regular dynamical systems}
\label{ssec:3d-main}
To further assess the scalability of ECHO, we provide in \cref{tab:model-comparison-3d} an evaluation on 3D spatio-temporal datasets with trajectory segments respectively with about $29M$ and $25M$ points. The datasets are described in appendix \ref{app:datasetsdetails} - \cref{tab:datasets}.

\begin{table}[htbp]
\vspace{-5mm}
\centering
\caption{Forward time-stepping on 3D benchmarks. Relative $L^2$ error over
the first frame (1), the first 8 frames (1--8) and the first 16 frames (1--16)
of the predicted trajectory. Lower is better. Best results in \textbf{bold},
second best \underline{underlined}; ``--'' indicates non-convergence.}
\label{tab:model-comparison-3d}
\small
\renewcommand{\arraystretch}{1.15}
\setlength{\tabcolsep}{8pt}
\begin{tabular}{l ccc@{\hspace{24pt}}ccc}
\toprule
& \multicolumn{3}{c}{MHD} & \multicolumn{3}{c}{TGC} \\
\cmidrule(lr){2-4}\cmidrule(lr){5-7}
Model & 1 & 1--8 & 1--16 & 1 & 1--8 & 1--16 \\
\midrule
FNO  & \bsn{1.69}{-1} & --             & --             & \bsn{3.15}{-2} & --             & -- \\
AViT & \sn{3.40}{-1}  & \sn{5.68}{-1}  & \sn{7.51}{-1}  & \sn{1.26}{-1}  & \usn{6.22}{-1} & \underline{1.32} \\
BCAT & \sn{3.29}{-1}  & \usn{5.43}{-1} & \usn{6.70}{-1} & \usn{1.04}{-1} & \sn{9.89}{-1}  & 1.62 \\
ECHO & \usn{2.70}{-1} & \bsn{3.82}{-1} & \bsn{5.50}{-1} & \sn{1.13}{-1}  & \bsn{5.55}{-1} & \bsn{6.67}{-1} \\
\bottomrule
\end{tabular}
\end{table}

\paragraph{Results} 
ECHO successfully scales to 3D scenarios compared to existing baselines. Its full-trajectory generation design accumulates less errors than other baselines and makes it competitive on these datasets. In particular, we observe that FNO performs very well on first frame generation, but quickly diverges as the dynamic evolve, whereas ECHO provides consistent prediction along the entire trajectory. 

\subsection{Evaluation on irregular meshes}
\label{ssec:no_exp}

\paragraph{Setting.} Following ECHO's success on regular grids (\cref{ssec:dyn_forecasting_exp}), we now consider demanding irregular, dense meshes. We compare arbitrary-resolution neural operators that operate in a compressed latent space and can be queried at arbitrary resolutions. Unless stated otherwise, all methods use matched encoder–decoder compression rates. Each encoder-decoder is first trained and evaluated on reconstruction to assess physical information preservation. Next, to test latent expressivity on a forward task, we fix ECHO as the shared processor due to its superior performance (\cref{tab:model-comparison}). All methods thus generate one-shot trajectories (\cref{tab:exp_no,tab:1024x1024}). We evaluate two scenarios: fully observed (100\% input available) and partially observed (20\% randomly sampled points predicting the full grid).

\paragraph{Baselines.}  

We benchmark ECHO against a diverse set of encoders: GINO~\citep{li2023geometry,zhou2025} 
implicit neural representation models CORAL~\citep{serrano2023}; transformer-based auto-encoders AROMA and ENMA~\citep{enma}; and CALM-PDE~\citep{calmpde}, a continuous-convolution model – details in appendix \ref{app:imp_details}. 

\begin{table}[htbp]
\centering
\caption{Comparison of encoder--decoder modules from operator surrogates on
irregular scenarios. \textit{Rec.} and \textit{For.} denote reconstruction and
forward generation; \textit{Irr.} gives the strategy for handling irregular
points. $\mathcal{X}_{te}$ is the spatial sampling ratio of the conditioning
frames at inference. Colors indicate encoder scalability:
\textcolor{LimeGreen}{green} = no issues; \textcolor{Goldenrod}{yellow} = minor
modification; \textcolor{BrickRed}{red} = model size modification required to
fit in GPU memory (see \cref{app:imp_details}). Metrics are relative $L^2$
errors on the full grid of the test set (lower is better). Best results in
\textbf{bold}, second best \underline{underlined}.}
\label{tab:exp_no}
\footnotesize
\renewcommand{\arraystretch}{1.15}
\setlength{\tabcolsep}{4pt}
\begin{tabular}{ll cc cc cc cc}
\toprule
& & \multicolumn{2}{c}{Vorticity} & \multicolumn{2}{c}{Shallow-Water}
  & \multicolumn{2}{c}{Eagle} & \multicolumn{2}{c}{Cylinder Flow} \\
\cmidrule(lr){3-4}\cmidrule(lr){5-6}\cmidrule(lr){7-8}\cmidrule(lr){9-10}
Irr. & Encoder & Rec. & For. & Rec. & For. & Rec. & For. & Rec. & For. \\
\midrule
\multicolumn{10}{l}{\textit{Full resolution} ($\mathcal{X}_{te} = 100\%$)} \\
\addlinespace[2pt]
Graph & GINO
  & \cellcolor{LimeGreen}\sn{9.99}{-1} & 1.00
  & \cellcolor{LimeGreen}\sn{8.68}{-1} & 1.11
  & \cellcolor{LimeGreen}\sn{5.58}{-1} & 1.89
  & \cellcolor{LimeGreen}\sn{7.94}{-1} & \sn{8.65}{-1} \\
INR & CORAL
  & \cellcolor{LimeGreen}\sn{5.13}{-1} & 1.34
  & \cellcolor{LimeGreen}\sn{2.29}{-1} & \sn{6.89}{-1}
  & \cellcolor{LimeGreen}\sn{6.04}{-1} & 1.54
  & \cellcolor{LimeGreen}\sn{2.94}{-1} & \sn{5.08}{-1} \\
\multirow{2}{*}{Attention} & AROMA
  & \cellcolor{softred}\sn{5.13}{-1}    & \sn{8.42}{-1}
  & \cellcolor{softred}\usn{2.51}{-2}   & \usn{4.21}{-2}
  & \cellcolor{LimeGreen}\usn{2.81}{-1} & \usn{3.09}{-1}
  & \cellcolor{LimeGreen}\bsn{2.76}{-2} & \sn{2.21}{-1} \\
& ENMA
  & \cellcolor{softred}\sn{4.38}{-1}   & \sn{4.38}{-1}
  & \cellcolor{softred}\sn{7.17}{-2}   & \sn{7.35}{-2}
  & \cellcolor{LimeGreen}\sn{2.85}{-1} & \sn{3.29}{-1}
  & \cellcolor{LimeGreen}\sn{7.50}{-2} & \usn{1.04}{-1} \\
\multirow{2}{*}{Convolution} & CALM-PDE
  & \cellcolor{Goldenrod}\usn{2.69}{-1} & \usn{4.20}{-1}
  & \cellcolor{Goldenrod}\sn{2.56}{-1}  & \sn{2.80}{-1}
  & \cellcolor{Goldenrod}\sn{9.01}{-1}  & \sn{9.24}{-1}
  & \cellcolor{LimeGreen}\sn{2.44}{-1}  & \sn{3.56}{-1} \\
& ECHO
  & \cellcolor{LimeGreen}\bsn{6.88}{-2} & \bsn{1.71}{-1}
  & \cellcolor{LimeGreen}\bsn{1.78}{-2} & \bsn{1.96}{-2}
  & \cellcolor{LimeGreen}\bsn{1.71}{-1} & \bsn{2.55}{-1}
  & \cellcolor{LimeGreen}\usn{4.68}{-2} & \bsn{8.82}{-2} \\
\midrule
\multicolumn{10}{l}{\textit{Subsampled} ($\mathcal{X}_{te} = 20\%$)} \\
\addlinespace[2pt]
Graph & GINO
  & \sn{9.99}{-1} & 1.00
  & \sn{8.73}{-1} & 1.71
  & \sn{6.77}{-1} & 1.71
  & \sn{7.95}{-1} & \sn{8.65}{-1} \\
INR & CORAL
  & \sn{5.19}{-1} & 1.33
  & \sn{2.29}{-1} & \sn{6.87}{-1}
  & \sn{6.31}{-1} & 1.54
  & \sn{3.31}{-1} & \sn{5.18}{-1} \\
\multirow{2}{*}{Attention} & AROMA
  & \sn{5.14}{-1}  & \sn{8.43}{-1}
  & \usn{3.40}{-2} & \usn{4.57}{-2}
  & \sn{4.01}{-1}  & \sn{4.83}{-1}
  & \bsn{4.59}{-2} & \sn{2.27}{-1} \\
& ENMA
  & \sn{4.41}{-1}  & \sn{4.41}{-1}
  & \sn{8.16}{-2}  & \sn{8.09}{-1}
  & \usn{3.66}{-1} & \bsn{3.99}{-1}
  & \sn{8.68}{-2}  & \usn{1.20}{-1} \\
\multirow{2}{*}{Convolution} & CALM-PDE
  & \bsn{2.95}{-1} & \usn{4.38}{-1}
  & \sn{2.77}{-1}  & \sn{2.95}{-1}
  & \sn{9.61}{-1}  & \sn{9.75}{-1}
  & \sn{2.68}{-1}  & \sn{3.77}{-1} \\
& ECHO
  & \usn{3.66}{-1} & \bsn{2.21}{-1}
  & \bsn{2.61}{-2} & \bsn{2.61}{-2}
  & \bsn{2.46}{-1} & \usn{4.72}{-1}
  & \usn{5.51}{-2} & \bsn{1.03}{-1} \\
\bottomrule
\end{tabular}
\end{table}
These experiments highlight ECHO’s consistent superiority. While graph and INR approaches fail to encode these complex dynamics, attention-based AROMA proves powerful on smaller datasets like \textit{Cylinder Flow} but struggles scaling to \textit{Vorticity} due to high memory costs. Conversely, the convolution-based CALM-PDE handles \textit{Vorticity} better given its lighter footprint. Alongside ECHO, this shows that hierarchical convolutional architectures excel on large-scale data. Furthermore, operating in a spatio-temporally compressed latent space mitigates generation error accumulation. This is evident on \textit{Cylinder Flow}: AROMA yields the best reconstruction, yet ECHO achieves higher forward task accuracy.

\begin{table}[htbp]
\centering
\caption{Super-resolution forecasting on a $1024\times1024$ \textit{Vorticity}
grid (200\% of the training grid, more than 20M points). Relative $L^2$ error
of forward generation (\textit{For.}); OOM = out of memory on an 80\,GB H100.}
\label{tab:1024x1024}
\small
\renewcommand{\arraystretch}{1.15}
\setlength{\tabcolsep}{8pt}
\begin{tabular}{lcccccc}
\toprule
Model & GINO & CORAL & AROMA & CALM-PDE & Transolver++ & ECHO \\
\midrule
For.  & OOM  & OOM   & OOM   & OOM      & OOM          & \bsn{3.88}{-1} \\
\bottomrule
\end{tabular}
\end{table}

\begin{figure}[htbp]
\centering
\includegraphics[width=0.6\textwidth]{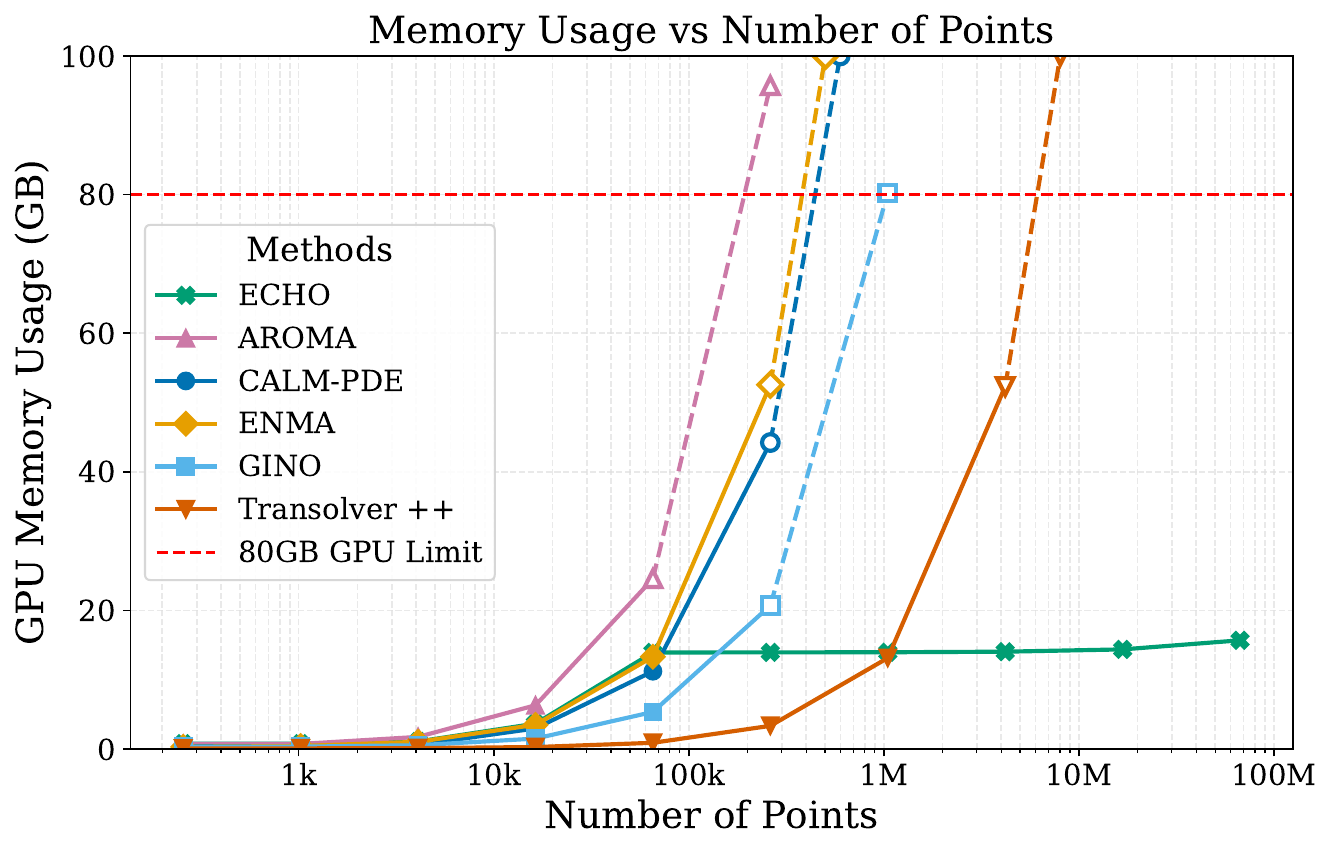}
\caption{Scaling comparison: memory consumption at inference vs.\ number of
points in the input mesh.}
\label{fig:memory_scaling}
\end{figure}

\paragraph{Large-scale evaluation}
To further study the scaling properties of ECHO and other baselines, we perform two scaling experiments, pushing the models beyond the capacity of our GPU. First, \Cref{tab:1024x1024} presents results for a spatio-temporal forecasting task on our large-scale \textit{Vorticity} dataset with a grid size of $1024 \times 1024$. All baselines ran out of memory on the H100 GPU with 80\,GB used for these experiments; only ECHO scaled successfully. 
Then, \Cref{fig:memory_scaling} shows the memory consumption at inference for several grid sizes. This figure shows the superior scaling capacity of ECHO as it surpasses all baselines in term of memory consumption for grid sizes beyond $100$k points and scales up to $100$M points (details in \cref{app:ssec:fairness}). 

\subsection{Additional experiments}
\label{ssec:add_exp}
Extensive additional experiments are provided in \cref{app:add_exp}. Specifically, we further stress-test ECHO with extreme settings: OOD evaluation on PDE parameters \cref{ssec:ood_vorticity} - Long range prediction  beyond the training horizon \cref{app:sec:long_range} - Data sampling \cref{app:ssec:generative_expe} - Probabilistic evaluation \cref{app:ssec:probabilistic} - Initial Value Problem \cref{app:ssec:addexp_ivp} - Interpolation task complete evaluation \cref{app:ssec:addtasks} - Additional Evaluation on spatially sparse data appendix \ref{app:ssec:addexp_aeood}. We also provide ablations of key components of ECHO in appendix \ref{app:ssec:addexp_ablation}: hierarchical compression, refinement stage and the generative process. Finally, qualitative samples generated by ECHO are presented in \cref{app:visu}.

\section{Conclusion and limitations}
\label{sec:conclusion}

We introduced ECHO, a generative transformer-based neural operator for large-scale PDE modeling on arbitrary domains. Our central premise is that trajectory-level generative modeling becomes tractable at scale only when performed in strongly compressed yet information-preserving spatio-temporal latent spaces. To this end, ECHO combines hierarchical spatio-temporal compression with latent trajectory generation, enabling efficient modeling of million-point systems while mitigating the error accumulation associated with autoregressive rollouts.
Experiments across large-scale 2D and 3D benchmarks demonstrate strong performance on forward prediction, interpolation, backward prediction, and long-horizon forecasting, including simulations with up to 20 million points.

\paragraph{Limitations.}
The trade-off between compression and reconstruction fidelity is central to this framework and is analyzed empirically in Appendix \ref{ssec:ablation_studies} and \cref{fig:memory_scaling}. More generally, ECHO relies on the existence of compact spatio-temporal latent representations on which trajectory-level generation remains stable and informative. Consequently, extreme compression may affect the recovery of very fine-scale or rapidly evolving structures. Finally, as with any data-driven surrogate model, performance depends on adequate coverage of the underlying data distribution.

\section{Acknowledgments}
We acknowledge the financial support provided by DL4CLIM (ANR-19-CHIA-0018-01), DEEPNUM (ANR-21-CE23-0017-02), PHLUSIM (ANR-23-CE23-0025-02), MELISSA (ANR-24-CE23-7140-01), PEPR Sharp (ANR-23-PEIA0008, ANR, FRANCE 2030), and  cluster IA  PostGenAI@Paris (ANR-23-IACL-0007).

This project was provided with computer and storage resources
by GENCI at IDRIS thanks to the grants 2025-AD011016890R1, 2025-AD011014938R1, 2026-AD011016890R2, 2026-AD011014938R2, and 2025-A0191016890 on the supercomputer Jean Zay’s V100/A100/H100 partitions.

\clearpage

\bibliography{neurips_2026}
\bibliographystyle{icml2026}


\appendix
\input{appendix}


\newpage


\end{document}

%% file: math_commands.tex
\usepackage{amsmath,amsfonts,bm}

\def\eqref#1{equation~\ref{#1}}

\def\1{\bm{1}}

\def\vb{{\bm{b}}}
\def\vc{{\bm{c}}}

\def\vf{{\bm{f}}}

\def\vu{{\bm{u}}}

\def\vz{{\bm{z}}}

\DeclareMathAlphabet{\mathsfit}{\encodingdefault}{\sfdefault}{m}{sl}
\SetMathAlphabet{\mathsfit}{bold}{\encodingdefault}{\sfdefault}{bx}{n}



%% file: appendix.tex
\cleardoublepage
\phantomsection
\addcontentsline{toc}{section}{Appendix: Table of Contents}
\pdfbookmark[1]{Appendix: Table of Contents}{appendix_toc}
\section*{Appendix: Table of Contents}

\begin{itemize}
    \item[\ref{statements}] \textbf{\hyperref[statements]{Statements}} \dotfill \pageref{statements}
    \begin{itemize}
        \item[\ref{impact}] \hyperref[impact]{Impact Statement} \dotfill \pageref{impact}
        \item[\ref{reproductibility}] \hyperref[reproductibility]{Reproductibility Statement} \dotfill \pageref{reproductibility}
        \item[\ref{llms}] \hyperref[llms]{Use of LLMs} \dotfill \pageref{llms}
    \end{itemize}
    
    \item[\ref{app:rw}] \textbf{\hyperref[app:rw]{Related works}} \dotfill \pageref{app:rw}
    \begin{itemize}
        \item[\ref{app:ssec_ol}] \hyperref[app:ssec_ol]{Operator learning} \dotfill \pageref{app:ssec_ol}
        \item[\ref{app:ssec_gen}] \hyperref[app:ssec_gen]{Generative models} \dotfill \pageref{app:ssec_gen}
    \end{itemize}

    \item[\ref{app:datasetsdetails}] \textbf{\hyperref[app:datasetsdetails]{Dataset details}} \dotfill \pageref{app:datasetsdetails}
    \begin{itemize}
        \item[\ref{app:ssec_vorticity}] \hyperref[app:ssec_vorticity]{Vorticity} \dotfill \pageref{app:ssec_vorticity}
        \item[\ref{app:ssec_shallowwater}] \hyperref[app:ssec_shallowwater]{Shallow-water} \dotfill \pageref{app:ssec_shallowwater}
        \item[\ref{app:ssec_eagle}] \hyperref[app:ssec_eagle]{Eagle} \dotfill \pageref{app:ssec_eagle}
        \item[\ref{app:ssec_cylinder}] \hyperref[app:ssec_cylinder]{Cylinder-flow} \dotfill \pageref{app:ssec_cylinder}
        \item[\ref{app:ssec_rb}] \hyperref[app:ssec_rb]{Rayleigh-Bénard} \dotfill \pageref{app:ssec_rb}
        \item[\ref{app:ssec_gs}] \hyperref[app:ssec_gs]{Gray-Scott} \dotfill \pageref{app:ssec_gs}
        \item[\ref{app:ssec_activematter}] \hyperref[app:ssec_activematter]{Active matter} \dotfill \pageref{app:ssec_activematter}
        \item[\ref{app:ssec_asm}] \hyperref[app:ssec_asm]{Acoustic scattering maze} \dotfill \pageref{app:ssec_asm}
        \item[\ref{app:ssec_mhd}] \hyperref[app:ssec_mhd]{Magnetohydrodynamic (MHD)} \dotfill \pageref{app:ssec_mhd}
        \item[\ref{app:ssec_tgc}] \hyperref[app:ssec_tgc]{Turbulence gravity cooling (TGC)} \dotfill \pageref{app:ssec_tgc}
    \end{itemize}

    \item[\ref{app:arch}] \textbf{\hyperref[app:arch]{Architecture details}} \dotfill \pageref{app:arch}
    \begin{itemize}
        \item[\ref{app:Inference_instances}] \hyperref[app:Inference_instances]{Inference model} \dotfill \pageref{app:Inference_instances}
        \item[\ref{app:arch_ae}] \hyperref[app:arch_ae]{Encoder-decoder} \dotfill \pageref{app:arch_ae}
        \begin{itemize}
            \item[\ref{app:sssec-interplayer}] \hyperref[app:sssec-interplayer]{Interpolation with continuous convolution} \dotfill \pageref{app:sssec-interplayer}
            \item[\ref{app:sssec-complayer}] \hyperref[app:sssec-complayer]{Spatial and temporal compression} \dotfill \pageref{app:sssec-complayer}
        \end{itemize}
    \end{itemize}

    \item[\ref{app:imp_details}] \textbf{\hyperref[app:imp_details]{Implementation details}} \dotfill \pageref{app:imp_details}
    \begin{itemize}
        \item[\ref{app:ssec_imp_process}] \hyperref[app:ssec_imp_process]{Evaluation protocol of the process} \dotfill \pageref{app:ssec_imp_process}
        \begin{itemize}
            \item[\ref{app:sssec_baselinear}] \hyperref[app:sssec_baselinear]{Baseline details} \dotfill \pageref{app:sssec_baselinear}
            \item[\ref{app:sssec_trainar}] \hyperref[app:sssec_trainar]{Training details} \dotfill \pageref{app:sssec_trainar}
            \item[\ref{app:ssec:fairness}] \hyperref[app:ssec:fairness]{Fairness of Baseline Comparisons and Scaling Limits.} \dotfill \pageref{app:ssec:fairness}
        \end{itemize}
        \item[\ref{app:ssec_encedecimp}] \hyperref[app:ssec_encedecimp]{Encoder-decoder implementation protocol} \dotfill \pageref{app:ssec_encedecimp}
        \begin{itemize}
            \item[\ref{app:sssec_archencdec}] \hyperref[app:sssec_archencdec]{ECHO encoder-decoder architecture} \dotfill \pageref{app:sssec_archencdec}
            \item[\ref{app:ssec_basdetails_encdec}] \hyperref[app:ssec_basdetails_encdec]{Baseline details} \dotfill \pageref{app:ssec_basdetails_encdec}
            \item[\ref{app:sssec_train_enc_dec}] \hyperref[app:sssec_train_enc_dec]{Training details} \dotfill \pageref{app:sssec_train_enc_dec}
        \end{itemize}
    \end{itemize}

    \item[\ref{app:add_exp}] \textbf{\hyperref[app:add_exp]{Additional experiments}} \dotfill \pageref{app:add_exp}
    \begin{itemize}
        \item[\ref{ssec:ood_vorticity}] \hyperref[ssec:ood_vorticity]{Out-of-distribution evaluation} \dotfill \pageref{ssec:ood_vorticity}
        \item[\ref{app:sec:long_range}] \hyperref[app:sec:long_range]{Long-range prediction} \dotfill \pageref{app:sec:long_range}
        \begin{itemize}
            \item[\ref{ssec:long-range-vort}] \hyperref[ssec:long-range-vort]{Long-range prediction on Vorticity} \dotfill \pageref{ssec:long-range-vort}
            \item[\ref{app:ssec:long_range_sw}] \hyperref[app:ssec:long_range_sw]{Long-range prediction on Shallow-Water} \dotfill \pageref{app:ssec:long_range_sw}
            \item[\ref{app:ssec:long_range_vort_spec}] \hyperref[app:ssec:long_range_vort_spec]{Spectral analysis on the vorticity generation} \dotfill \pageref{app:ssec:long_range_vort_spec}
        \end{itemize}
        \item[\ref{app:ssec:generative_expe}] \hyperref[app:ssec:generative_expe]{Generative data sampling} \dotfill \pageref{app:ssec:generative_expe}
        \item[\ref{app:ssec:probabilistic}] \hyperref[app:ssec:probabilistic]{Probabilistic evaluation} \dotfill \pageref{app:ssec:probabilistic}
        \item[\ref{app:ssec:addexp_ivp}] \hyperref[app:ssec:addexp_ivp]{Initial value problem} \dotfill \pageref{app:ssec:addexp_ivp}
        \item[\ref{app:ssec:addtasks}] \hyperref[app:ssec:addtasks]{Interpolation tasks} \dotfill \pageref{app:ssec:addtasks}
        \item[\ref{app:ssec:addexp_aeood}] \hyperref[app:ssec:addexp_aeood]{Encoding sparse data} \dotfill \pageref{app:ssec:addexp_aeood}
        \item[\ref{app:ssec:addexp_ablation}] \hyperref[app:ssec:addexp_ablation]{Ablation studies} \dotfill \pageref{app:ssec:addexp_ablation}
        \begin{itemize}
            \item[\ref{ssec:ablation_studies}] \hyperref[ssec:ablation_studies]{Impact of hierarchical compression} \dotfill \pageref{ssec:ablation_studies}
            \item[\ref{app:sssec_refine}] \hyperref[app:sssec_refine]{Impact of the refinement stage} \dotfill \pageref{app:sssec_refine}
            \item[\ref{app:sssec_gen_ablation}] \hyperref[app:sssec_gen_ablation]{Ablations on the generative process} \dotfill \pageref{app:sssec_gen_ablation}
        \end{itemize}
    \end{itemize}

    \item[\ref{app:visu}] \textbf{\hyperref[app:visu]{Visualizations}} \dotfill \pageref{app:visu}
    \begin{itemize}
        \item[\ref{app:ssec:visu_Vorticity}] \hyperref[app:ssec:visu_Vorticity]{Vorticity} \dotfill \pageref{app:ssec:visu_Vorticity}
        \item[\ref{app:ssec:visu_gs}] \hyperref[app:ssec:visu_gs]{Gray-Scott} \dotfill \pageref{app:ssec:visu_gs}
        \item[\ref{app:ssec:visu_rb}] \hyperref[app:ssec:visu_rb]{Rayleigh-Bénard} \dotfill \pageref{app:ssec:visu_rb}
        \item[\ref{app:ssec:visu_am}] \hyperref[app:ssec:visu_am]{Active-Matter} \dotfill \pageref{app:ssec:visu_am}
        \item[\ref{app:ssec:visu_asm}] \hyperref[app:ssec:visu_asm]{Acoustic Scattering Maze} \dotfill \pageref{app:ssec:visu_asm}
        \item[\ref{app:ssec:visu_eagle}] \hyperref[app:ssec:visu_eagle]{Eagle} \dotfill \pageref{app:ssec:visu_eagle}
    \end{itemize}
\end{itemize}
\clearpage

\clearpage
\section{Statements}
\label{statements}

\subsection{Impact Statement}
\label{impact}
PDEs are involved in many applications of science and engineering. This paper introduces a new method to improve the performance and scalability of surrogate models for solving PDEs. While we do not directly target such real-world applications in this paper, one should acknowledge that solvers can be used in a wide range of scenarios including weather, climate, medical, aerodynamics, industry, and military applications. 

\subsection{Reproducibility Statement}
\label{reproductibility}
We provide implementation details in \cref{app:imp_details} including architecture details, training hyper-parameters and baseline descriptions. Moreover, we will publicly release the code base and the generated dataset upon acceptance. All our experiments run on Nvidia H100 GPUs with 80Gb memory. 
We estimate the total compute budget for the development of ECHO to be approximately of 50 days of compute.

\subsection{Use of LLMs}
\label{llms}
LLMs were used for polishing the text formulation of this publication. 
  
\clearpage
\section{Related Works}
\label{app:rw}

\begin{table}[htbp]
\centering
\caption{Comparison of operator learning approaches for spatio-temporal PDE forecasting.}
\label{tab:method_comparison}
\footnotesize
\renewcommand{\arraystretch}{1.15}
\setlength{\tabcolsep}{0pt}
\newcommand{\tabhead}[1]{\begin{tabular}[b]{@{}c@{}}#1\end{tabular}}
\begin{tabular*}{\textwidth}{@{\extracolsep{\fill}}ll cccccc@{}}
\toprule
Model & Reference
  & \tabhead{Arbitrary\\domains}
  & \tabhead{Compression}
  & \tabhead{High-res.\\encoding}
  & \tabhead{Hierarchical\\encoding}
  & \tabhead{Generative\\modeling}
  & \tabhead{Multi-task\\prediction} \\
\midrule
\multicolumn{8}{@{}l}{\textit{Graph}} \\
MPPDE       & \citet{Brandstetter2022} & \greencheck & \redcross   & \redcross   & \redcross   & \redcross   & \redcross \\
RIGNO       & \citet{rigno}            & \greencheck & \greencheck & \redcross   & \redcross   & \redcross   & \redcross \\
\addlinespace[3pt]
\multicolumn{8}{@{}l}{\textit{Operator}} \\
Transolver++ & \citet{luotransolver++} & \greencheck & \redcross   & \greencheck & \redcross   & \redcross   & \redcross \\
GINO        & \citet{li2023geometry}   & \greencheck & \greencheck & \redcross   & \redcross   & \redcross   & \redcross \\
TEXT2PDE    & \citet{zhou2025}         & \greencheck & \greencheck & \redcross   & \greencheck & \greencheck & \redcross \\
\addlinespace[3pt]
\multicolumn{8}{@{}l}{\textit{INRs}} \\
CORAL       & \citet{serrano2023}      & \greencheck & \greencheck & \redcross   & \redcross   & \redcross   & \redcross \\
DiNo        & \citet{Yin2022}          & \greencheck & \greencheck & \redcross   & \redcross   & \redcross   & \redcross \\
\addlinespace[3pt]
\multicolumn{8}{@{}l}{\textit{Attention}} \\
UPT         & \citet{alkin2024upt}     & \greencheck & \greencheck & \redcross   & \redcross   & \redcross   & \redcross \\
ENMA        & \citet{enma}             & \greencheck & \greencheck & \redcross   & \greencheck & \greencheck & \redcross \\
\addlinespace[3pt]
\multicolumn{8}{@{}l}{\textit{Convolution}} \\
CALM-PDE    & \citet{calmpde}          & \greencheck & \greencheck & \redcross   & \greencheck & \redcross   & \redcross \\
ECHO        & Ours                     & \greencheck & \greencheck & \greencheck & \greencheck & \greencheck & \greencheck \\
\bottomrule
\end{tabular*}
\end{table}

\subsection{Operator Learning}
\label{app:ssec_ol}

Operator learning has emerged as a powerful paradigm for modeling mappings between infinite-dimensional function spaces, enabling neural surrogates to learn the solution operators of partial differential equations (PDEs) directly. Early breakthroughs such as DeepONet and the Fourier Neural Operator (FNO) established neural operators (NOs) as effective and mesh-independent tools for approximating PDE solution maps \citep{Li2020fno,Lu2019}. Since then, research has focused on improving both the expressiveness and scalability of NOs. For example, multi-scale modeling has been explored through factorized representations and wavelet-based decompositions \citep{gupta2021multiwavelet}, while latent-space formulations have been introduced to support complex and irregular geometries \citep{Tran2023ffno,li2023geometry}. Implicit neural representations (INRs), as in CORAL, have also been proposed to flexibly handle variable spatial discretizations at inference time \citep{Yin2022,serrano2023,wang2025gridmix}.

A major trend in recent years has been the adoption of \emph{attention-based architectures} for operator learning on irregular meshes. OFormer introduced transformers for embedding input–output function pairs, demonstrating their strong representation capabilities for operator approximation \citep{li2023transformer}. This line of work inspired more advanced designs such as GNOT and Transolver, which refine input encoding and generalization to complex domains \citep{hao2023gnot,wu2024Transolver}. However, transformer-based NOs often suffer from quadratic complexity and memory usage, making them difficult to scale to real-world scenarios where spatial resolutions can reach millions of points.

These limitations have motivated the development of \emph{efficient neural operator surrogates} aimed at reducing computational and memory costs without sacrificing accuracy. Perceiver-inspired approaches such as Aroma and UPT learn compact latent representations that decouple the number of tokens from the raw input size, significantly improving scalability \citep{serrano2024aroma,alkin2024upt}. While these methods achieve strong results on moderately large inputs, our empirical analysis in \cref{fig:experimental_analysis} (left) shows that attention-only latent encoders degrade at extreme compression ratios, limiting their effectiveness for high-resolution generative modeling.

A complementary direction has explored \emph{hierarchical compression}. CALM-PDE \citep{calmpde} demonstrated that progressively reducing spatial resolution via continuous convolutions improves efficiency while preserving fidelity. Building on this idea, we introduce a spatio-temporal hierarchical encoder that extends this principle beyond static fields to dense time-dependent trajectories. By gradually compressing both space and time from a dense latent grid to coarser levels, our approach constructs highly informative and compact latent spaces, enabling efficient generative modeling even on million-point spatio-temporal inputs. The continuous convolution that maps irregular inputs onto a regular latent grid is closely related to the \textsc{ConvDeepSet} layer of Convolutional Conditional Neural Processes \citep{gordon2020convcnp}, which also aggregates off-the-grid observations onto a grid permutation-invariantly through a learned kernel.

\subsection{Generative Models}
\label{app:ssec_gen}
While most operator learning methods are deterministic, generative modeling introduces key capabilities for physical systems—most notably the ability to capture uncertainty and represent one-to-many mappings. This is especially valuable in chaotic or partially observed regimes, where deterministic predictions can quickly diverge. Two main generative paradigms have emerged for scientific modeling: diffusion-based methods and autoregressive transformers.

\paragraph{Diffusion Transformers.}
Diffusion models synthesize data by learning to reverse a progressive noising process through a sequence of denoising steps \citep{ho2020denoising}. In computer vision, Latent Diffusion Transformers (DiTs) extend this approach by operating in the latent space of a variational autoencoder (VAE), achieving strong generative quality and scalability \citep{Peebles2022DiT}.  

Recently, diffusion has been adapted to physical modeling. \citet{kohl2024benchmarking} introduced an autoregressive diffusion framework for PDE dynamics, aiming to better capture stochasticity in turbulent flows. \citet{Lippe2023} improved the treatment of chaotic high-frequency regimes by modulating noise variance during denoising. Generative diffusion is particularly attractive under partial observations or incomplete physics \citep{diffusionpde}. For example, \citet{li2025videopde} reformulate PDE solving as a video inpainting problem to handle both forward and backward tasks, but the method remains limited to regular grids and is expensive at inference. \citet{zhou2025} further extend DiTs to physics-driven data generation from textual prompts.

Despite their flexibility, diffusion models are computationally heavy at inference due to the large number of denoising steps. \emph{Flow Matching} provides a more efficient alternative by learning a continuous-time velocity field that transports one distribution to another via an ordinary differential equation (ODE), enabling much faster sampling with far fewer steps \citep{lipman2023,lipman2024flowmatchingguidecode}. Flow matching has recently been applied to PDE surrogates \citep{holzschuh2025pdetransformer}, mostly in multi-physics settings and under fixed, regular discretizations. More recently, \citep{Li2025DiT}, proposed to use Flow matching to solve PDE framewise. The main difference with our proposed method lies in the formulation that we adopted, that allows us to solve several tasks without retraining. Moreover, ECHO makes use of continuous convolution, allowing our model to solve PDE on very dense grids. 

\paragraph{Autoregressive Transformers.}
Autoregressive (AR) models, originally designed for language modeling, have been extended to visual domains by sequentially generating spatial and temporal tokens. A common design couples a vector-quantized variational autoencoder (VQ-VAE) with a causal or bidirectional transformer to model discrete token sequences \citep{oord2017neural, esser2021taming, chang2022maskgit}. In video generation, \textit{Magvit} and \textit{Magvit2} \citep{yu2023magvit, yu2023language} use 3D CNNs to encode spatiotemporal structure and autoregressively predict quantized latent tokens for future frames.

This paradigm has recently been adapted to PDE surrogates. \textit{Zebra} \citep{serrano2024zebra} combines a spatial VQ-VAE with a causal transformer for in-context forecasting of dynamical systems. However, reliance on discrete codebooks can limit expressiveness and hinder accurate representation of fine-grained continuous physical phenomena. \citet{nguyen2025physix} extend this design to multi-physics settings by adding a decoder refinement stage to reduce quantization artifacts, at the cost of extra complexity. Alternatively, \citet{enma} replace discrete tokens with continuous ones and train next-token generation via a flow-matching objective.

Yet, whether discrete or continuous, autoregressive strategies remain prone to long-horizon error accumulation, particularly in compressed latent spaces where information loss amplifies drift from the true solution \citep{pedersen2025thermalizer}. Our work departs from the next-step paradigm: we train a flow-matching transformer to synthesize entire trajectory segments directly rather than predicting one token at a time. This design mitigates temporal drift and enables more stable long-range rollouts than both diffusion-based and deterministic baselines (see the middle panel of \cref{fig:experimental_analysis}).

\clearpage
\section{Dataset details}
\label{app:datasetsdetails}
The following section details the datasets used in the experiments. They are first summarized in \cref{tab:datasets} and then, we detail and illustrate each dataset. 

\begin{table}[htbp]
\centering
\caption{Datasets used in the experiments. Sizes are given as
$T \times (H \times W \textcolor{gray}{\times D}) \times C$, where the depth
$D$ only applies to 3D datasets. ASM = Acoustic Scattering Maze;
MHD = Magnetohydrodynamics; TGC = Turbulence Gravity Cooling;
BC = boundary conditions.}
\label{tab:datasets}
\footnotesize
\renewcommand{\arraystretch}{1.15}
\setlength{\tabcolsep}{5pt}
\newcommand{\tabhead}[1]{\begin{tabular}[b]{@{}c@{}}#1\end{tabular}}
\begin{tabular}{lrllc}
\toprule
Dataset & \# points & Grid & PDE parameters
  & \tabhead{Size\\$T \times (H \times W \textcolor{gray}{\times D}) \times C$} \\
\midrule
\multicolumn{5}{l}{\textit{2D}} \\
Vorticity \citep{koupai2024geps}             & 20M   & Dense     & Parametric      & $20\times\bm{1048576}\times1$ \\
Shallow-Water \citep{Yin2022}                & 2.6M  & Spherical & Single instance & $40\times32768\times2$ \\
Gray--Scott \citep{ohana2024well}            & 1.3M  & --        & Varying         & $40\times16384\times2$ \\
Rayleigh--B\'enard \citep{ohana2024well}     & 3.9M  & --        & Varying         & $15\times65536\times\bm{4}$ \\
ASM \citep{ohana2024well}                    & 2.9M  & --        & Varying         & $15\times65536\times\bm{3}$ \\
Active Matter \citep{ohana2024well}          & 14.4M & --        & Varying         & $20\times65536\times\bm{11}$ \\
Eagle \citep{janny2023eagle}                 & 0.2M  & Irregular & BC              & $20\times2500\times4$ \\
Cylinder Flow \citep{pfaff2020}              & 0.3M  & Irregular & Single instance & $60\times2000\times3$ \\
\midrule
\multicolumn{5}{l}{\textit{3D}} \\
MHD \citep{ohana2024well}                    & 29M   & --        & Varying         & $16\times262144\times\bm{7}$ \\
TGC \citep{ohana2024well}                    & 25M   & --        & Varying         & $16\times262144\times\bm{6}$ \\
\bottomrule
\end{tabular}
\end{table}

\subsection{Vorticity}
\label{app:ssec_vorticity}

We consider a 2D turbulence model and focus on the evolution of the vorticity field $\omega$, which captures the local rotation of the fluid and is defined as \( \omega = \nabla \times \mathbf{u} \), where $\mathbf{u}$ is the velocity field. The governing equation is:
\begin{equation}
    \frac{\partial \omega}{\partial t} + (\mathbf{u} \cdot \nabla) \omega - \nu \nabla^2 \omega = 0,
\end{equation}
where $\nu$ denotes the kinematic viscosity, defined as $\nu = 1/\text{Re}$. We sample $\nu \sim \mathcal{U}([10^{-3}, 10^{-2}])$. The initial conditions are generated from the energy spectrum:
\begin{equation}
    E(k) = \frac{4}{3} \sqrt{\pi} \left(\frac{k}{k_0}\right)^4 \frac{1}{k_0} \exp\left(-\left(\frac{k}{k_0}\right)^2\right),
\end{equation}
where $k_0$ denotes the characteristic wavenumber. Vorticity is linked to energy by the following equation : 
\begin{equation}
    \omega(k) = \sqrt{\frac{E(k)}{\pi k}}
\end{equation}
We generate very dense-grid with $1,024\times1,024$ spatial resolution and $20$ timesteps, resulting in trajectories with more than $20$M points.

\begin{figure}[htbp]
        \centering
        \includegraphics[width=\textwidth]{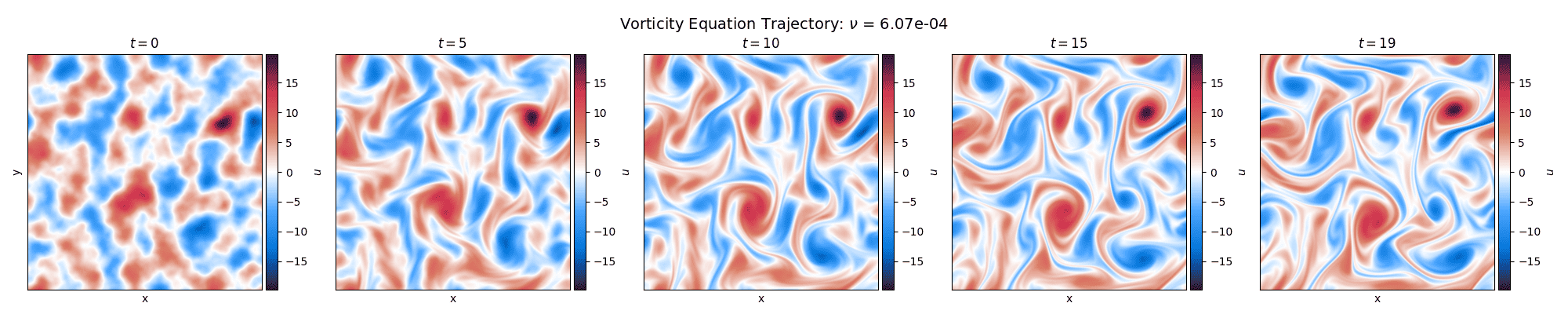}
        \caption{Sample of the Vorticity dataset. }
  \label{fig:vis_vorticity}
\end{figure}

\subsection{Shallow-Water}
\label{app:ssec_shallowwater}
\paragraph{3D-Spherical Shallow-Water (\textit{\textmd{Shallow-Water}}).} 
We consider the shallow-water equations on a rotating sphere, modeling large-scale atmospheric flows:
\begin{align}
    \frac{\partial u}{\partial t} &= -f \cdot k \times u - g \nabla h + \nu \Delta u, \\
    \frac{\partial h}{\partial t} &= -h \nabla \cdot u + \nu \Delta h,
\end{align}
where $k$ is the outward unit normal to the sphere, $u$ is the tangential velocity field (with vorticity $w = \nabla \times u$), and $h$ denotes the fluid height.  

Data is generated with the \textit{Dedalus} software \citep{Burns2020}, following \citet{Yin2022}. The setup produces symmetric dynamics in both hemispheres. The initial zonal velocity $u_0$ contains two symmetric jets parallel to latitude circles:
\begin{equation}
u_0(\phi,\theta) = 
\begin{cases}
\left( \tfrac{u_{max}}{e_n} \exp\!\Bigl(\tfrac{1}{(\phi - \phi_0)(\phi - \phi_1)} \Bigr), 0 \right), & \phi \in (\phi_0, \phi_1), \\[6pt]
\left( \tfrac{u_{max}}{e_n} \exp\!\Bigl(\tfrac{1}{(\phi + \phi_0)(\phi + \phi_1)} \Bigr), 0 \right), & \phi \in (-\phi_1, -\phi_0), \\[6pt]
(0,0), & \text{otherwise},
\end{cases}
\end{equation}
where $\phi$ and $\theta$ are latitude and longitude, $u_{max}$ is the maximum velocity, $\phi_0=\tfrac{\pi}{7}$, $\phi_1=\tfrac{\pi}{2}-\phi_0$, and $e_n=\exp\!\bigl(-\tfrac{4}{(\phi_1 - \phi_0)^2}\bigr)$.  

The initial fluid height $h_0$ is computed by solving a boundary value problem as in \citet{Galewsky2004}, perturbed with:
\begin{equation}
h^\prime_0(\phi, \theta) = \hat{h} \cos(\phi)\exp\!\left(-\left(\tfrac{\theta}{\alpha}\right)^2 \right) 
\Bigl[ \exp\!\left(-\left(\tfrac{\phi_2 - \phi}{\beta}\right)^2 \right) + \exp\!\left(-\left(\tfrac{\phi_2 + \phi}{\beta}\right)^2 \right) \Bigr],
\end{equation}
with constants $\phi_2 = \tfrac{\pi}{4}$, $\hat{h}=120$ m, $\alpha=\tfrac{1}{3}$, and $\beta=\tfrac{1}{15}$ \citep{Galewsky2004}. Simulations are run on a latitude–longitude grid of size $128 \times 256$.  

For data generation, $u_{max}$ is sampled from $\mathcal{U}(60,80)$. Trajectories span 320 hours with hourly snapshots, yielding 320 frames each. To emphasize meaningful dynamics, we retain only the last 160 frames. Each long trajectory is split into sub-trajectories of 40 frames. The resulting dataset comprises 64 training trajectories and 8 test trajectories. Finally, we rescale the fields for numerical stability: heights $h$ by $3 \times 10^3$ and vorticity $w$ by 2.

\begin{figure}[htbp]
        \centering
        \includegraphics[width=\textwidth]{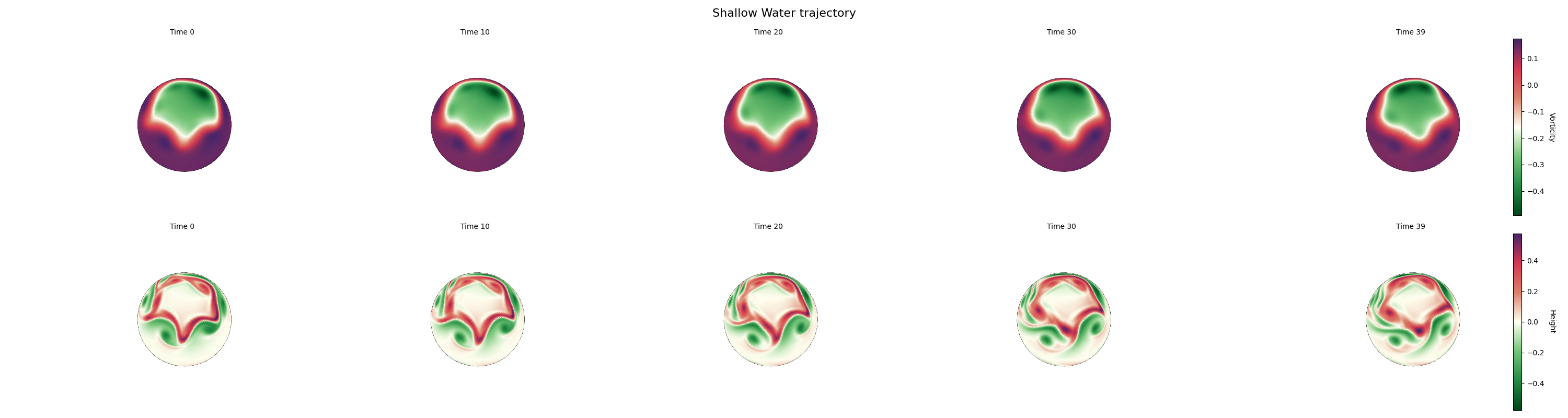}
        \caption{Sample of the Shallow-water dataset. }
  \label{fig:vis_sw}
\end{figure}

\subsection{Eagle}
\label{app:ssec_eagle}
\begin{figure}[htbp]
        \centering
        \includegraphics[width=\textwidth]{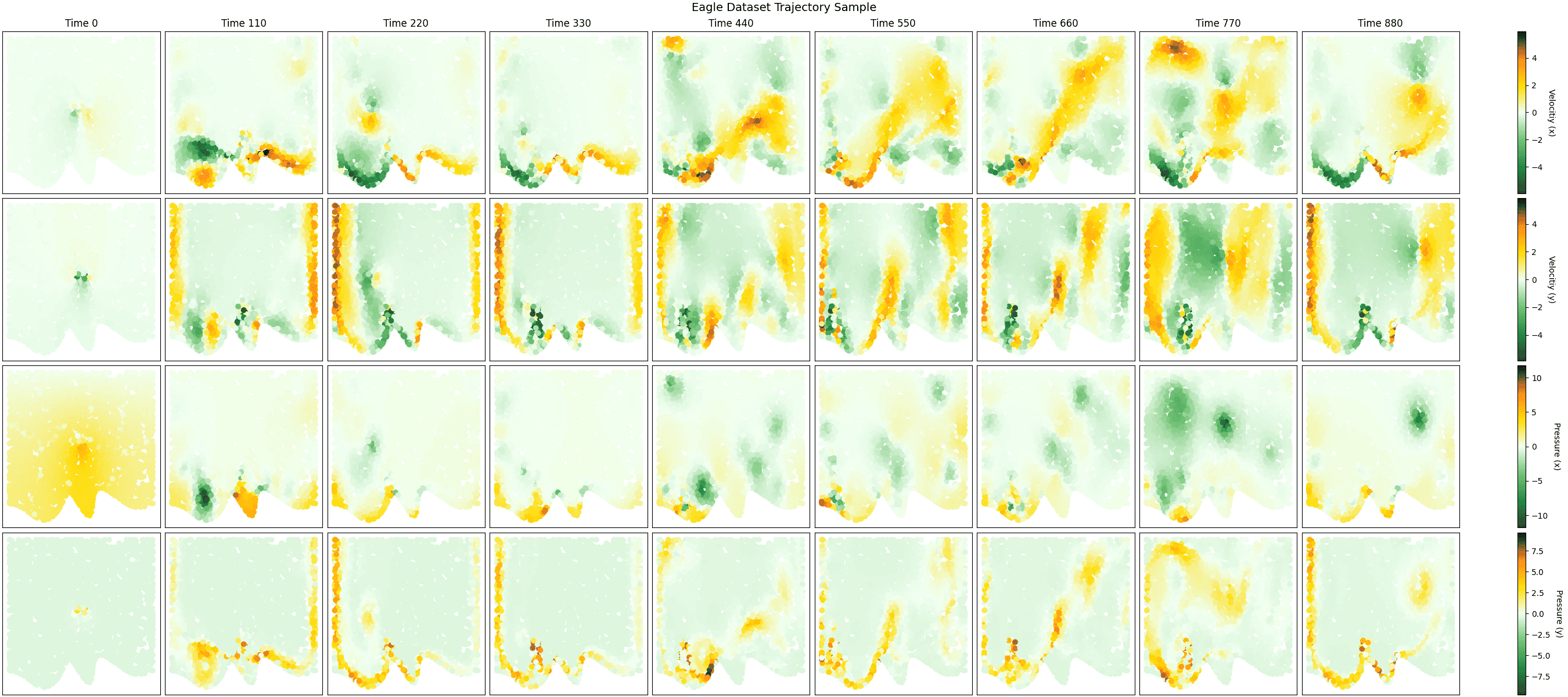}
        \caption{Sample of the Eagle dataset. }
        \label{fig:vis_eagle}
\end{figure}

Eagle is taken from \citep{janny2023eagle} and represents the fluid velocity and pressure around a moving source. This dataset has different boundary shapes that create varying flow around the source. Moreover, trajectories are long ($990$ timesteps), making this dataset well-suited for long-rollout tasks. For additional deatails on dataset generation, we refer readers to \citep{janny2023eagle}. We provide a visualization of a trajectory in \cref{fig:vis_eagle}

\subsection{Cylinder-Flow}
\label{app:ssec_cylinder}
We use the dataset introduced by \citet{pfaff2020}, which simulates water flow in a channel with a cylinder serving as an obstacle. The governing equations are the 2D incompressible Navier--Stokes equations with constant density:
\begin{align}
\partial_t \mathbf{v} &= 0, \\
\rho_0 \left( \partial_t \mathbf{v} + \mathbf{v} \cdot \nabla \mathbf{v} \right) + \nabla p &= \mu \nabla^2 \mathbf{v}, \\
\mathbf{v} &:= \mathbf{v}(t, \omega), \quad p := p(t, \omega), \quad \omega \in \Omega, \, t \in [0,T],
\end{align}
where $\rho_0$ denotes the constant density, $\mathbf{v}$ the velocity field, and $p$ the pressure. Each sample varies in the cylinder’s diameter and position, and the task of neural surrogates is to predict both velocity and pressure fields. 
For our experiments, we subsample trajectories to 15 timesteps. Further details about the setup can be found in \citet{pfaff2020}.

\begin{figure}[htbp]
        \centering
        \includegraphics[width=\textwidth]{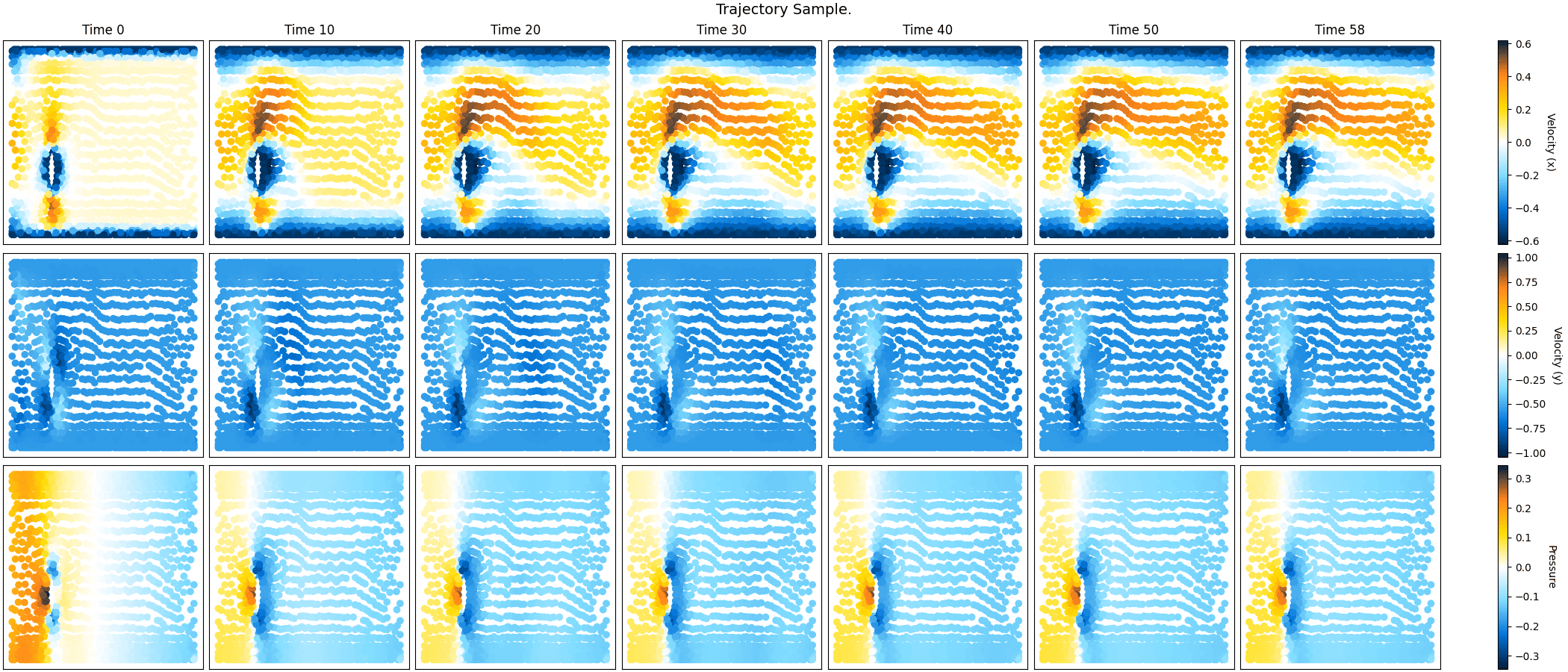}
        \caption{Sample of the Cylinder Flow dataset. }
        \label{fig:vis_cylinder}
\end{figure}

\subsection{Rayleigh-Bénard}
\label{app:ssec_rb}
We consider a 2D horizontally periodic fluid subject to buoyancy-driven convection. The dataset is taken from \citep{ohana2024well}, which provides the full setting for data generation. The state variables are the velocity $u = (u_x, u_z)$, buoyancy $b$, and pressure $p$.  
Heating from below and cooling from above create density variations that drive convection, producing characteristic Bénard cells where hot fluid rises and cold fluid sinks.  

\begin{figure}[htbp]
    \begin{subfigure}{0.48\textwidth} 
        \centering
        \includegraphics[width=\textwidth]{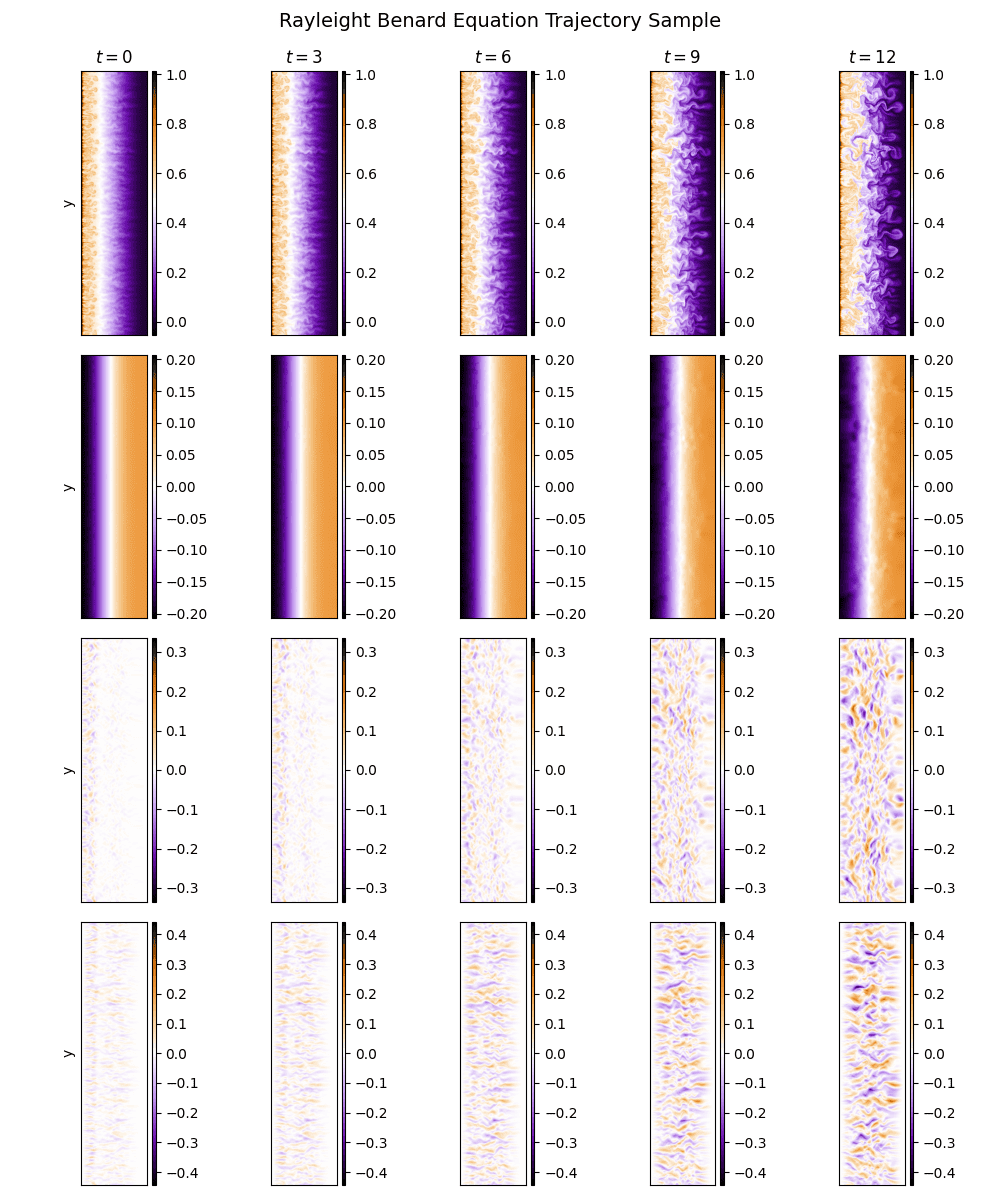}
        \caption{}
        \label{fig:vis_rb1}
  \end{subfigure}
  \hfill
  \begin{subfigure}{0.48\textwidth} 
        \centering
        \includegraphics[width=\textwidth]{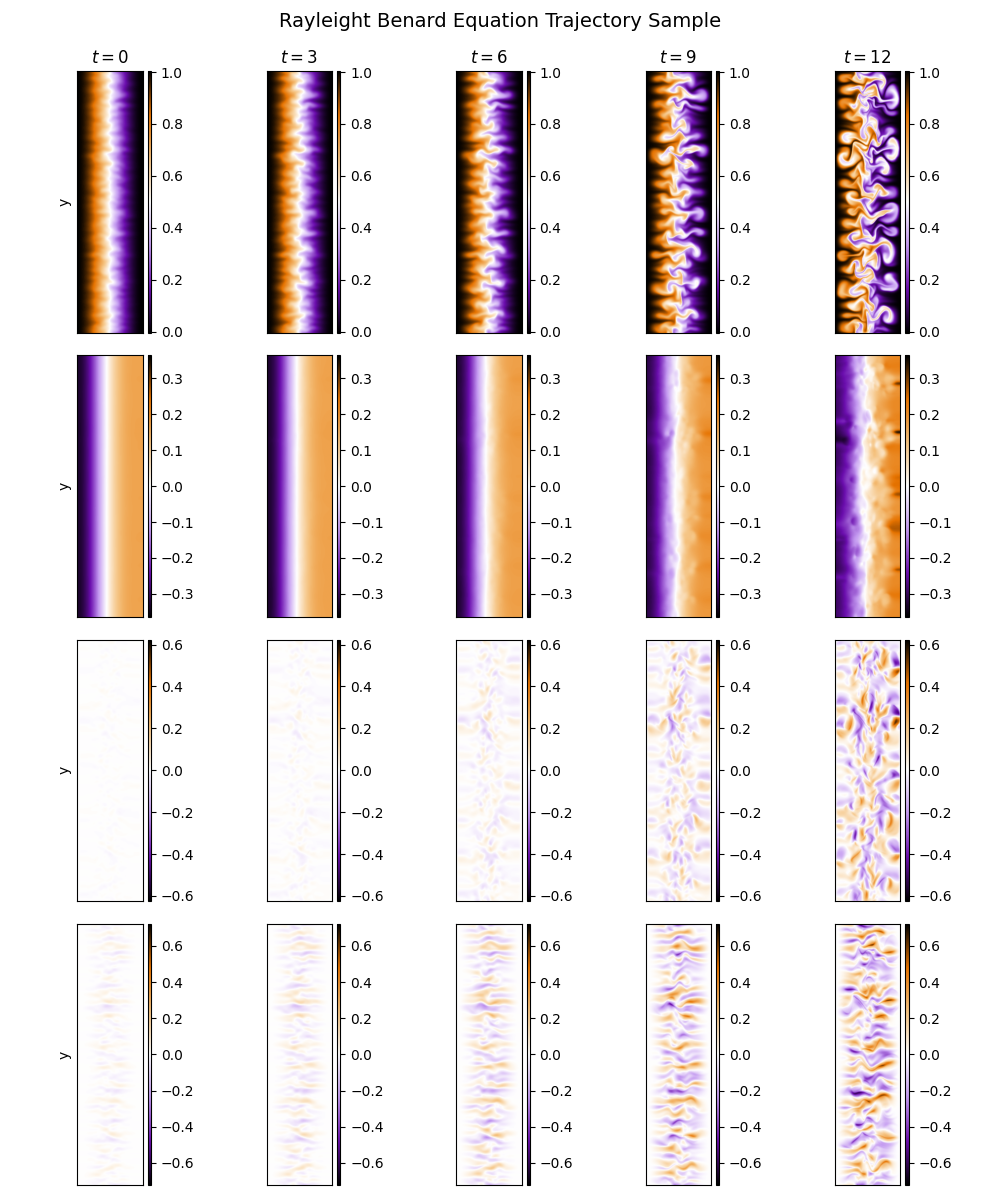}
        \caption{}
        \label{fig:vis_rb2}
  \end{subfigure}
  \caption{Samples from the \textit{Rayleigh-Bénard} dataset. }
  \label{fig:vis_rb}
\end{figure}

The dynamics are governed by
\begin{align}
\frac{\partial b}{\partial t} - \kappa \Delta b &= - u \cdot \nabla b, \\
\frac{\partial u}{\partial t} - \nu \Delta u + \nabla p - b e_z &= - u \cdot \nabla u,
\end{align}
with $\Delta = \nabla \cdot \nabla$ the Laplacian, $e_z$ the vertical unit vector, and $\int p = 0$ a gauge constraint.  
The first equation describes convection–diffusion of buoyancy, while the second is the Navier–Stokes equation with buoyancy forcing.  
The domain is periodic in $x$ with vertical boundary conditions $u(z=0)=u(z=L_z)=0$ and $b(z=0)=b(z=L_z)=0$. The system is parameterized by the Rayleigh and Prandtl numbers through thermal diffusivity $\kappa$ and viscosity $\nu$:
\begin{align}
\kappa &= (\text{Rayleigh} \times \text{Prandtl})^{-1/2}, \quad
\nu = \left(\frac{\text{Rayleigh}}{\text{Prandtl}}\right)^{-1/2}.
\end{align}
Here, the Rayleigh number measures the ratio of buoyancy to diffusion forces, while the Prandtl number controls the balance between momentum and thermal diffusion. We provide in \cref{fig:vis_rb} visualizations of $2$ trajectories. 

\subsection{Gray-Scott}
\label{app:ssec_gs}
The Gray--Scott system \citep{gs} is a pair of coupled reaction--diffusion equations modeling the interaction of two chemical species, $A$ and $B$, whose concentrations evolve in space and time:
\begin{align}
\frac{\partial A}{\partial t} &= \delta_A \Delta A - AB^2 + f(1 - A), \label{eq:grayscott1} \\
\frac{\partial B}{\partial t} &= \delta_B \Delta B + AB^2 - (f + k)B. \label{eq:grayscott2}
\end{align}

Here, $f$ and $k$ are the \emph{feed} and \emph{kill} rates: $f$ controls the supply of species $A$, while $k$ regulates the removal of species $B$.  
The diffusion coefficients $\delta_A$ and $\delta_B$ determine the transport rates of the two species.   

\begin{figure}[htbp]
    \centering
    \includegraphics[width=\linewidth]{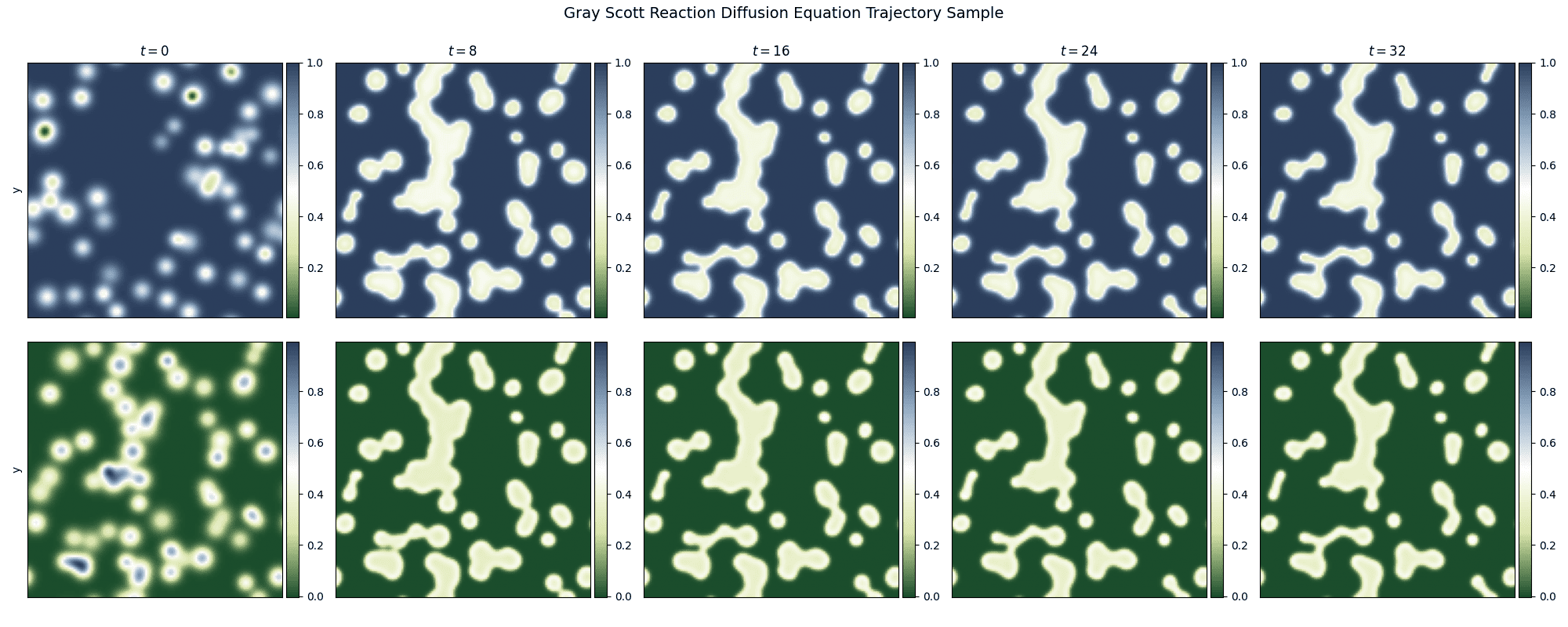}
    \caption{Sample of the \textit{Gray-Scott Reaction-Diffusion} dataset}
    \label{fig:gs_vis0}
\end{figure}

The dataset used in our experiments is provided by \citet{ohana2024well}, which includes further details on its generation process.See \cref{fig:gs_vis0} for a visualization of a trajectory. 

\subsection{Active Matter} 
\label{app:ssec_activematter}
\begin{figure}[htbp]
    \centering
    \includegraphics[height=0.8\textheight]{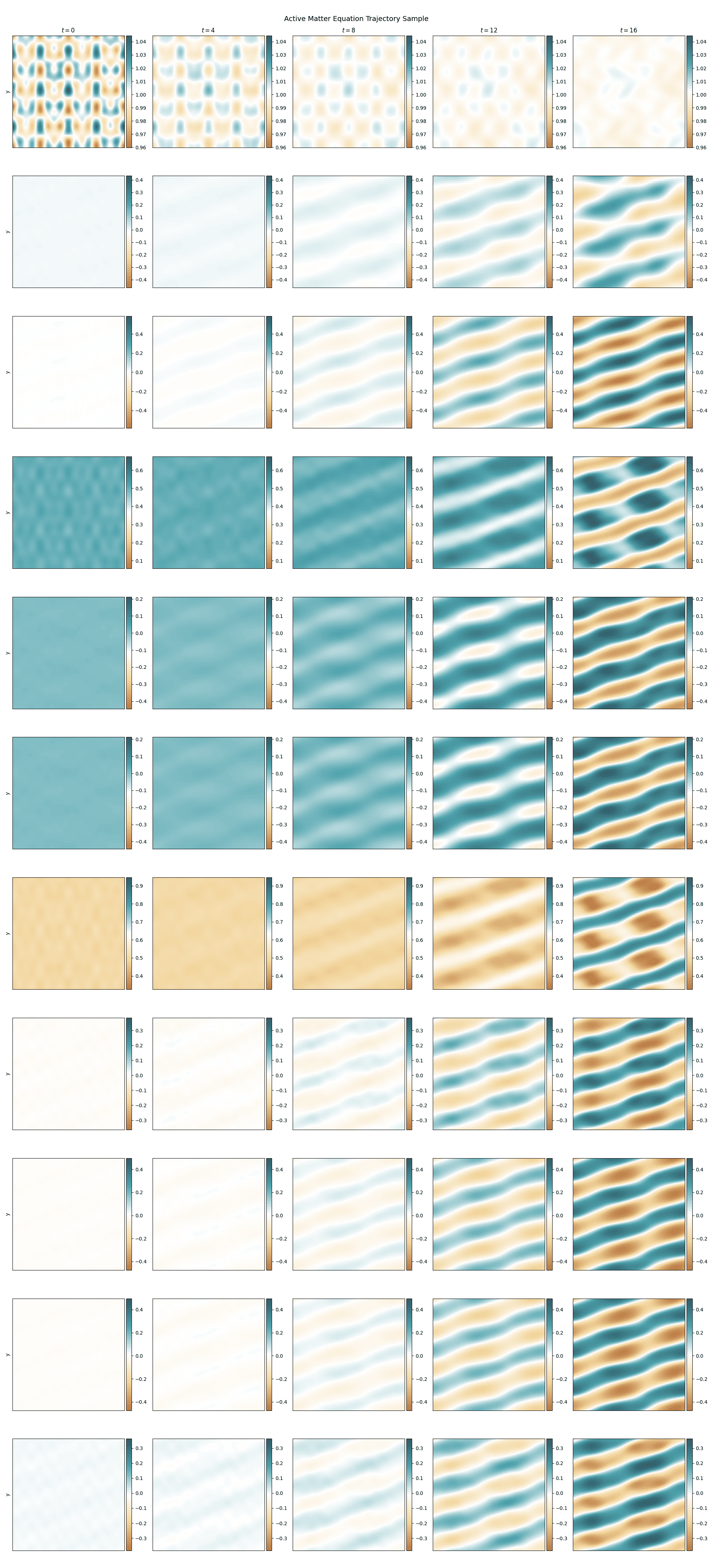}
    \caption{Trajectory sample of \textit{Active Matter} dataset. }
    \label{fig:active_matter_vis}
\end{figure}

Active Matter features simulations of rod-like biological active particles immersed in a Stokes flow. The active particles
transfer chemical energy into mechanical work, leading to stresses that are communicated across the system. Furthermore, particle coordination causes complex behavior inside the flow. More details are provided in \citep{ohana2024well} for the dataset generation setup.We provide a visualization of a trajectory in \cref{fig:active_matter_vis}. 

\clearpage

\subsection{Acoustic Scattering Maze}
\label{app:ssec_asm}
We use the \textit{Acoustic Scattering Maze} dataset introduced by \citet{ohana2024well}. 
The dynamics are governed by acoustic equations that describe the propagation of a pressure wave through materials with spatially varying density. 
The governing system is:
\begin{align}
\frac{\partial p}{\partial t} + K(x,y) \left( \frac{\partial u}{\partial x} + \frac{\partial v}{\partial y} \right) &= 0 \label{eq:acoustic1} \\
\frac{\partial u}{\partial t} + \frac{1}{\rho(x,y)} \frac{\partial p}{\partial x} &= 0 \label{eq:acoustic2} \\
\frac{\partial v}{\partial t} + \frac{1}{\rho(x,y)} \frac{\partial p}{\partial y} &= 0 \label{eq:acoustic3}
\end{align}
where $\rho$ denotes the material density, $u$ and $v$ the velocity components in the $x$ and $y$ directions, 
$p$ the pressure, and $K$ the bulk modulus. Together, $\rho$ and $K$ determine the speed of sound; 
in these simulations, $\rho$ varies spatially while $K$ is fixed to a constant value of $4$.

\begin{figure}[htbp]
    \centering
    \includegraphics[width=\linewidth]{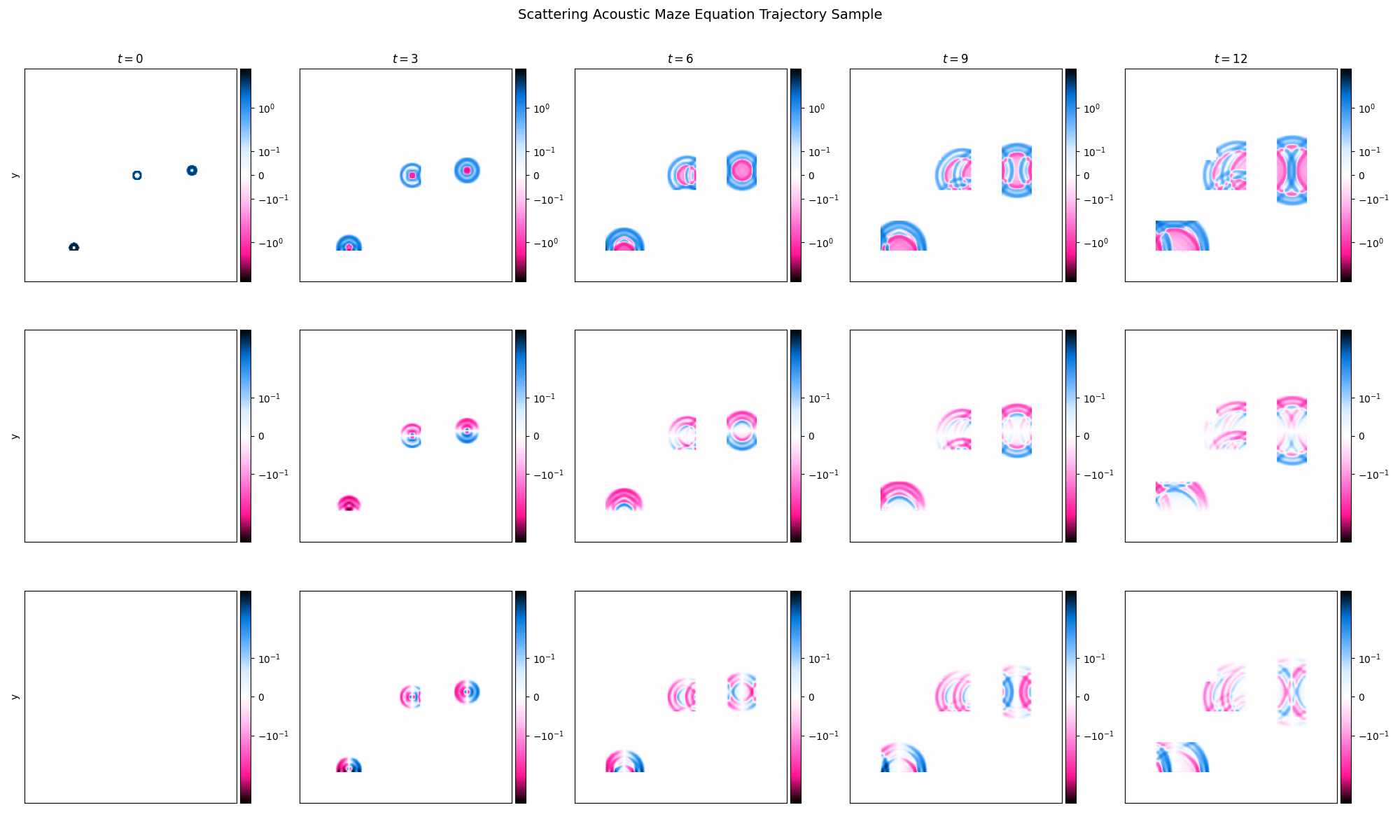}
    \caption{Sample of the \textit{Acoustic Scattering Maze} dataset. }
    \label{fig:scatteringmaze_vis}
\end{figure}

\subsection{Magnetohydrodynamic (MHD)}
\label{app:ssec_mhd}
\begin{figure}[htbp]
    \begin{subfigure}{0.33\textwidth}
        \centering
        \includegraphics[width=0.7\linewidth]{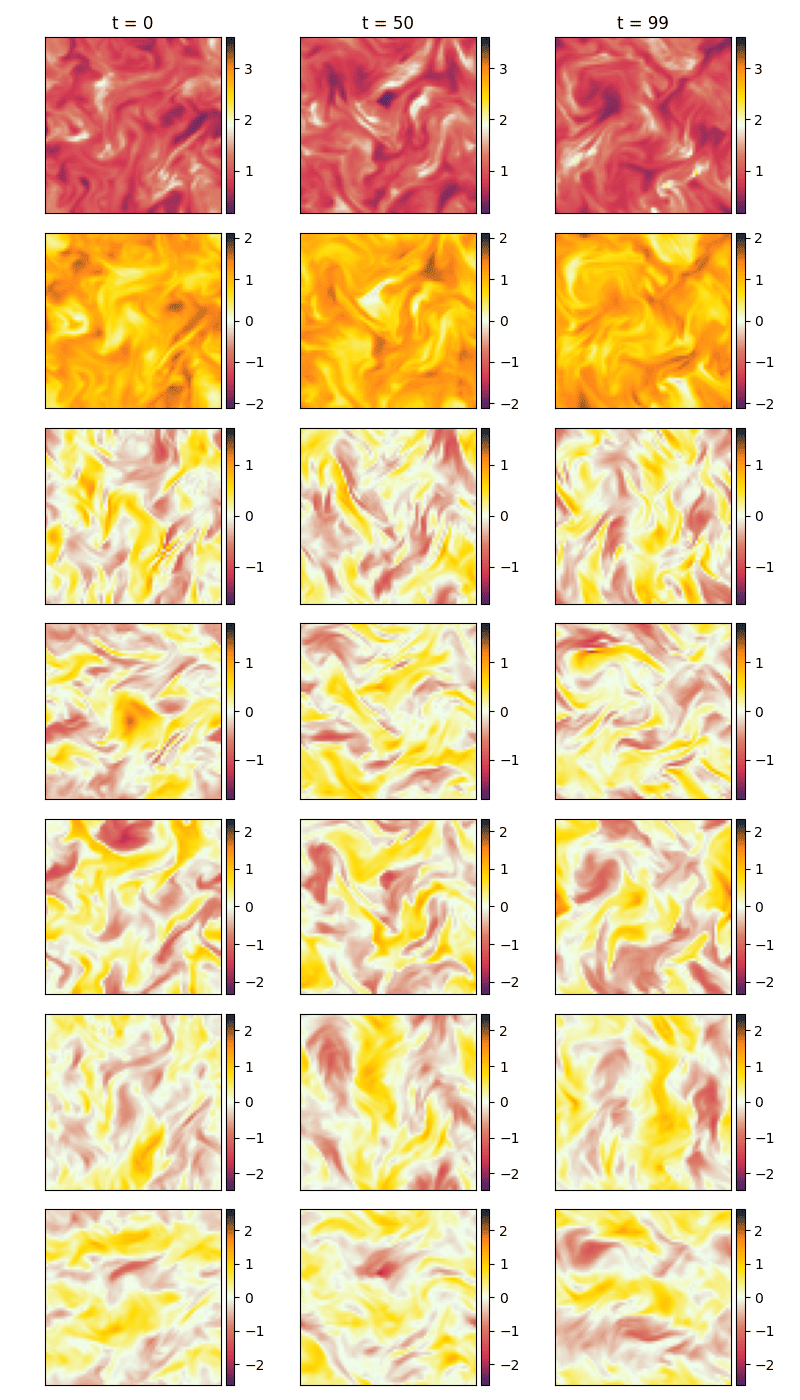}
        \caption{Slice along x.}
        \label{fig:mhd_vis_x}
    \end{subfigure}
    \begin{subfigure}{0.33\textwidth}
        \centering
        \includegraphics[width=0.7\linewidth]{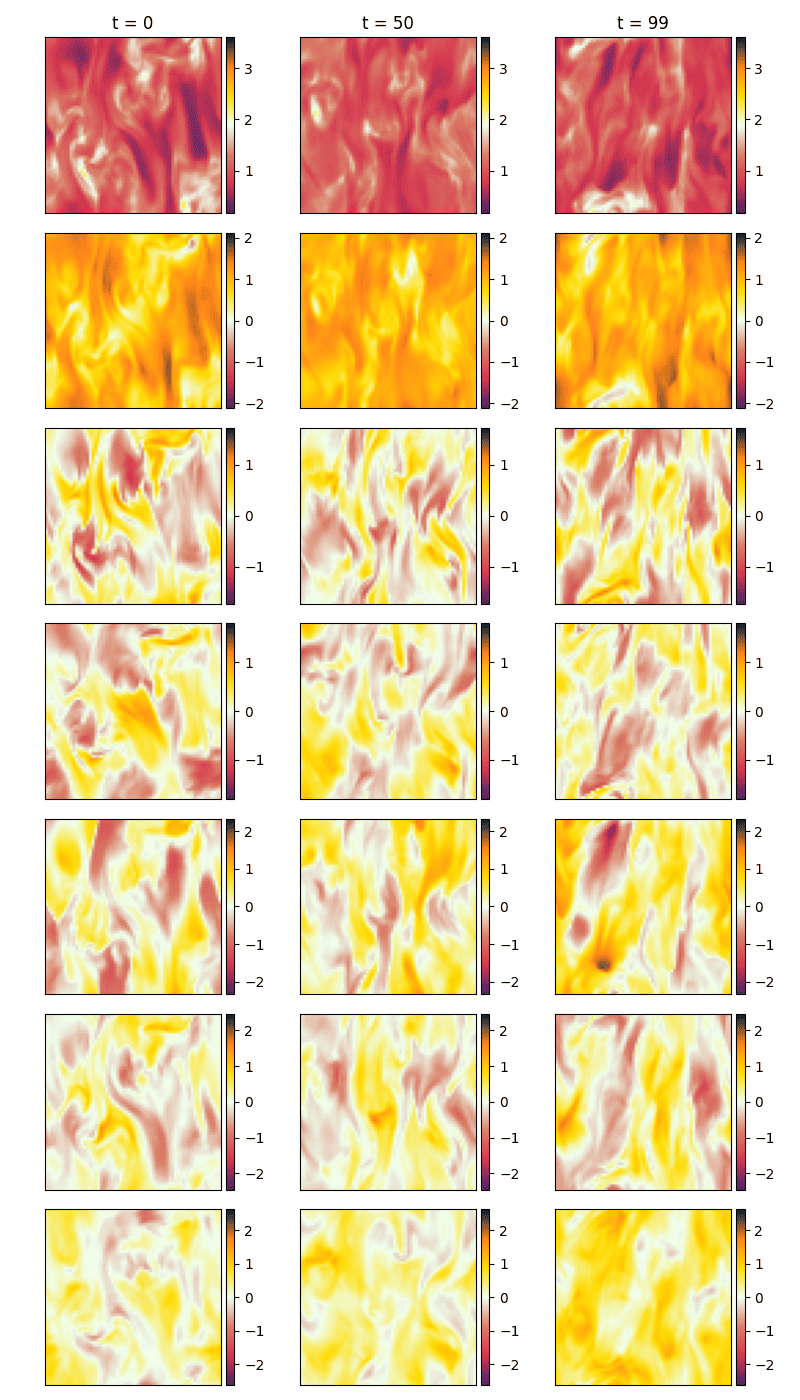}
        \caption{Slice along y.}
        \label{fig:mhd_vis_y}
    \end{subfigure}
    \begin{subfigure}{0.33\textwidth}
        \centering
        \includegraphics[width=0.7\linewidth]{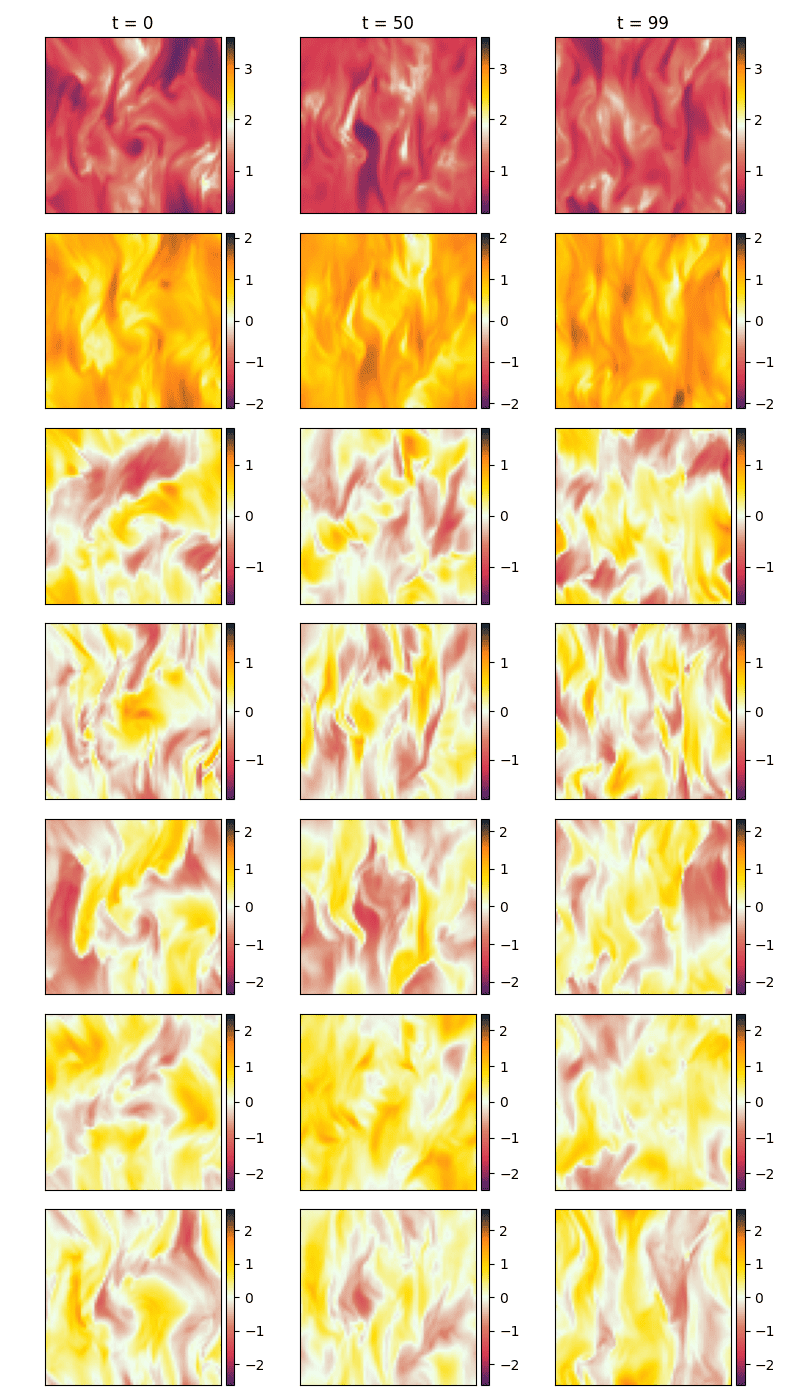}
        \caption{Slice along z.}
        \label{fig:mhd_vis_z}
    \end{subfigure}
    \caption{Trajectory sample of \textit{MHD} dataset sliced along each axis. }
    \label{fig:mhd_vis}
\end{figure}

The MHD dataset \citep{ohana2024well}, simulates a magnetohydrodynamic (MHD) turbulence. Such dynamics are often used to describe space events such as solar winds or galaxy formation. We refer to \citep{ohana2024well} for additional details. We provide in \cref{fig:mhd_vis} a trajectory sample. 

\vspace{10cm}
\subsection{Turbulence Gravity Cooling (TGC)}
\label{app:ssec_tgc}
\begin{figure}[htbp]
    \begin{subfigure}{0.33\textwidth}
        \centering
        \includegraphics[width=0.7\linewidth]{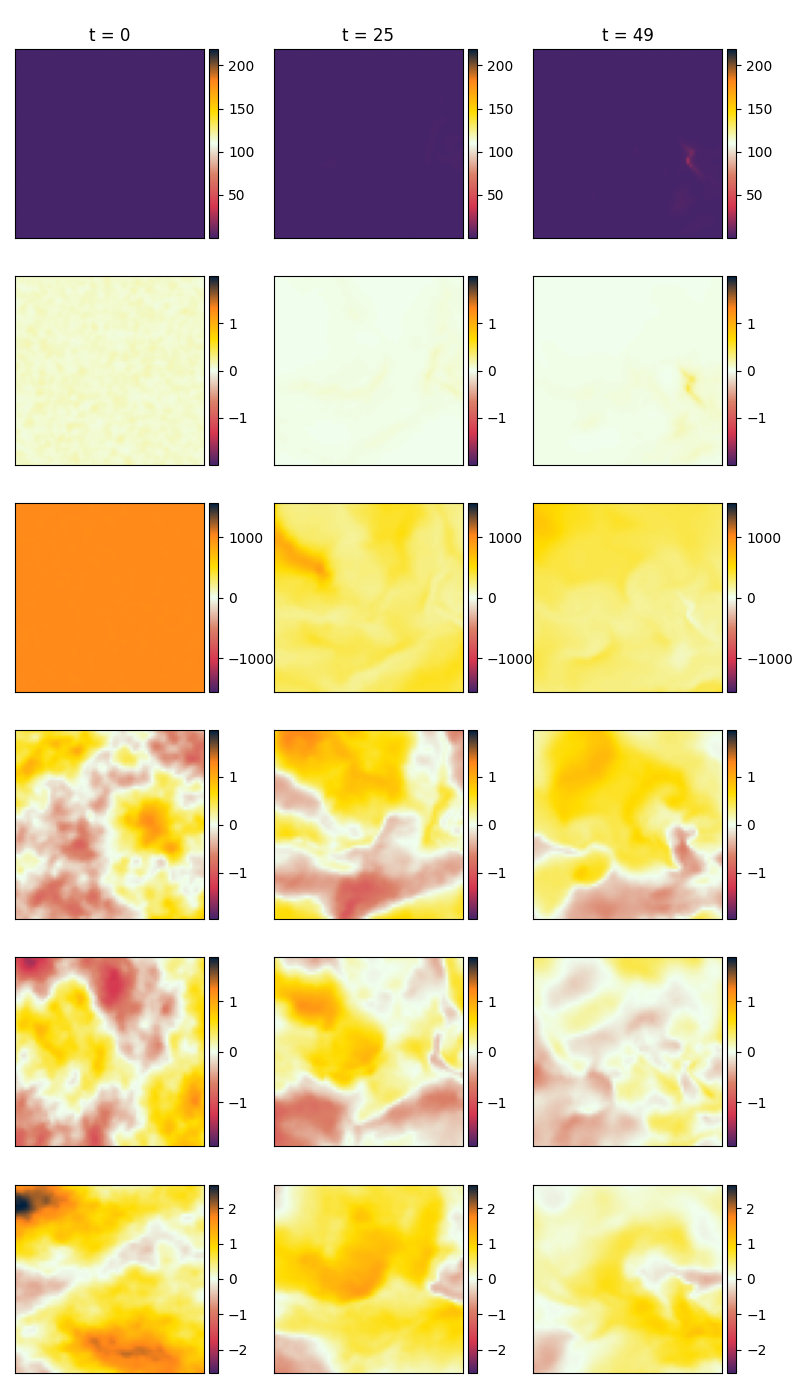}
        \caption{Slice along x.}
        \label{fig:tgc_vis_x}
    \end{subfigure}
    \begin{subfigure}{0.33\textwidth}
        \centering
        \includegraphics[width=0.7\linewidth]{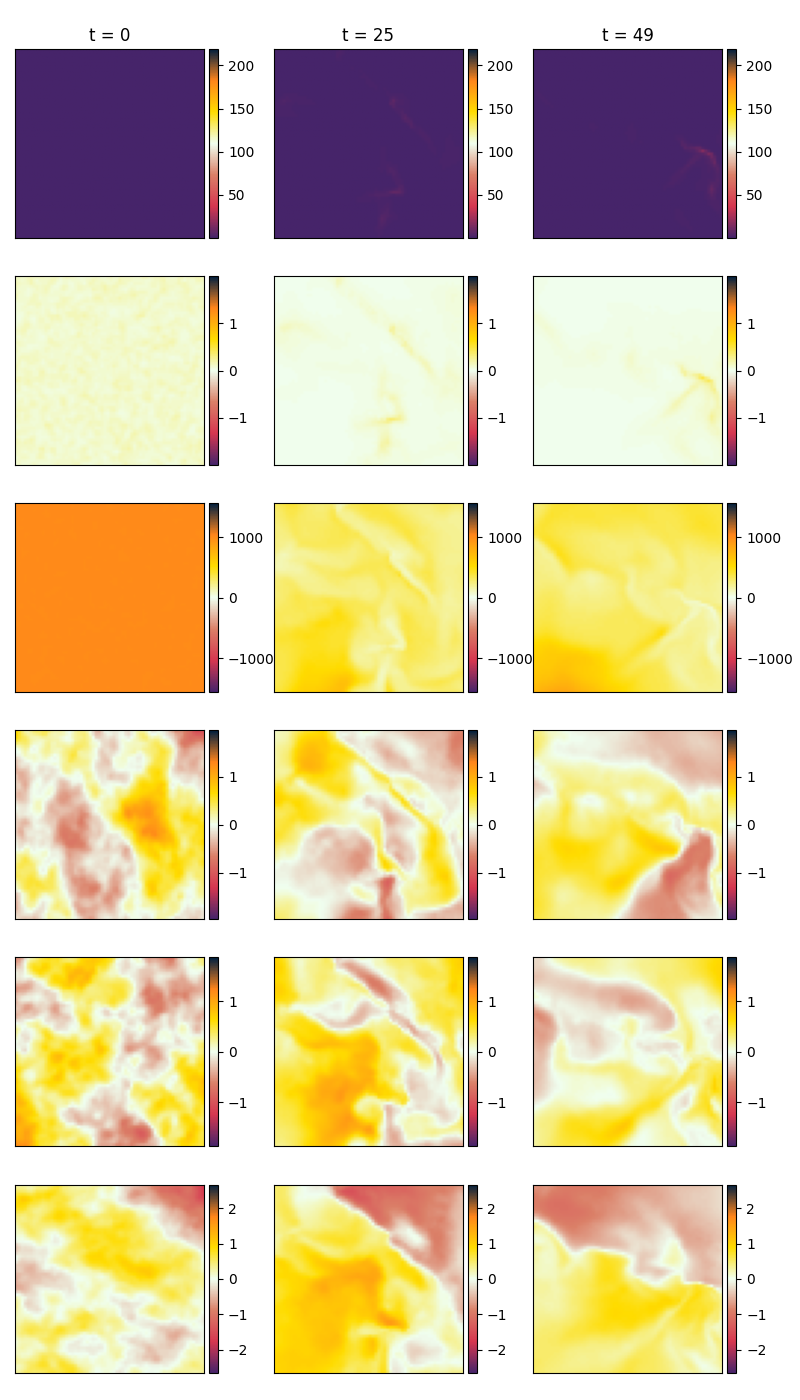}
        \caption{Slice along y.}
        \label{fig:tgc_vis_y}
    \end{subfigure}
    \begin{subfigure}{0.33\textwidth}
        \centering
        \includegraphics[width=0.7\linewidth]{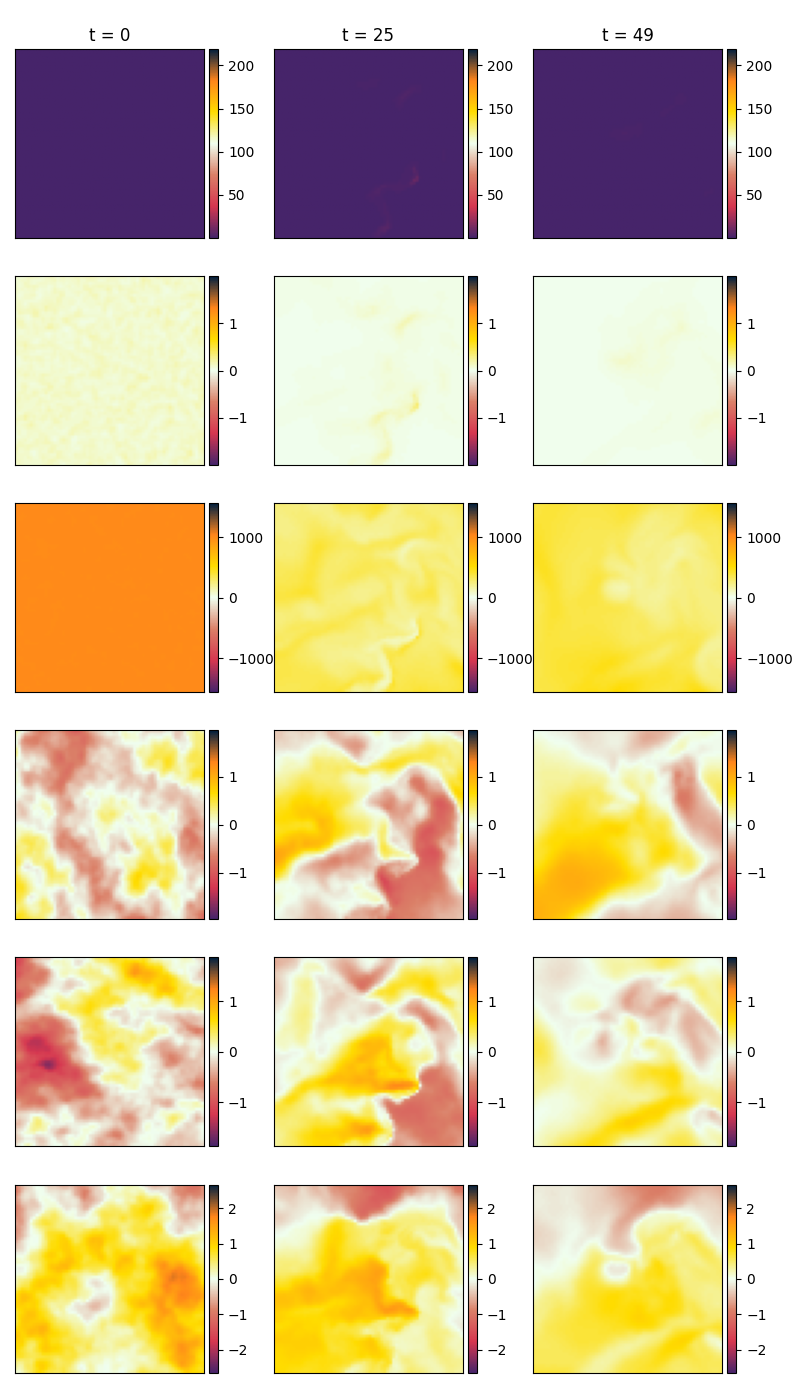}
        \caption{Slice along z.}
        \label{fig:tgc_vis_z}
    \end{subfigure}
    \caption{Trajectory sample of \textit{TGC} dataset sliced along each axis. }
    \label{fig:tgc_vis}
\end{figure}

The TGC dataset \citep{ohana2024well}, simulates a turbulent fluid with gravity. We provide in \cref{fig:tgc_vis} a trajectory sample. Additional details can be found in \citep{ohana2024well}. 

\clearpage
\section{Architecture details}
\label{app:arch}
We provide additional architecture details about ECHO's auto encoder in \cref{app:arch_ae} and generative model \cref{ssec:architecture}. 

\subsection{Inference model}
\label{app:Inference_instances}

We describe here several instance of the inference process corresponding to different tasks, that can be handled by ECHO. This is an illustration of the inference formalism introduces in section \ref{ssec:inference_model}.

\begin{itemize}
    \item \textbf{Forward prediction:} $\mathcal{O}$ contains the first $L$ frames of a trajectory, encoded as $\vz^{0:L'-1} = E_\phi(\vu^{0:L-1}) \in \mathbb{R}^{M \times L' \times d}$. The model predicts the missing tokens $\vz^{\mathcal{M}'} = \{\vz^{L'}, \dots, \vz^{T'}\}$, completing the latent trajectory.
    
    \item \textbf{Interpolation:} $\mathcal{O}$ consists of $L$ non-consecutive frames $\{\vu^{t_1}, \dots, \vu^{t_L}\}$ with $\{t_0, \dots, t_{L-1}\} \subset [0,T]$. Each observed frame is encoded spatially as $\vz^{t_\ell} = E_\phi(\vu^{t_\ell}) \in \mathbb{R}^{M \times d}$, yielding a set of latent observations $\vz^{\mathcal{O}'}$. The generative model then reconstructs the missing temporal dynamics by producing the full latent trajectory  $\vz^{0:T'}$, thus interpolating the unobserved tokens $\vz^{\mathcal{M}'} = \vz^{0:T'} \setminus \vz^{\mathcal{O}'}$. 
    
    \item \textbf{Backward prediction:} $\mathcal{O}$ contains the last frames of a trajectory $\{\vu^{t_{T-L+1}}, \dots, \vu^{t_T}\}$, encoded as $\vz^{\mathcal{O}'}$. The model reconstructs earlier dynamics by generating $\vz^{0:T'}$, recovering $\vz^{\mathcal{M}'}$.
    
    \item \textbf{Initial value problem:} $\mathcal{O}$ contains only $\vu^0$ (optionally with $\bm{\gamma}$), encoded as $\vz^0 \in \mathbb{R}^{M \times 1 \times d}$. The model then generates $\vz^{1:T'} = \vz^{\mathcal{M}'}$, yielding the full trajectory $\vz^{0:T'}$.
    
    \item \textbf{Conditional / unconditional generation:} $\mathcal{O} = \varnothing$. The model initializes $\vz^{\mathcal{M}'}_0 \sim \mathcal{N}(0,I) \in \mathbb{R}^{M \times T' \times d}$ and generates $\vz^{0:T'}$, either unconditionally or conditioned on PDE parameters $\bm{\gamma}$.
\end{itemize}

\paragraph{Spatial tasks}
ECHO's encoder–decoder architecture naturally supports a variety of spatial tasks. Thanks to the continuous convolution layers, the model can encode an arbitrary number of input points and decode at any desired spatial location. As a result, ECHO can perform spatial interpolation, super-resolution, and related tasks simply by querying the decoder at the appropriate target points.

\begin{itemize}
    \item \textbf{Spatial interpolation}: $\Xi$ denotes an incomplete irregular grid. During decoding, we query the decoder at the missing locations, effectively interpolating the field on the unobserved points.
    \item \textbf{Super-resolution}: During decoding, we query a finer grid that includes intermediate locations between the original points, thereby interpolating the input data at a higher spatial resolution.
\end{itemize}

We provide examples of super-resolution in \cref{tab:1024x1024} and spatial interpolation (reconstruction from irregular partial grids) in \cref{tab:exp_no,tab:exp_no_ood}.

\subsection{Encoder-Decoder}
\label{app:arch_ae}
\begin{figure}[htbp]
        \centering
        \includegraphics[width=\linewidth]{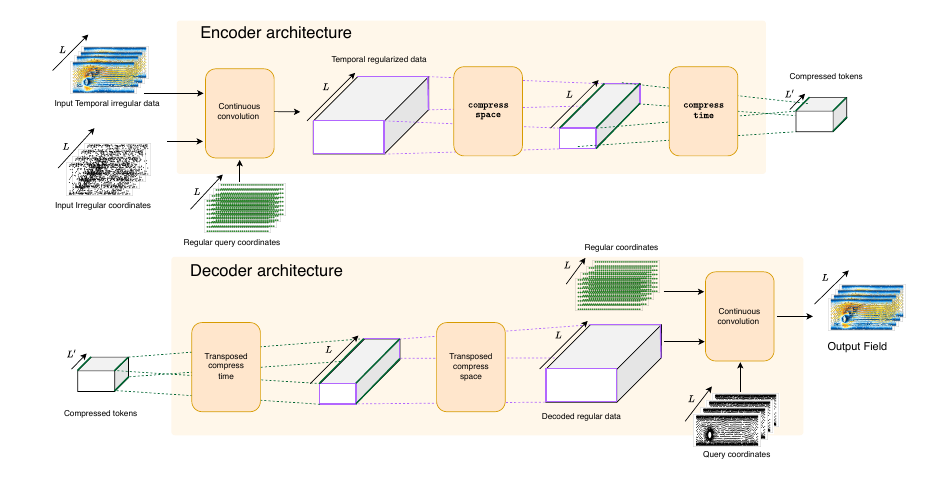}
        \caption{Encoder–decoder architecture of ECHO. The auto-encoder uses a continuous convolution layer to process irregular input grids with an arbitrary number of points, mapping the dynamics onto a regular latent grid. This latent representation is then compressed by a hierarchical CNN encoder. The decoder mirrors this process with a hierarchical CNN decoder followed by a continuous convolution layer, enabling queries at any desired output location.}
        \label{fig:eencdec-arch}
    \end{figure}

The encoder–decoder architecture consists of two stages. First, a continuous convolution layer regularizes the (potentially) irregular input grid (\cref{app:sssec-interplayer}). Second, a spatio-temporal CNN compresses the regularized representation to the desired token dimension (\cref{app:sssec-complayer}), as illustrated in \cref{fig:eencdec-arch}.
    
\subsubsection{Interpolation with continuous convolution}
\label{app:sssec-interplayer}
When encoding (possibly) irregular high-dimensional data, the computational burden falls on the encoder, which must process a large number of spatial points. The first layer plays a crucial role in reconstruction performance: applying a very high compression ratio leads to information loss, whereas avoiding compression results in significant computational cost (as detailed in \cref{sec:experiments}). In ECHO, we leverage continuous convolutions \citep{calmpde} to handle irregular grids. 

We now detail the continuous convolution layers used in the encoder and decoder.

\paragraph{Continuous convolution for the encoder.}
In the encoder, the inputs are:
(i) a physical field with $C_{\mathrm{in}}$ channels,
$f^{\mathcal{O}}(\mathcal{X})$, defined on coordinates
$\mathcal{X} = \{x_p\}_{p=1}^{|\mathcal{X}|}$ (with $|\mathcal{X}|$ the number
of spatial points), and
(ii) a regular latent grid $\Xi = \{\xi_j\}_{j=1}^{S}$, where $S$ is the number
of latent grid points.

With these notations, the continuous convolution at an output location
$\xi_j \in \Xi$ is given by
\begin{equation}
    (f^{\mathcal{O}} \ast k)_o(\xi_j)
    =
    \sum_{i=1}^{C_{\mathrm{in}}}
    \sum_{p \in \mathrm{RF}(\xi_j)}
    f_i(x_p)\, k_{i,o}(\xi_j - x_p),
\end{equation}
where $\mathrm{RF}(\xi_j)$ denotes the receptive field around $\xi_j$, and
$k_{i,o}$ is a kernel parameterized by a neural network. Each output on the
regular latent grid is thus obtained by convolving the input field with this
learned kernel. Importantly, there is no constraint on the number of input
points $|\mathcal{X}|$, so the encoder can handle arbitrary point clouds.

\paragraph{Continuous convolution for the decoder.}
The final decoder layer has a similar form. Let $f^{\mathrm{DEC}}$ be the
decoded latent representation with $d$ channels, defined on the regular grid
$\Xi$ (the same latent grid as in the encoder). Given a set of query
coordinates $\mathcal{X} = \{x_i\}$ in physical space, the reconstructed field
at a query point $x_i$ is computed as
\begin{equation}
    (f^{\mathrm{DEC}} \ast k)_o(x_i)
    =
    \sum_{c=1}^{d}
    \sum_{p \in \mathrm{RF}(x_i)}
    f^{\mathrm{DEC}}_{c}(\xi_p)\,
    k_{c,o}(x_i - \xi_p),
\end{equation}
where $k_{c,o}$ is again a learned kernel and $\mathrm{RF}(x_i)$ denotes the
receptive field around the query location $x_i$.

Because the convolution kernel $k$ is learned during training and the formulation
is defined for arbitrary query coordinates, the model can be evaluated at any
location $x_i$. This enables a range of spatial tasks such as spatial
interpolation and super-resolution, as demonstrated in
\cref{sec:experiments,app:add_exp}.


\paragraph{ECHO's architectural modifications}
In order to improve the scalability of this layer, we perform some architectural modifications. \begin{itemize}
    \item \textbf{Fixed latent grid}: The latent grid is fixed to a dense regular grid (up to 36k points for \textit{Vorticity}). This leads to a lower memory consumption at training.  This is also motivated by the use of CNN-based compression layers that needs to operate on regular grids and enables efficient and scalable encoding. 
    \item \textbf{Chunk}: We perform chunking for very dense grids, i.e. we compute the output locally by iterative on smaller amount of query points. 
\end{itemize} 

\subsubsection{Spatial and Temporal compression}
\label{app:sssec-complayer}

For the architecture of the CNN, we use a MagViT architecture (see for example \citet{yu2023magvit,enma} for additional details). 
In our case, the encoder CNN takes as input the output of the interpolation module detailed above, ie a regular representation of the input field. The regularized physical space tensor is then compressed through a stack of three building blocks: $\texttt{residual}$, $\texttt{compress\_space}$, and $\texttt{compress\_time}$. Finally, a last layer projects the representation back to the target token dimension. The decoder mirrors this compression pipeline. This type of architecture has been shown to be effective for spatio-temporal compression~\citep{serrano2024zebra,yu2023magvit,enma}. The three types of layers used in the compression module are detailed below (see \citep{enma}). 

\paragraph{$\texttt{residual}$ blocks:}
The $\texttt{residual}$ block processes the input while preserving its original shape. It consists of a causal convolution with kernel size $k$, followed by a linear layer and a Global Context layer adapted from \citet{cao2020globalcontextnetworks}. If the output dimensionality differs from the input’s, an additional convolution is used to project the spatial channels to the desired size.

\paragraph{$\texttt{compress\_space}$ blocks:}
The $\texttt{compress\_space}$ block reduces spatial resolution by a factor of $2^d$, i.e., each spatial dimension is downsampled by a factor of $2$ using a convolutional layer with stride $s = 2$. The kernel size $k$ and padding $p$ are set accordingly, with $p = k // 2$. To ensure that only spatial dimensions are compressed, inputs are reshaped so that this operation does not affect the temporal axis.

\paragraph{$\texttt{compress\_time}$ blocks:}
The $\texttt{compress\_time}$ block performs temporal compression similarly to the spatial case, but operates along the time dimension. To preserve causality, padding is applied only to the past, with size $p = k - 1$, so that a frame at time $t$ attends only to frames at times $< t$. A convolution with stride $s = 2$ is used to reduce the temporal resolution by a factor of $2$.

\paragraph{ECHO's architectural choices} In our experiments, we compress physical fields by separating spatial and temporal compression. We start by stacking several $\texttt{compress\_space}$ blocks to reduce the spatial dimensions of the tokens. Then, we process it using a $\texttt{residual}$ layer. Since these layers, make use of 3D causal convolution, they allow to process jointly the spatial and temporal dimensions. Finally, we compress the temporal dimensions with $\texttt{compress\_time}$ layers. This gives us an encoded representation of the input physics. The detailed number of layers and hyper-parameters related to the encoder/decoder modules are shown in \cref{tab:dania-hp}. 

\clearpage
\section{Implementation details}
\label{app:imp_details}
The code has been written in Pytorch \citep{torch}. All experiments were conducted on a H100. We estimate the total compute budget—including development and evaluation—to be approximately 1000 GPU-days.

\subsection{Evaluation protocol of the process}
\label{app:ssec_imp_process}

\subsubsection{Baseline details}
\label{app:sssec_baselinear}
We detail the baseline architectures used to evaluate ECHO in both forward and backward prediction, corresponding to experiments in \cref{ssec:dyn_forecasting_exp}.  
All baselines follow a comparable training protocol (see \cref{app:sssec_trainar}).  
For forward and backward experiments, we assume $4$ known timesteps and autoregressively predict the remainder, except for ECHO (and its deterministic variant), which generates the full trajectory in a single forward pass.

\paragraph{FNO.}  
For the Fourier Neural Operator (FNO) \citep{Li2020fno}, we followed the authors’ recommendations and concatenated temporal history directly to the input channels. We used $10$ Fourier modes in both 1D and 2D, a channel width of $128$, and stacked $4$ spectral layers.

\paragraph{BCAT.}  
BCAT \citep{liu2025bcat} is a deterministic block-wise causal transformer for spatio-temporal dynamics, originally designed for multi-physics problems. We adapt it to parametric PDEs. BCAT performs autoregression in physical space and uses spatial patches to reduce token count, similar to Vision Transformers \citep{Dosovitskiy2020}.

\begin{table}[htbp]
\centering
\caption{BCAT hyperparameters used across all datasets.}
\label{tab:hp-bcat}
\small
\renewcommand{\arraystretch}{1.15}
\begin{tabular}{lc}
\toprule
Hyperparameter       & Value \\
\midrule
Patch size           & 8 \\
Transformer depth    & 6 \\
Hidden size          & 512 \\
MLP ratio            & 2 \\
Number of heads      & 8 \\
QK normalization     & True \\
Normalization type   & RMS \\
Activation           & SwiGLU \\
Positional embedding & Sinusoidal \\
\bottomrule
\end{tabular}
\end{table}

\paragraph{AViT.}  
The Axial Vision Transformer (AViT) \citep{mccabe2023multiple} applies attention separately along spatial and temporal dimensions.  
We use the same configuration as BCAT (Table~\ref{tab:hp-bcat}), which we found to perform best on our PDE datasets.

\paragraph{Transolver++.}  
Transolver++ \citep{luotransolver++} extends \citep{Tran2023ffno} for dense input grid, with improved parameterization and efficiency. We followed the recommended setup summarized below, provided in the reference paper.
\begin{table}[htbp]
\centering
\caption{Transolver++ hyperparameters used across all datasets.}
\label{tab:hp-transolver}
\small
\renewcommand{\arraystretch}{1.15}
\begin{tabular}{lc}
\toprule
Hyperparameter    & Value \\
\midrule
Number of layers  & 8 \\
Hidden size       & 128 \\
Attention heads   & 8 \\
MLP ratio         & 4 \\
Dropout           & 0 \\
Physics slices    & 32 \\
Activation        & GELU \\
\bottomrule
\end{tabular}
\end{table}

\paragraph{ENMA.}  
For ENMA \citep{enma}, we implemented the generation architecture given in the original paper. We followed the authors recommenation for the hyper-parameters. Its configuration across datasets is summarized in \cref{tab:hp-enma}.

\begin{table}[htbp]
\centering
\caption{ENMA generation hyperparameters.}
\label{tab:hp-enma}
\small
\renewcommand{\arraystretch}{1.15}
\begin{tabular}{lcccc}
\toprule
Hyperparameter            & Active Matter & Rayleigh--B\'enard & Gray--Scott & ASM \\
\midrule
VAE embedding dim         & 4          & 4          & 4          & 8 \\
Number of tokens          & 16         & 16         & 64         & 256 \\
Patch size                & 2          & 2          & 1          & 2 \\
Spatial Transformer depth & 8          & 8          & 8          & 8 \\
Causal Transformer depth  & 8          & 8          & 8          & 8 \\
Hidden size               & 512        & 512        & 512        & 512 \\
MLP ratio                 & 2          & 2          & 2          & 2 \\
Attention heads           & 8          & 8          & 8          & 8 \\
Dropout                   & 0          & 0          & 0          & 0 \\
QK normalization          & True       & True       & True       & True \\
Normalization type        & RMS        & RMS        & RMS        & RMS \\
Activation                & SwiGLU     & SwiGLU     & SwiGLU     & SwiGLU \\
Positional embedding      & Sinusoidal & Sinusoidal & Sinusoidal & Sinusoidal \\
FM steps                  & 10         & 10         & 10         & 10 \\
\bottomrule
\end{tabular}
\end{table}

\subsubsection{Training details}
\label{app:sssec_trainar}

All models were trained under the same protocol for fair comparison. Unless specified, we used the AdamW optimizer with $\beta_1=0.9$, $\beta_2=0.95$, a cosine learning rate schedule from $10^{-3}$ to $10^{-5}$, and linear warm-up over the first 2000 steps. Baselines were trained with the same schedule and selection criterion as ECHO, using the best checkpoint on train error.
\begin{table}[htbp]
\centering
\caption{Training hyperparameters used across datasets.}
\label{tab:hp-training}
\small
\renewcommand{\arraystretch}{1.15}
\begin{tabular}{lc}
\toprule
Hyperparameter            & Value \\
\midrule
Epochs                    & 200 \\
Batch size                & 32 \\
Learning rate             & $10^{-3}$ (cosine decay) \\
Weight decay              & $10^{-4}$ \\
Grad clip norm            & 1 \\
Betas $(\beta_1,\beta_2)$ & (0.9, 0.95) \\
\bottomrule
\end{tabular}
\end{table}

\subsubsection{Fairness of Baseline Comparisons and Scaling Limits.}
\label{app:ssec:fairness}
To ensure a rigorous and fair comparison, we carefully controlled for parameter counts, memory tuning, and training budgets across all baseline models. 
For memory-intensive baselines, we carefully searched for the best tradeoff between memory consumption and performance, to fit them within the 80GB memory limit of our H100 GPUs. 
Despite these tuning efforts, models relying on dense spatial attention (e.g., AROMA) or direct physical-space processing inherently scale poorly with grid resolution. 
The out-of-memory (OOM) failures reported for baselines on \cref{tab:1024x1024} are therefore not due to a lack of optimization, but rather fundamental architectural bottlenecks that ECHO effectively circumvents via its joint spatio-temporal compression. Additionally, this experiment corresponds to a super-resolution task on the Vorticity dataset, because the grid size considered during training was limited to $512\times 512$. 

On the other hand, \cref{fig:memory_scaling} shows a separate experiment. The setting is the following: we instantiate models with a fixed latent size, except for Transolver ++, for which we considered the setting of the authors \citep{luotransolver++}. 
Then we generate random grid, with growing size, that we fed into the networks (input and query size are similar). 
The objective here is to find the limits of the models in terms of memory consumption, rather than evaluation their capacities on the super-resolution task. 

\paragraph{Model sizes.}
\Cref{tab:params_full} reports the number of parameters of the models in \cref{tab:model-comparison}. ECHO is smaller than FNO, AViT and ENMA, so its accuracy cannot be attributed to a larger parameter count. For Transolver++, larger configurations exceeded GPU memory.

\begin{table}[htbp]
\centering
\caption{Number of parameters of the models compared in \cref{tab:model-comparison}.}
\label{tab:params_full}
\small
\renewcommand{\arraystretch}{1.15}
\begin{tabular}{lcccccc}
\toprule
Dataset            & Transolver++ & FNO & AViT & BCAT & ENMA & ECHO \\
\midrule
Rayleigh--B\'enard & 5.5M         & 75M & 51M  & 27M  & 72M  & 48M \\
Active Matter      & 5.5M         & 75M & 53M  & 28M  & 72M  & 48M \\
\bottomrule
\end{tabular}
\end{table}

\paragraph{Cause of the out-of-memory failures.}
At moderate resolution (\cref{tab:exp_no}), the cost of encoding the raw mesh is paid by shrinking the models to fit the memory budget. Attention-based encoders must use a smaller latent size on \emph{Vorticity} and \emph{Shallow-Water} and compress the mesh in a single stage. The first layer of CALM-PDE is more memory-intensive than ours, so it admits fewer latent points and therefore fewer hierarchical blocks. At higher resolution (\cref{tab:1024x1024}), this cost can no longer be paid and every baseline runs out of memory, including those without attention (GINO, CORAL, CALM-PDE).

\clearpage
\subsection{Encoder-Decoder implementation protocol}
\label{app:ssec_encedecimp}

\subsubsection{ECHO encoder-decoder's Architecture hyper-parameters}
\label{app:sssec_archencdec}
We present in \cref{tab:dania-hp} the hyper-parameters used for the experiments done in \cref{ssec:no_exp} for our encoder-decoder module.
\begin{table}[htbp]
\centering
\caption{Hyper-parameters of the encoder and decoder components of ECHO.}
\label{tab:dania-hp}
\small
\renewcommand{\arraystretch}{1.15}
\setlength{\tabcolsep}{6pt}
\begin{tabular}{lcccc}
\toprule
Parameter                       & Vorticity        & Shallow-Water        & Eagle          & Cylinder Flow \\
\midrule
Token dimension                 & 32               & 32                   & 32             & 32 \\
\midrule
\multicolumn{5}{l}{\textit{Interpolation module}} \\
\addlinespace[2pt]
Regularized grid                & $192\times192$   & $64\times128\times1$ & $64\times32$   & $64\times16$ \\
Hidden dimension                & 64               & 64                   & 64             & 64 \\
Receptive field                 & 0.01             & 0.01                 & 0.01           & 0.01 \\
Softmax temperature             & 1                & 1                    & 1              & 1 \\
Chunk size                      & 4096             & 4096                 & 4096           & 4096 \\
Max neighbors                   & 512              & 512                  & 512            & 16 \\
Latent grid type                & Euclidean        & Spherical            & Euclidean      & Euclidean \\
Spatial dimension               & 2                & 2                    & 3              & 2 \\
\midrule
\multicolumn{5}{l}{\textit{Compression module}} \\
\addlinespace[2pt]
Spatial compression layers      & 3                & 2                    & 2              & 2 \\
Temporal compression layers     & 1                & 2                    & 2              & 2 \\
Residual layers                 & \multicolumn{4}{c}{One between each compression layer} \\
Compression kernel size         & 3                & 3                    & 3              & 3 \\
Causal input layer kernel size  & 3                & 3                    & 3              & 3 \\
Causal output layer kernel size & 3                & 3                    & 3              & 3 \\
\bottomrule
\end{tabular}
\end{table}

\subsubsection{Baseline details}
\label{app:ssec_basdetails_encdec}
We detail here the hyper-parameters used for the baselines considered in \cref{ssec:no_exp}. If not explicitly stated, we follow the authors recommendation in the respective papers. 

In \cref{tab:token-sizes}, we report the shapes of the trajectory in the physical space and in the latent space for each baseline. GINO, CORAL, AROMA, ENMA and CALM-PDE only compress information spatially. In ECHO, both temporal and spatial space can be compressed.

\paragraph{Matching latent size rather than parameter count.}
Encoder--decoders are matched on compression rate, that is, on the size of the latent representation, rather than on parameter count. For each baseline, we follow the authors' recommendations and target a comparable latent size, so that all methods compress the input to approximately the same information budget; for instance, on \emph{Vorticity}, GINO uses a latent representation of $10\times576\times32=184{,}320$ values. The exceptions are ECHO, whose temporal compression allows a smaller latent size, and AROMA, whose latent size is constrained by memory. For hierarchical methods such as CALM-PDE, we stack as many compression blocks as needed to reach this size. For Perceiver-style encoders such as AROMA and ENMA, enlarging the encoder beyond this point has limited effect, since the single cross-attention projection has already discarded much of the input signal. \Cref{ssec:ablation_studies} provides the controlled version of this observation: at a fixed final latent size, adding intermediate compression stages improves reconstruction, an effect that parameter count does not capture.

\Cref{tab:params_encoders} reports the resulting encoder sizes. AROMA, ENMA and CALM-PDE do not fit in GPU memory in their published configurations (colored cells in \cref{tab:exp_no}); the counts correspond to the reduced configurations used in our experiments. Model size plays a minor role in scalability: AROMA and CALM-PDE use encoders two orders of magnitude smaller than that of ECHO and still exceed memory on the $1024\times1024$ grid, because their cost grows with the number of input points (\cref{fig:memory_scaling}). All methods use a batch size of $1$ at inference and the same temporal window.

\begin{table}[htbp]
\centering
\caption{Number of parameters of the encoder--decoders compared in \cref{tab:exp_no}.}
\label{tab:params_encoders}
\small
\renewcommand{\arraystretch}{1.15}
\begin{tabular}{lcccccc}
\toprule
Dataset       & GINO & CORAL & AROMA & ENMA & CALM-PDE & ECHO \\
\midrule
Vorticity     & 18M  & 23M   & 0.5M  & 1.9M & 0.5M     & 33M \\
Shallow-Water & 19M  & 21M   & 0.5M  & 5M   & 0.2M     & 30M \\
\bottomrule
\end{tabular}
\end{table}

\begin{table}[htbp]
\centering
\caption{Physical and latent space shapes in the experiments of \cref{tab:exp_no},
in the format temporal size $\times$ spatial size $\times$ token dimension. The
first row gives the shape of the trajectories. On datasets with many input
points, some baselines use a smaller latent shape (in \textcolor{red}{red}) to
fit in GPU memory.}
\label{tab:token-sizes}
\footnotesize
\renewcommand{\arraystretch}{1.15}
\setlength{\tabcolsep}{4pt}
\begin{tabular}{lcccc}
\toprule
Model             & Vorticity & Shallow-Water & Eagle & Cylinder Flow \\
\midrule
\textit{Trajectory} & $10\times(512\times512)\times1$ & $10\times(128\times256)\times2$ & $200\times3000\times3$ & $60\times2000\times3$ \\
\midrule
GINO     & $10\times576\times32$ & $40\times512\times32$ & $100\times512\times32$ & $60\times64\times32$ \\
CORAL    & $10\times18432$ & $40\times16384$ & $200\times4096$ & $50\times2048$ \\
AROMA    & \textcolor{red}{$10\times64\times32$} & \textcolor{red}{$40\times128\times32$} & $200\times128\times32$ & $60\times64\times32$ \\
CALM-PDE & $10\times576\times32$ & $40\times512\times32$ & $200\times128\times32$ & $60\times64\times32$ \\
ECHO     & $5\times24\times24\times32$ & $10\times16\times8\times32$ & $50\times16\times8\times32$ & $15\times16\times4\times32$ \\
\bottomrule
\end{tabular}
\end{table}

\paragraph{GINO}
In GINO, a graph operator is proposed relying on a GNO block to encode mesh points into a latent space. The resulting latents are then processed by $8$ FNO layers to increase model expressivity. We use the default hyperparameters provided in the \texttt{neuralop} package \citep{kossaifi2025librarylearningneuraloperators, Kovachki2022}.

\paragraph{CORAL}
CORAL is a operator learning framework relying on cordinate-based networks for solving PDEs on general geometries. The INRs are learned via a second-order CAVIA-style meta-learning scheme that optimizes shared parameters so that, for each new function, only a few inner-loop gradient steps are required to obtain an accurate per-sample latent code from sparse observations. We use $3$ inner-loop steps with an inner learning rate of $10^{-2}$. The INR has $6$ layers of width $256$, and the hypernetwork has $1$ hidden layer with width $256$. We refer to \citep{serrano2023} for additional details.

\paragraph{AROMA}
The AROMA baseline uses cross-attention to directly compress trajectories into a latent space. We follow the configuration proposed in \citep{serrano2024aroma} and set the number of latent tokens as detailed in \cref{tab:token-sizes}. Due to GPU memory limitations, we use fewer tokens than the other baselines on two datasets (Vorticity and Shallow-Water).

\paragraph{CALM-PDE}
For CALM-PDE, we follow the configuration from \citep{calmpde}, using $2$ or $3$ hierarchical convolutional layers where the spatial resolution is gradually reduced while the number of channels increases. For the first layer, which processes irregular grids, we select the largest latent grid that fits in GPU memory, and then apply a sequence of $\times 2$ downsampling steps until we reach the desired token resolution. As reported in \cref{tab:exp_no}, GPU memory constraints forced us to use a smaller first-layer latent grid than ECHO on three datasets: $2048$ for Vorticity, $1024$ for Shallow-Water, and $512$ for Eagle.

\subsubsection{Training details}
\label{app:sssec_train_enc_dec}
\paragraph{Reconstruction performances}
To ensure fairness, we trained all encoder–decoders using a unified procedure. Specifically, we optimized a relative $L^2$ loss between the auto-encoder output and the original function evaluated on the full spatial grid (the $100\%$ setting).

Models are trained with the AdamW optimizer, using an initial learning rate of $10^{-3}$ that is annealed to $10^{-7}$ with a cosine schedule. Each model is trained for $20$ hours, and the batch size is tuned to best utilize the available GPU memory on an NVIDIA H100 (80,GB), with the chosen values reported in \cref{tab:bs-tab-exp-no}.

\begin{table}[htbp]
\centering
\caption{Training batch sizes used across datasets, for reconstruction
(\textit{Rec.}) and time-stepping (\textit{TS}).}
\label{tab:bs-tab-exp-no}
\small
\renewcommand{\arraystretch}{1.15}
\begin{tabular}{llcccc}
\toprule
Model & Task & Vorticity & Shallow-Water & Eagle & Cylinder Flow \\
\midrule
\multirow{2}{*}{GINO}     & Rec. & 8 & 16 & 32 & 32 \\
                          & TS   & 4 & 16 & 8  & 16 \\
\addlinespace[3pt]
\multirow{2}{*}{CORAL}    & Rec. & 1 & 2  & 16 & 32 \\
                          & TS   & 1 & 4  & 16 & 16 \\
\addlinespace[3pt]
\multirow{2}{*}{AROMA}    & Rec. & 1 & 1  & 4  & 32 \\
                          & TS   & 1 & 1  & 4  & 16 \\
\addlinespace[3pt]
\multirow{2}{*}{ENMA}     & Rec. & 1 & 1  & 1  & 8 \\
                          & TS   & 1 & 2  & 4  & 16 \\
\addlinespace[3pt]
\multirow{2}{*}{CALM-PDE} & Rec. & 1 & 1  & 2  & 2 \\
                          & TS   & 8 & 16 & 8  & 16 \\
\bottomrule
\end{tabular}
\end{table}
For ECHO, as we do spatio-temporal encoding, we divide our training into two stages, as explained in \cref{ssec:training}. We report the hyper-parameters considered in \cref{tab:hp-training-resolution}.

\begin{table}[htbp]
\centering
\caption{Training hyperparameters used across datasets for the low- and high-resolution stages.}
\label{tab:hp-training-resolution}
\small
\renewcommand{\arraystretch}{1.15}
\begin{tabular}{lcc}
\toprule
Hyperparameter            & Low resolution           & High resolution \\
\midrule
Epochs                    & 200                      & 10 \\
Batch size                & 8                        & 8 \\
Learning rate             & $10^{-3}$ (cosine decay) & $10^{-4}$ (cosine decay) \\
Weight decay              & $10^{-4}$                & $10^{-4}$ \\
Grad clip norm            & 1                        & 1 \\
Betas $(\beta_1,\beta_2)$ & (0.9, 0.95)              & (0.9, 0.95) \\
\bottomrule
\end{tabular}
\end{table}
\paragraph{Generative process training}
All models are evaluated on the forward task, where 4 time-steps are considered as context go generate the trajectory. Hyper-parameters are ported in \cref{tab:hp-training-generative} across all baselines. For this training, we considered a maximum budget of $20$ hours.

\begin{table}[htbp]
\centering
\caption{Training hyperparameters used across datasets for the generative stage.}
\label{tab:hp-training-generative}
\small
\renewcommand{\arraystretch}{1.15}
\begin{tabular}{lc}
\toprule
Hyperparameter            & Generative stage \\
\midrule
Epochs                    & 200 \\
Batch size                & 32 \\
Learning rate             & $10^{-3}$ (cosine decay) \\
Weight decay              & $10^{-4}$ \\
Grad clip norm            & 1 \\
Betas $(\beta_1,\beta_2)$ & (0.9, 0.95) \\
\bottomrule
\end{tabular}
\end{table}

\clearpage

\section{Additional experiments}
\label{app:add_exp}

To further assess ECHO’s capabilities beyond the main-text experiments, we report a series of additional studies covering long-range extrapolation, unconditional generation, robustness to sparse and irregular observations, and key architectural design choices. Specifically, we consider:
\begin{itemize}
    \item \textbf{Out-of-distribution evaluation} on the Vorticity equation: we evaluate the OOD capacities of ECHO on new unseen physical parameters (\cref{ssec:ood_vorticity}). 
    \item \textbf{Long-range prediction} on Shallow-Water, where ECHO is evaluated up to $4\times$ its training horizon (\cref{app:sec:long_range}). We also provide a spectra analysis on the Vorticity equation;
    \item \textbf{Generative trajectory sampling} in an unconditional setting on Shallow-Water, including Fréchet Physics Distance (FPD) and efficiency metrics (\cref{app:ssec:generative_expe});
    \item \textbf{Probabilistic evaluation} of ECHO with $16$-member ensembles (CRPS, calibration error, spread-skill ratio) on forward and backward prediction (\cref{app:ssec:probabilistic});
    \item \textbf{Initial value problem (IVP) evaluation}, where ECHO generates full trajectories from a single initial frame and PDE parameters (\cref{app:ssec:addexp_ivp});
    \item \textbf{Temporal interpolation tasks}, in which ECHO reconstructs long missing segments between observed frames across four datasets (\cref{app:ssec:addtasks});
    \item \textbf{Encoding sparse data / spatial interpolation}, where only $50\%$ or $10\%$ of the input grid is observed and the full grid must be reconstructed, with $10\%$ constituting a strong OOD sparsity level (\cref{app:ssec:addexp_aeood}).
\end{itemize}
In addition, we provide a set of ablation studies in \cref{app:ssec:addexp_ablation}, analyzing:
\begin{itemize}
    \item the impact of \emph{hierarchical compression depth} in the encoder on reconstruction accuracy (\cref{ssec:ablation_studies});
    \item the contribution of the \emph{refinement stage} for full-resolution Vorticity reconstruction (\cref{app:sssec_refine});
    \item the robustness of the \emph{flow-matching generative process} to the choice of ODE solver and number of integration steps at sampling time (\cref{app:sssec_gen_ablation}).
\end{itemize}

\subsection{Out-of-distribution evaluation }
\label{ssec:ood_vorticity}
In this section, we probe ECHO’s ability to (i) generalize to out-of-distribution PDE parameters and (ii) specialize at a specific task (Forward generation here) on new data. We vary the viscosity parameter: training uses $\nu \in [10^{-3},10^{-4}]$, while OOD evaluation uses $\nu \in [10^{-4},10^{-5}]$, yielding more turbulent, higher-Reynolds flows with different initial conditions. We consider two settings: (i) a strict \textbf{zero-shot} scenario, where the pre-trained model is directly applied to the new regime, and (ii) a \textbf{few-shot adaptation} scenario, where the model is lightly fine-tuned on a small set of OOD trajectories for the forward prediction task.
\begin{table}[htbp]
\centering
\caption{OOD generalization on Vorticity: performance of ECHO under
distribution shift, with zero-shot and few-shot adaptation. Relative $L^2$
error (lower is better).}
\label{tab:ood_vorticity}
\small
\renewcommand{\arraystretch}{1.15}
\setlength{\tabcolsep}{12pt}
\begin{tabular}{lc}
\toprule
Setting         & Rel.\ $L^2$ \\
\midrule
In-distribution & \sn{1.17}{-1} \\
OOD zero-shot   & \sn{3.39}{-1} \\
OOD few-shot    & \sn{1.12}{-1} \\
\bottomrule
\end{tabular}
\end{table}
\paragraph{Results} As shown in Table~\ref{tab:ood_vorticity}, ECHO displays strong zero-shot transfer to the more turbulent regime. Modest few-shot fine-tuning on OOD forward trajectories further improves accuracy, illustrating how a multi-task pre-trained operator can be efficiently adapted to new dynamics, paving the way toward universal PDE surrogates that are trained once and quickly specialized at low cost.

\subsection{Long-range prediction}
\label{app:sec:long_range}

\subsubsection{Long-range prediction on Vorticity}
\label{ssec:long-range-vort}
We further illustrate ECHO’s ability to generate long-range predictions beyond the training horizon on the Vorticity dataset, which exhibits turbulent regimes. \Cref{fig:long_range_vort} illustrates that ECHO remains consistent in long-rollouts, even on turbulent dynamics such as the Vorticity PDE. Note that we observed in \cref{fig:experimental_analysis} (middle) that ECHO accumulates less error than baselines.

\subsubsection{Long-range prediction on Shallow-Water}
\label{app:ssec:long_range_sw}

To further assess the ability of ECHO to perform long-range prediction, we extend the experiments of \cref{ssec:long-range-vort} to the Shallow-Water equation. On this dataset, ECHO is trained on trajectories of length $T = 40$ time steps, and evaluated in an out-of-distribution regime up to $4T = 160$.
\begin{figure*}[htbp]
\centering
\begin{minipage}[t]{0.48\textwidth}
  \vspace{0pt}
  \centering
  \includegraphics[width=\linewidth]{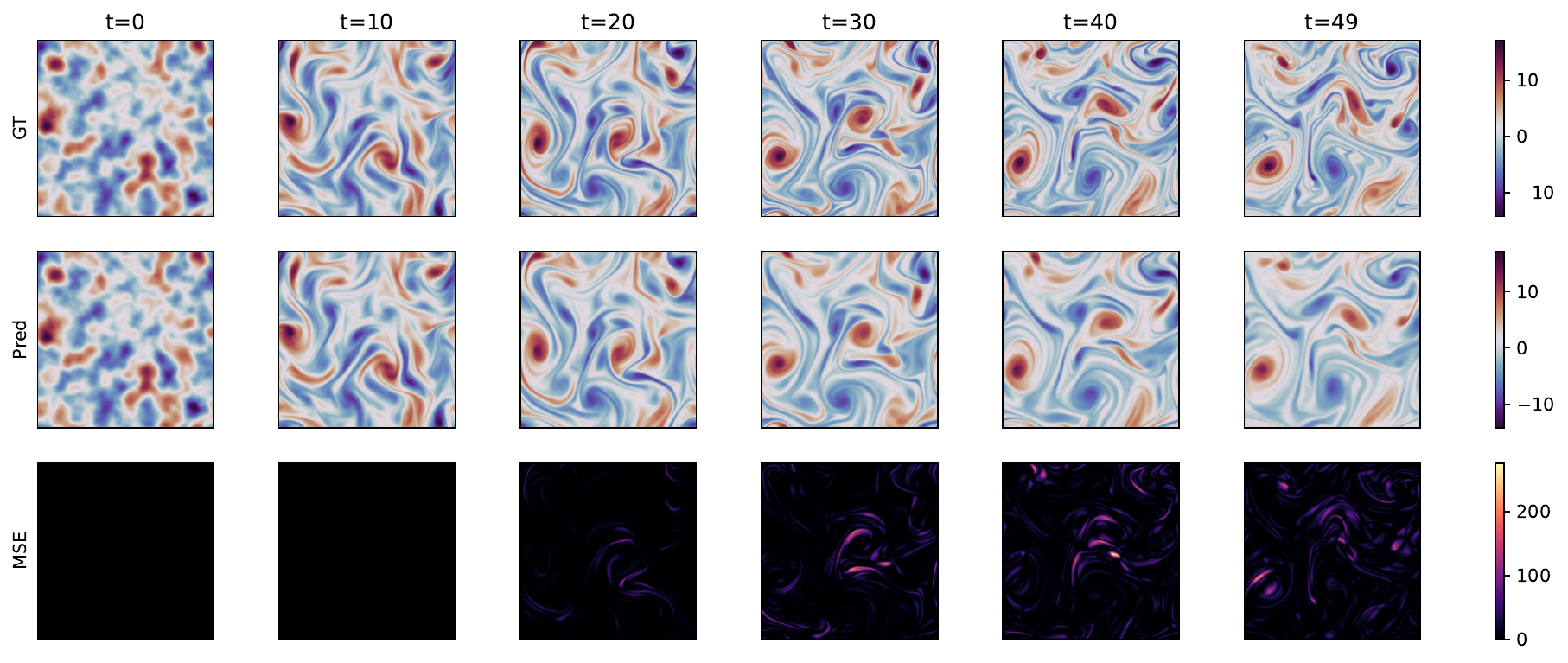}

  \vspace{0.5em}
  \footnotesize
  \renewcommand{\arraystretch}{1.15}
  \setlength{\tabcolsep}{0pt}
  \begin{tabular*}{\linewidth}{@{\extracolsep{\fill}}lccccc@{}}
  \toprule
  Horizon        & $0.5T$        & $T$           & $1.5T$        & $2T$          & $2.5T$ \\
  \midrule
  Rel.\ $L^2$    & \sn{7.70}{-2} & \sn{1.72}{-1} & \sn{2.62}{-1} & \sn{3.79}{-1} & \sn{4.79}{-1} \\
  \bottomrule
  \end{tabular*}
  \caption{Vorticity: long-range prediction with ECHO. Relative $L^2$ error
  at increasing horizons.}
  \label{fig:long_range_vort}
\end{minipage}%
\hfill
\begin{minipage}[t]{0.48\textwidth}
  \vspace{0pt}
  \centering
  \includegraphics[width=\linewidth]{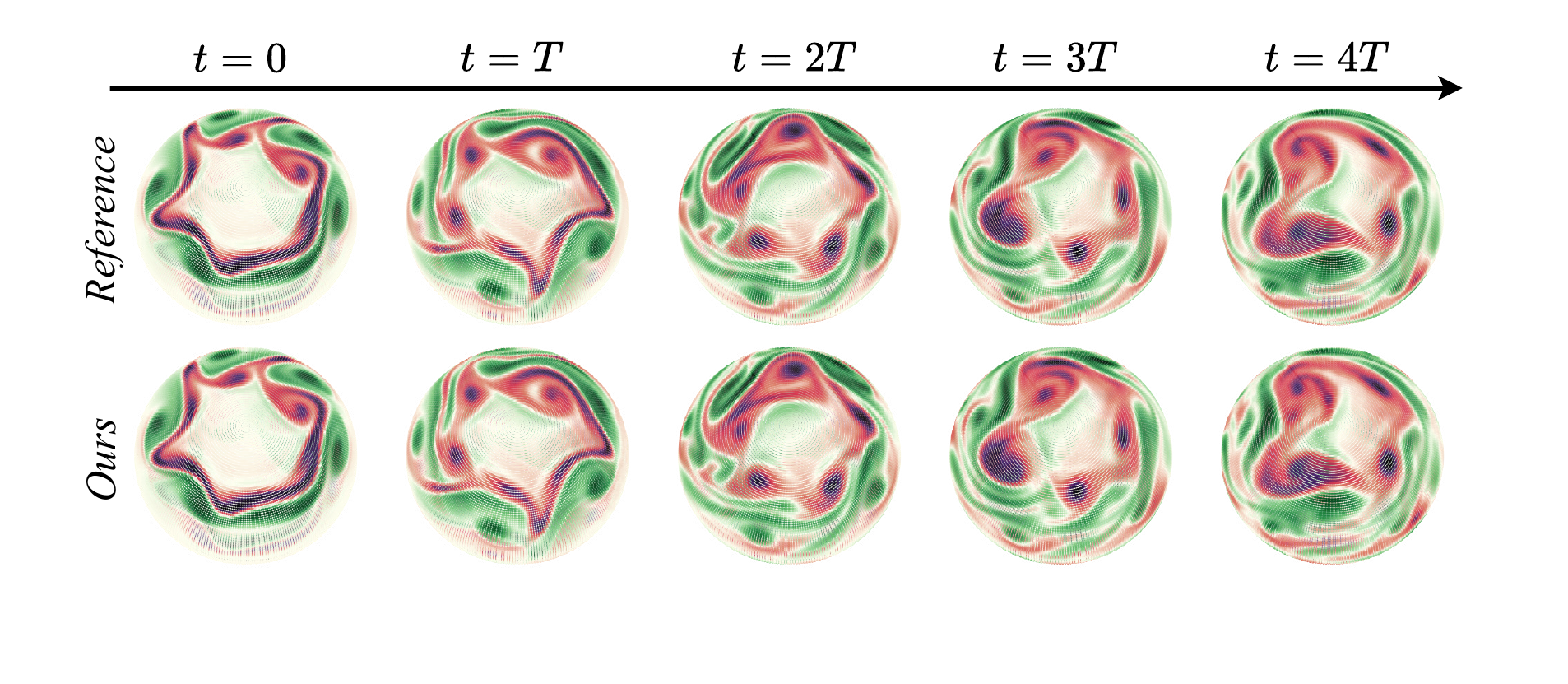}

  \vspace{0.5em}
  \footnotesize
  \renewcommand{\arraystretch}{1.15}
  \setlength{\tabcolsep}{0pt}
  \begin{tabular*}{\linewidth}{@{\extracolsep{\fill}}lcccc@{}}
  \toprule
  Horizon        & $T$           & $2T$          & $3T$          & $4T$ \\
  \midrule
  Rel.\ $L^2$    & \sn{1.21}{-2} & \sn{2.02}{-2} & \sn{2.11}{-2} & \sn{3.15}{-2} \\
  \bottomrule
  \end{tabular*}
  \caption{Shallow-Water: long-range prediction with ECHO ($T = 40$,
  evaluated up to $4T$). Relative $L^2$ error at increasing horizons.}
  \label{fig:long_range_sw}
\end{minipage}
\end{figure*}

\paragraph{Results} \Cref{fig:long_range_sw} shows that the error increases smoothly with the rollout horizon, from $1.21\times 10^{-2}$ at $T$ to $3.15\times 10^{-2}$ at $4T$. This moderate and monotonic degradation indicates that ECHO produces stable long-range forecasts even when extrapolating four times beyond its training horizon.

\subsubsection{Spectral analysis on the Vorticity generation}
\label{app:ssec:long_range_vort_spec}

While relative L2 is a standard metric in the PDE literature, it can hide over-smoothing, especially on turbulent fields where high-frequency features are important.

To better assess ECHO’s ability to preserve small-scale structures, we perform a spectral analysis on the \textit{Vorticity} dataset, both within the in-distribution training horizon and far beyond it. This type of analysis is standard in fluid dynamics and characterizes how energy is distributed across spatial scales, from large eddies at low wavenumbers $k$ to fine-scale structures at high $k$.

We proceed as follows: (i) we transform each physical field $u(x)$ into the frequency domain via FFT, obtaining Fourier coefficients $\mathcal{F}(u)(k)$ at accessible wavenumbers $k$; (ii) we compute the corresponding energy spectrum $E(k) = \tfrac{1}{2}|\mathcal{F}(u)(k)|^2$; (iii) we average the energy over all wavevectors with the same magnitude $|k|$ to obtain an isotropic spectrum.

\begin{figure}[htbp]
    \begin{subfigure}{0.48\textwidth}
        \centering
        \includegraphics[width=0.8\linewidth]{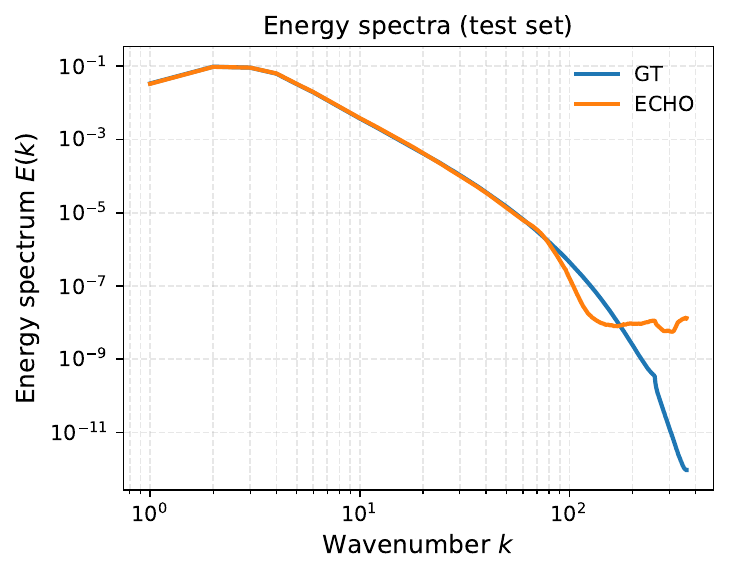}
        \caption{Spectral analysis for generation of a trajectory inside the training horizon ($T=20$). }
        \label{fig:spectre_iid}
    \end{subfigure}
   \begin{subfigure}{0.48\textwidth}
        \centering
        \includegraphics[width=0.8\linewidth]{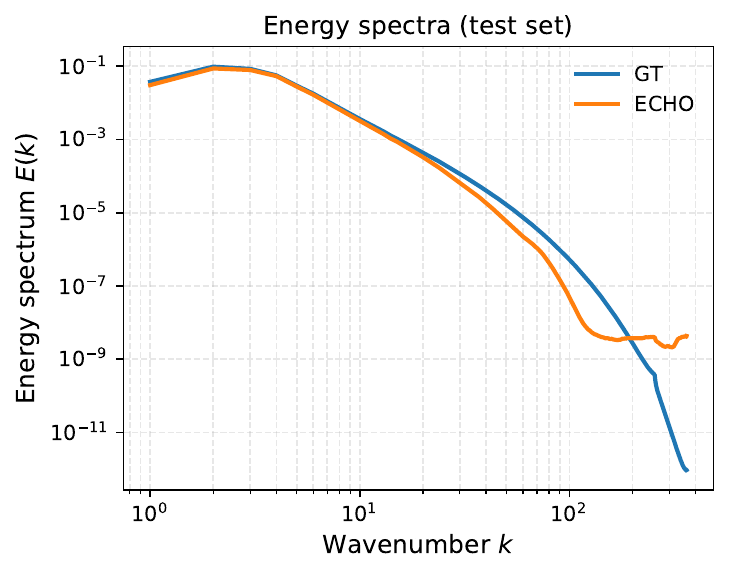}
        \caption{Spectral analysis for generation of a trajectory beyond the training horizon ($T=50$). }
        \label{fig:spectre_ood}
    \end{subfigure}
    \caption{Spectral analysis on a trajectory generation with ECHO on the \textit{Vorticity} dataset.}
    \label{fig:spectre}
\end{figure}

The resulting spectra are shown in \cref{fig:spectre} (done for the forward task): generation within the training horizon (up to $T=20$, left) and long-range generation beyond the training horizon (up to $T=50$, right). Each curve reports the energy spectrum averaged over all samples and time-steps in the corresponding temporal range.

\paragraph{Results} ECHO faithfully reconstructs the physics of the system across the most energy-dominant scales. The only discrepancy is a slight overestimation of energy at the very finest scales (high spatial frequencies), where most neural surrogates are known to fail. For a simulation running beyond its training horizon, ECHO shows good stability (the large scales are accurate). Beyond the training distribution, ECHO succeeds to preserve the low and medium scales, but fail to capture high and very-high spatial frequencies.

\paragraph{Origin of the high-frequency discrepancy.}
The deviation at high wavenumbers has two sources. The auto-encoder is trained with a relative $L^2$ reconstruction loss, which is dominated by the energy-carrying large scales and therefore fits low frequencies first, so the finest scales are the first affected by compression. Within the training horizon, this appears as a small excess of energy at the highest wavenumbers (\cref{fig:spectre}a). Over long rollouts, small errors at fine scales propagate to intermediate scales, so the departure from the reference moves toward lower wavenumbers beyond the training horizon (\cref{fig:spectre}b). Frequency-aware reconstruction objectives are a possible direction to reduce this effect.

\subsection{Generative data sampling}
\label{app:ssec:generative_expe}
As a demonstration of \textbf{ECHO}’s generative capability, we assess its ability to produce diverse and physically plausible trajectories (on Shallow-Water equation) and compare its efficiency to existing neural surrogates. We evaluate its generative quality using the \textbf{Fréchet Physics Distance (FPD)} \citep{enma}, an adaptation of the Fréchet Inception Distance widely used in image synthesis. FPD is computed in a compact 64-dimensional feature space extracted by a lightweight CNN encoder trained on physical trajectories, enabling semantically meaningful comparisons while avoiding the curse of dimensionality. Lower FPD indicates that generated trajectories are statistically closer to the training data distribution. For this evaluation, ECHO is trained in a fully \emph{unconditional} setting, where the entire trajectory is noised and then generated.

Beyond fidelity, we report standard \textbf{efficiency metrics} relevant for large-scale PDE surrogates:
\begin{itemize}
    \item \textbf{Throughput (Training)} — the number of samples processed per second during training (higher is better);
    \item \textbf{Latency} — average wall-clock time to generate a single trajectory at inference (lower is better);
    \item \textbf{Parameter count} — total learnable parameters, which provides a proxy for model complexity.
\end{itemize}

\begin{table}[htbp]
\centering
\caption{Efficiency and generative fidelity on Shallow-Water. Every
encoder--decoder is combined with our generative model. Throughput and latency
are measured on identical hardware; FPD measures the distributional alignment
with ground-truth trajectories. Best results in \textbf{bold}, second best
\underline{underlined}.}
\label{tab:throughput-latency}
\small
\renewcommand{\arraystretch}{1.15}
\setlength{\tabcolsep}{8pt}
\newcommand{\tabhead}[1]{\begin{tabular}[b]{@{}c@{}}#1\end{tabular}}
\begin{tabular}{lcccc}
\toprule
Encoder--decoder & \#Params
  & \tabhead{Training throughput\\(samples/s) $\uparrow$}
  & Latency (s) $\downarrow$
  & FPD $\downarrow$ \\
\midrule
GINO     & 19M  & \underline{55.7} & 1.83              & \sn{1.68}{-1} \\
CORAL    & 21M  & 4.3              & 0.75              & \sn{1.31}{-1} \\
AROMA    & 0.5M & 32.8             & \textbf{0.19}     & \usn{1.55}{-2} \\
CALM-PDE & 2M   & \textbf{60.5}    & 0.24              & \sn{1.02}{-1} \\
ECHO     & 30M  & 35.0             & \underline{0.21}  & \bsn{1.12}{-3} \\
\bottomrule
\end{tabular}
\end{table}

\paragraph{Results}
ECHO achieves the lowest FPD by a wide margin ($1.1\times 10^{-3}$), indicating its generated trajectories align extremely closely with the true data distribution, significantly outperforming existing efficient surrogates such as AROMA ($1.5\times 10^{-2}$) and CALM-PDE ($1.0\times 10^{-1}$). While slightly heavier in parameters than minimalistic encoders, ECHO maintains competitive \textbf{throughput} (35 samples/s) and extremely low \textbf{inference latency} (0.21s/trajectory), close to the fastest baseline AROMA (0.19s) while being substantially more accurate.

\subsection{Probabilistic evaluation}
\label{app:ssec:probabilistic}

All results in the main text are computed from a single sample per test trajectory, for generative and deterministic models alike. Single-sample errors do not assess the predictive distribution of a generative model. We therefore sample $16$ trajectories per test case and report three standard probabilistic metrics, together with the relative $L^2$ error of the ensemble mean. The \textbf{CRPS}, in its fair ensemble form, is a proper scoring rule that evaluates the full predictive distribution. The \textbf{RMSCE} (root mean squared calibration error) measures the gap between nominal and empirical coverage of the ensemble quantiles. The \textbf{SSR} (spread-skill ratio, with finite-ensemble correction) compares the ensemble spread with the error of the ensemble mean: a value of $1$ indicates that the spread matches the realized error, and values below $1$ indicate over-confident ensembles.

\begin{table}[htbp]
\centering
\caption{Probabilistic evaluation with $16$-member ensembles, using the same
windows and number of generated frames as \cref{tab:model-comparison}.
Rel.~$L^2$ is computed on the ensemble mean. Bold marks the best forward result
per dataset; ECHO backward rows use identical settings.}
\label{tab:probabilistic}
\small
\renewcommand{\arraystretch}{1.15}
\setlength{\tabcolsep}{5pt}
\begin{tabular}{lllcccc}
\toprule
Dataset & Task & Model & CRPS $\downarrow$ & RMSCE $\downarrow$ & SSR $\rightarrow 1$ & Rel.~$L^2$ $\downarrow$ \\
\midrule
\multirow{3}{*}{Gray--Scott}
 & \multirow{2}{*}{Forward} & ENMA & \sn{8.68}{-3}  & \sn{3.16}{-1}  & 0.43          & \sn{3.96}{-2} \\
 &                          & ECHO & \bsn{8.66}{-3} & \bsn{9.34}{-2} & \textbf{0.93} & \bsn{3.90}{-2} \\
 \cmidrule(lr){2-7}
 & Backward                 & ECHO & \sn{9.69}{-3}  & \sn{7.94}{-2}  & 1.06          & \sn{4.41}{-2} \\
\midrule
\multirow{3}{*}{Rayleigh--B\'enard}
 & \multirow{2}{*}{Forward} & ENMA & \bsn{9.47}{-3} & \sn{2.03}{-1}  & 0.53          & \sn{9.38}{-2} \\
 &                          & ECHO & \sn{1.04}{-2}  & \bsn{8.97}{-2} & \textbf{0.86} & \bsn{9.02}{-2} \\
 \cmidrule(lr){2-7}
 & Backward                 & ECHO & \sn{1.30}{-2}  & \sn{4.77}{-2}  & 0.90          & \sn{1.14}{-1} \\
\midrule
\multirow{3}{*}{Active Matter}
 & \multirow{2}{*}{Forward} & ENMA & \sn{1.11}{-1}  & \sn{4.51}{-1}  & 0.12          & \sn{2.98}{-1} \\
 &                          & ECHO & \bsn{6.79}{-2} & \bsn{1.73}{-1} & \textbf{0.67} & \bsn{2.33}{-1} \\
 \cmidrule(lr){2-7}
 & Backward                 & ECHO & \sn{6.85}{-2}  & \sn{1.19}{-1}  & 0.80          & \sn{2.41}{-1} \\
\midrule
\multirow{3}{*}{ASM}
 & \multirow{2}{*}{Forward} & ENMA & \sn{5.16}{-3}  & \sn{1.97}{-1}  & 0.59          & \sn{3.41}{-1} \\
 &                          & ECHO & \bsn{1.65}{-3} & \bsn{1.37}{-1} & \textbf{0.82} & \bsn{1.19}{-1} \\
 \cmidrule(lr){2-7}
 & Backward                 & ECHO & \sn{1.77}{-3}  & \sn{1.27}{-1}  & 0.96          & \sn{1.20}{-1} \\
\bottomrule
\end{tabular}
\end{table}

\paragraph{Forward prediction.}
On forward prediction (\cref{tab:probabilistic}), ECHO is well calibrated, with RMSCE between \sci{9.0}{-2} and \sci{1.7}{-1} and SSR between $0.67$ and $0.93$. ENMA is under-dispersed on every system, with SSR between $0.12$ and $0.53$, meaning that its ensembles are over-confident relative to their realized error. ENMA obtains a slightly lower CRPS on \emph{Rayleigh-Bénard}; on the three other systems, ECHO is better on all four metrics.

\paragraph{Backward prediction.}
Backward prediction can be ill-posed, since several earlier states may be consistent with the observed frames, which makes the calibration of the ensembles the relevant property. Backward performance of ECHO is close to its forward performance on all four systems, with CRPS increasing by $1\%$ (\emph{Active Matter}) to $25\%$ (\emph{Rayleigh-Bénard}). The backward ensembles are well calibrated, with RMSCE between \sci{4.8}{-2} and \sci{1.3}{-1} and SSR between $0.80$ and $1.06$: the predicted spread reflects the actual uncertainty of the reconstruction.

\subsection{Initial Value Problem}
\label{app:ssec:addexp_ivp}
We further evaluate ECHO in an \emph{initial value problem} (IVP) setting, where the model uses as context only the first frame of the trajectory and the governing PDE parameter (here, the viscosity) as inputs. This setting tests the model’s ability to generate a full trajectory from minimal information, a common scenario in real-world PDE simulations.

We consider the \textbf{Vorticity} system on a $512\times512$ grid and report relative L2 reconstruction error over the full rollout. When conditioned only on the initial state and viscosity, ECHO achieves a relative L2 of \textbf{$1.15\times 10^{-1}$}, demonstrating robust trajectory generation despite extremely sparse input.

\begin{figure}[htbp]
    \centering
    \includegraphics[width=0.8\linewidth]{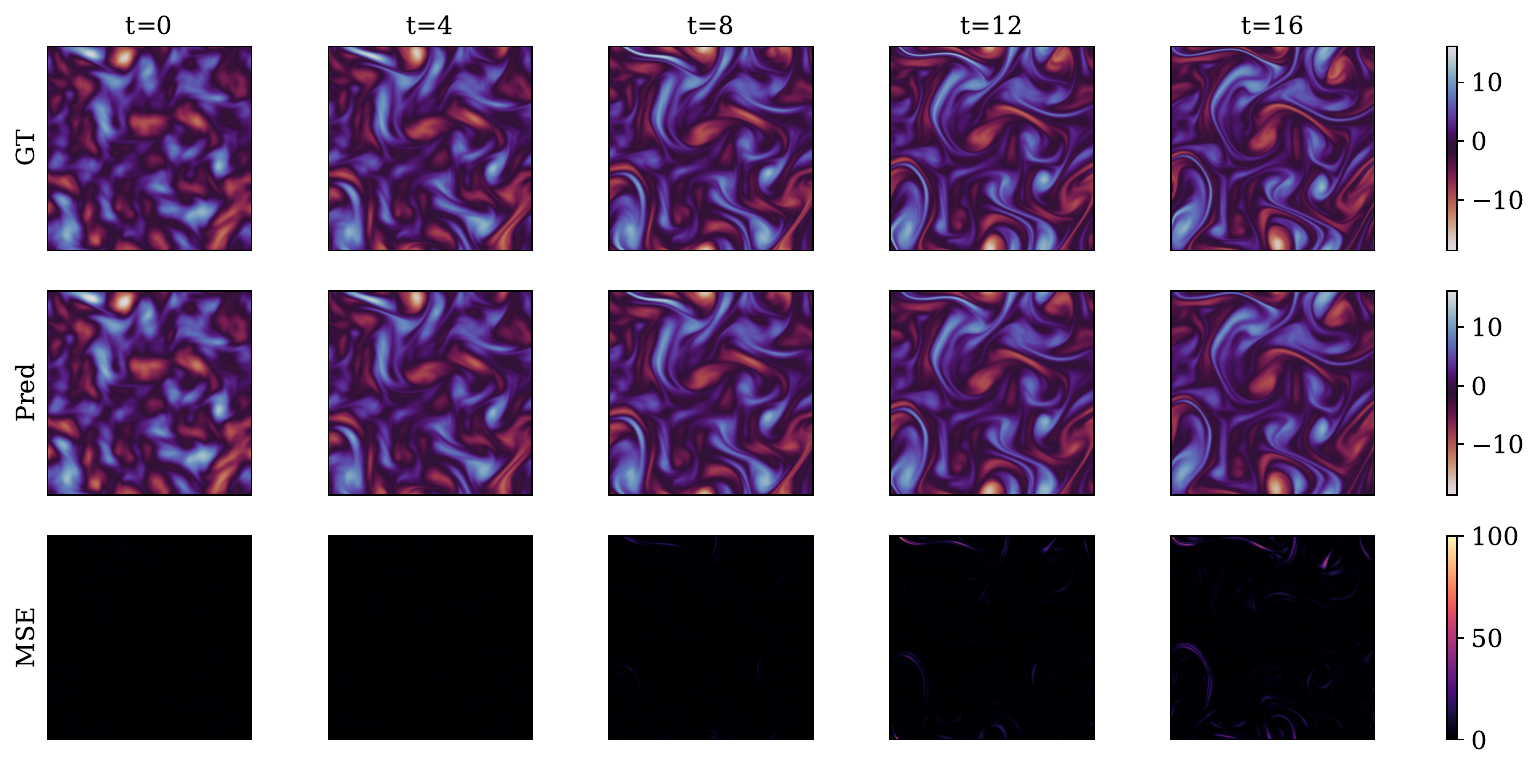}
    \caption{\textbf{Initial value problem on Vorticity ($512\times512$).} 
    Ground truth (GT), ECHO predictions, and per-frame error maps when generating the full trajectory from only the first state and PDE viscosity.}
    \label{fig:ivp_parameters}
\end{figure}

\subsection{Interpolation tasks}
\label{app:ssec:addtasks}

In addition to the forward and backward prediction settings discussed in the main text, we evaluate ECHO on a temporal interpolation task. Given the first $2$ and last $2$ frames of each trajectory as context, the model must reconstruct all intermediate frames (11 frames for Rayleigh--Bénard and ASM, 18 for Active Matter, and 36 for Gray--Scott). We report the Relative $L^2$ over the interpolated segment,together with backward and forward errors, in \cref{tab:model-comparison-app-interpolation}.

\begin{table}[htbp]
\centering
\caption{Backward (\textit{Back.}), forward (\textit{For.}) and temporal
interpolation (\textit{Interp.}) results across four dynamical systems.
Errors are relative $L^2$ errors (lower is better). Interpolation is evaluated
for ECHO only; latencies are reported in \cref{tab:model-comparison}. Best
results in \textbf{bold}, second best \underline{underlined}; ``--'' indicates
non-convergence.}
\label{tab:model-comparison-app-interpolation}
\small
\renewcommand{\arraystretch}{1.15}
\setlength{\tabcolsep}{6pt}
\begin{tabular}{l ccc@{\hspace{18pt}}ccc}
\toprule
& \multicolumn{3}{c}{Rayleigh--B\'enard} & \multicolumn{3}{c}{Gray--Scott} \\
\cmidrule(lr){2-4}\cmidrule(lr){5-7}
Model & Back. & For. & Interp. & Back. & For. & Interp. \\
\midrule
\multicolumn{7}{l}{\textit{Deterministic models}} \\
\addlinespace[2pt]
FNO          & \sn{2.47}{3}   & \sn{4.23}{-1}  &                & --             & --             & \\
Transolver++ & \sn{6.34}{-1}  & \sn{3.31}{-1}  &                & \sn{4.43}{-1}  & \sn{2.34}{-1}  & \\
BCAT         & \sn{1.91}{-1}  & \sn{1.06}{-1}  &                & \sn{2.19}{-1}  & \sn{8.82}{-2}  & \\
AViT         & \sn{4.50}{-1}  & \sn{1.01}{-1}  &                & \sn{1.66}{-1}  & \sn{7.42}{-2}  & \\
ECHO         & \sn{2.53}{-1}  & \sn{1.32}{-1}  & \sn{1.08}{-1}  & \usn{8.36}{-2} & \sn{7.66}{-2}  & \sn{3.12}{-2} \\
\midrule
\multicolumn{7}{l}{\textit{Generative models}} \\
\addlinespace[2pt]
ENMA         & \usn{1.71}{-1} & \usn{9.87}{-2} &                & \sn{1.08}{-1}  & \usn{5.44}{-2} & \\
ECHO         & \bsn{1.16}{-1} & \bsn{9.28}{-2} & \bsn{9.02}{-2} & \bsn{2.53}{-2} & \bsn{5.12}{-2} & \bsn{1.78}{-2} \\
\midrule[\heavyrulewidth]
& \multicolumn{3}{c}{Active Matter} & \multicolumn{3}{c}{ASM} \\
\cmidrule(lr){2-4}\cmidrule(lr){5-7}
Model & Back. & For. & Interp. & Back. & For. & Interp. \\
\midrule
\multicolumn{7}{l}{\textit{Deterministic models}} \\
\addlinespace[2pt]
FNO          & --             & --             &                & 1.87           & 1.52           & \\
Transolver++ & \sn{7.33}{-1}  & \sn{6.91}{-1}  &                & 1.03           & \sn{9.64}{-1}  & \\
BCAT         & \sn{4.98}{-1}  & \sn{4.56}{-1}  &                & \sn{1.95}{-1}  & \sn{2.18}{-1}  & \\
AViT         & \sn{4.50}{-1}  & \sn{4.62}{-1}  &                & \bsn{1.04}{-1} & \usn{1.52}{-1} & \\
ECHO         & \sn{5.74}{-1}  & \sn{3.44}{-1}  & \sn{1.47}{-1}  & 1.47           & \sn{2.53}{-1}  & \sn{1.88}{-1} \\
\midrule
\multicolumn{7}{l}{\textit{Generative models}} \\
\addlinespace[2pt]
ENMA         & \usn{4.27}{-1} & \usn{3.33}{-1} &                & 1.01           & \sn{4.12}{-1}  & \\
ECHO         & \bsn{3.55}{-1} & \bsn{2.87}{-1} & \bsn{1.11}{-1} & \usn{1.12}{-1} & \bsn{1.32}{-1} & \bsn{1.00}{-1} \\
\bottomrule
\end{tabular}
\end{table}

\paragraph{Results}  ECHO shows strong interpolation capabilities across all four systems, both for the deterministic and generative version. These results indicate that ECHO reliably fill in long missing segments between observed states.

\subsection{Encoding sparse data - Spatial interpolation}
\label{app:ssec:addexp_aeood}
In addition to \cref{tab:exp_no}, we evaluate the encoder--decoder of ECHO and all baselines under more challenging sparse-input regimes. In this experiment, the models receive only a subset of the input grid and must reconstruct the full-resolution output grid. We consider two sampling levels at test time: an intermediate sparsity of $50\%$ of the input points, and an extreme sparsity of $10\%$. 

The $10\%$ setting is out-of-distribution, since all models were trained with input samplings ranging only from $25\%$ to $75\%$. This evaluation therefore probes the robustness of each method to significantly more aggressive subsampling at inference. The corresponding results are reported in \cref{tab:exp_no_ood}.

\begin{table}[htbp]
\centering
\caption{Encoder--decoder results from sparse inputs: only a fraction
$\mathcal{X}_{te}$ of the input grid is observed at test time and the full grid
is reconstructed. The $10\%$ setting is out-of-distribution (training samplings
range from $25\%$ to $75\%$). \textit{Rec.} and \textit{For.} denote
reconstruction and forward generation; \textit{Irr.} gives the strategy for
handling irregular points, and \textit{Hier.} indicates hierarchical (\cmark)
or direct (\xmark) compression into the latent space. Metrics are relative
$L^2$ errors (lower is better). Best results in \textbf{bold}, second best
\underline{underlined}.}
\label{tab:exp_no_ood}
\footnotesize
\renewcommand{\arraystretch}{1.15}
\setlength{\tabcolsep}{4pt}
\begin{tabular}{llc cc cc cc cc}
\toprule
& & & \multicolumn{2}{c}{Vorticity} & \multicolumn{2}{c}{Shallow-Water}
    & \multicolumn{2}{c}{Eagle} & \multicolumn{2}{c}{Cylinder Flow} \\
\cmidrule(lr){4-5}\cmidrule(lr){6-7}\cmidrule(lr){8-9}\cmidrule(lr){10-11}
Irr. & Encoder & Hier. & Rec. & For. & Rec. & For. & Rec. & For. & Rec. & For. \\
\midrule
\multicolumn{11}{l}{\textit{Intermediate sparsity} ($\mathcal{X}_{te} = 50\%$)} \\
\addlinespace[2pt]
Graph & GINO & \xmark
  & \sn{9.99}{-1}  & 1.00
  & \sn{8.69}{-1}  & 1.11
  & \sn{5.80}{-1}  & 1.88
  & \sn{7.94}{-1}  & \sn{8.65}{-1} \\
INR & CORAL & \xmark
  & \sn{5.14}{-1}  & 1.34
  & \sn{2.29}{-1}  & \sn{6.88}{-1}
  & \sn{6.11}{-1}  & 1.54
  & \sn{3.04}{-1}  & \sn{5.07}{-1} \\
\multirow{2}{*}{Attention} & AROMA & \xmark
  & \sn{5.13}{-1}  & \sn{8.43}{-1}
  & \usn{2.69}{-2} & \usn{4.28}{-2}
  & \sn{3.09}{-1}  & \usn{3.34}{-1}
  & \bsn{3.17}{-2} & \sn{2.31}{-1} \\
& ENMA & \cmark
  & \sn{4.39}{-1}  & \sn{4.39}{-1}
  & \sn{7.42}{-2}  & \sn{7.53}{-2}
  & \usn{3.02}{-1} & \sn{3.42}{-1}
  & \sn{7.77}{-2}  & \usn{1.09}{-1} \\
\multirow{2}{*}{Convolution} & CALM-PDE & \cmark
  & \usn{2.76}{-1} & \usn{4.26}{-1}
  & \sn{2.61}{-1}  & \sn{2.84}{-1}
  & \sn{9.34}{-1}  & \sn{9.55}{-1}
  & \sn{2.50}{-1}  & \sn{3.59}{-1} \\
& ECHO & \cmark
  & \bsn{7.32}{-2} & \bsn{2.23}{-1}
  & \bsn{1.94}{-2} & \bsn{2.00}{-2}
  & \bsn{1.88}{-1} & \bsn{2.90}{-1}
  & \usn{4.72}{-2} & \bsn{9.63}{-2} \\
\midrule
\multicolumn{11}{l}{\textit{Extreme sparsity, out-of-distribution} ($\mathcal{X}_{te} = 10\%$)} \\
\addlinespace[2pt]
Graph & GINO & \xmark
  & \sn{9.99}{-1}  & 1.00
  & \sn{8.83}{-1}  & 1.11
  & \sn{7.82}{-1}  & 1.15
  & \sn{7.98}{-1}  & \sn{8.64}{-1} \\
INR & CORAL & \xmark
  & \sn{5.27}{-1}  & 1.33
  & \sn{2.31}{-1}  & \sn{6.86}{-1}
  & \sn{6.66}{-1}  & 1.53
  & \sn{3.77}{-1}  & \sn{5.19}{-1} \\
\multirow{2}{*}{Attention} & AROMA & \xmark
  & \sn{5.16}{-1}  & \sn{8.44}{-1}
  & \bsn{4.89}{-2} & \sn{5.43}{-1}
  & \sn{5.26}{-1}  & 1.01
  & \bsn{7.12}{-2} & \sn{2.36}{-1} \\
& ENMA & \cmark
  & \usn{4.45}{-1} & \usn{7.78}{-1}
  & \usn{9.35}{-2} & \usn{8.93}{-2}
  & \usn{4.72}{-1} & \bsn{6.86}{-1}
  & \usn{1.10}{-1} & \usn{1.44}{-1} \\
\multirow{2}{*}{Convolution} & CALM-PDE & \cmark
  & \bsn{3.24}{-1} & \bsn{4.59}{-1}
  & \sn{3.03}{-1}  & \sn{3.17}{-1}
  & \sn{9.79}{-1}  & 1.02
  & \sn{3.08}{-1}  & \sn{4.14}{-1} \\
& ECHO & \cmark
  & \sn{5.60}{-1}  & \sn{9.86}{-1}
  & \sn{1.06}{-1}  & \bsn{8.69}{-2}
  & \bsn{3.82}{-1} & \usn{7.31}{-1}
  & \sn{1.13}{-1}  & \bsn{1.41}{-1} \\
\bottomrule
\end{tabular}
\end{table}
\paragraph{Result}  Overall, \cref{tab:exp_no_ood} shows that convolution-based hierarchical models (CALM-PDE and ECHO) outperform graph- and INR-based baselines both for reconstruction and time-stepping. At $50\%$ sampling (in-distribution), ECHO achieves the best or second-best performance across most datasets. In the more challenging $10\%$ sampling regime, which constitutes a strong OOD setting relative to the training samplings, all models degrade, but ECHO remains competitive and often matches or surpasses alternative baselines, especially on the more complex Eagle. 

\clearpage
\subsection{Ablation studies}
\label{app:ssec:addexp_ablation}

\subsubsection{Impact of hierarchical compresion}
\label{ssec:ablation_studies}
To assess the benefit of hierarchical compression in the encoder, we conduct an ablation study where the final latent space size (the bottleneck resolution) is kept fixed, but the way we reach this compressed space is varied.

A hierarchical level refers to how many successive downsampling stages are applied in space; each level reduces the spatial resolution by a factor of two along each dimension. For instance, a 3-level hierarchy means three consecutive $2\times$ spatial reductions before reaching the final latent grid, whereas level 0 means a single direct projection from the input resolution to the final latent size (no intermediate stages). Importantly, in this experiment we only compress space (not time), so the temporal dimension remains unchanged (\cref{tab:hier}).

\begin{table}[htbp]
\centering
\caption{Impact of hierarchical depth in the encoder on reconstruction
(relative $L^2$ error). A higher level means more progressive spatial
downsampling, with the same final latent size.}
\label{tab:hier}
\small
\renewcommand{\arraystretch}{1.15}
\setlength{\tabcolsep}{8pt}
\begin{tabular}{lcccc}
\toprule
Hierarchical level & 0             & 1             & 2             & 3 \\
\midrule
Rel.\ $L^2$        & \sn{3.94}{-1} & \sn{3.64}{-1} & \sn{1.09}{-1} & \bsn{2.87}{-2} \\
\bottomrule
\end{tabular}
\end{table}

\paragraph{Results}  Table \ref{tab:hier} shows that increasing hierarchical depth dramatically improves reconstruction accuracy while keeping the same final latent size. Without hierarchy (level 0), Relative $L^2$ is high (3.94e-1), while introducing even one level of progressive compression significantly reduces error. The best performance is achieved with three levels (Relative $L^2$ 2.87e-2), highlighting that iterative, deep compression yields more expressive and informative latent representations than a single aggressive projection.

This validates our design choice of a deep hierarchical encoder: even at fixed compression ratio, progressive downsampling better preserves spatial structure and leads to markedly improved fidelity.

\subsubsection{Impact of the refinement stage}
\label{app:sssec_refine}

To evaluate the contribution of the \textit{refinement stage}, we conduct an ablation on the \textbf{Vorticity} dataset at full spatial resolution.  
Our training strategy first learns from spatio-temporal data that is \emph{spatially subsampled} to reduce memory cost.  
After this pretraining, we apply a refinement stage that re-encodes each frame individually at \emph{full spatial density}, improving the latent representation for high-resolution inputs.

\begin{table}[htbp]
\centering
\caption{Effect of the refinement stage on full-grid Vorticity reconstruction
(relative $L^2$ error, lower is better).}
\label{tab:refine}
\small
\renewcommand{\arraystretch}{1.15}
\setlength{\tabcolsep}{8pt}
\begin{tabular}{lcc}
\toprule
            & w/o refinement & w/ refinement \\
\midrule
Rel.\ $L^2$ & \sn{2.24}{-1}  & \bsn{6.88}{-2} \\
\bottomrule
\end{tabular}
\end{table}

\paragraph{Results}  This additional step proves particularly beneficial for highly detailed fields such as Vorticity, where local fine-scale structures are hard to preserve under strong compression.  
As shown in \cref{tab:refine}, adding the refinement stage significantly improves reconstruction accuracy.

\subsubsection{Ablations on the generative process}
\label{app:sssec_gen_ablation}

We conduct ablations on the vorticity dataset to validate the robustness of our flow-matching generative design. In particular, we vary both the ODE solver used at sampling time and the number of denoising steps, while keeping the learned model fixed. 

\begin{table}[htbp]
\centering
\caption{Ablation of the flow-matching generative process on Vorticity. We
vary the ODE solver and the number of integration steps used at sampling time.
Relative $L^2$ error (lower is better).}
\label{tab:fm_ode_ablation}
\small
\renewcommand{\arraystretch}{1.15}
\setlength{\tabcolsep}{8pt}
\begin{tabular}{llc}
\toprule
Ablation & Configuration & Rel.\ $L^2$ \\
\midrule
\multirow{4}{*}{Solver}
  & Euler    & \sn{1.62}{-1} \\
  & Midpoint & \sn{1.67}{-1} \\
  & RK4      & \sn{1.68}{-1} \\
  & DOPRI5   & \sn{1.71}{-1} \\
\midrule
\multirow{4}{*}{\# Steps}
  & 2 steps  & \sn{1.72}{-1} \\
  & 5 steps  & \sn{1.69}{-1} \\
  & 10 steps & \sn{1.68}{-1} \\
  & 25 steps & \sn{1.67}{-1} \\
\bottomrule
\end{tabular}
\end{table}

\paragraph{Results} 
Overall, the reconstruction errors remain in a very narrow range across all configurations, with differences on the order of $10^{-3}$. Simple solvers such as Euler already perform on par with higher-order methods (RK4, DOPRI5), and increasing the number of integration steps from $2$ to $25$ brings only marginal gains. These results indicate that our flow-matching generative process is stable with respect to the choice of ODE solver and discretization.

\clearpage
\section{Visualizations}
\label{app:visu}
We provide qualitative visualizations of ECHO's generated trajectories and a comparison with ground truth trajectories. 

\subsection{Vorticity}
\label{app:ssec:visu_Vorticity}

\begin{figure}[htbp]
    \centering
    \includegraphics[width=\linewidth]{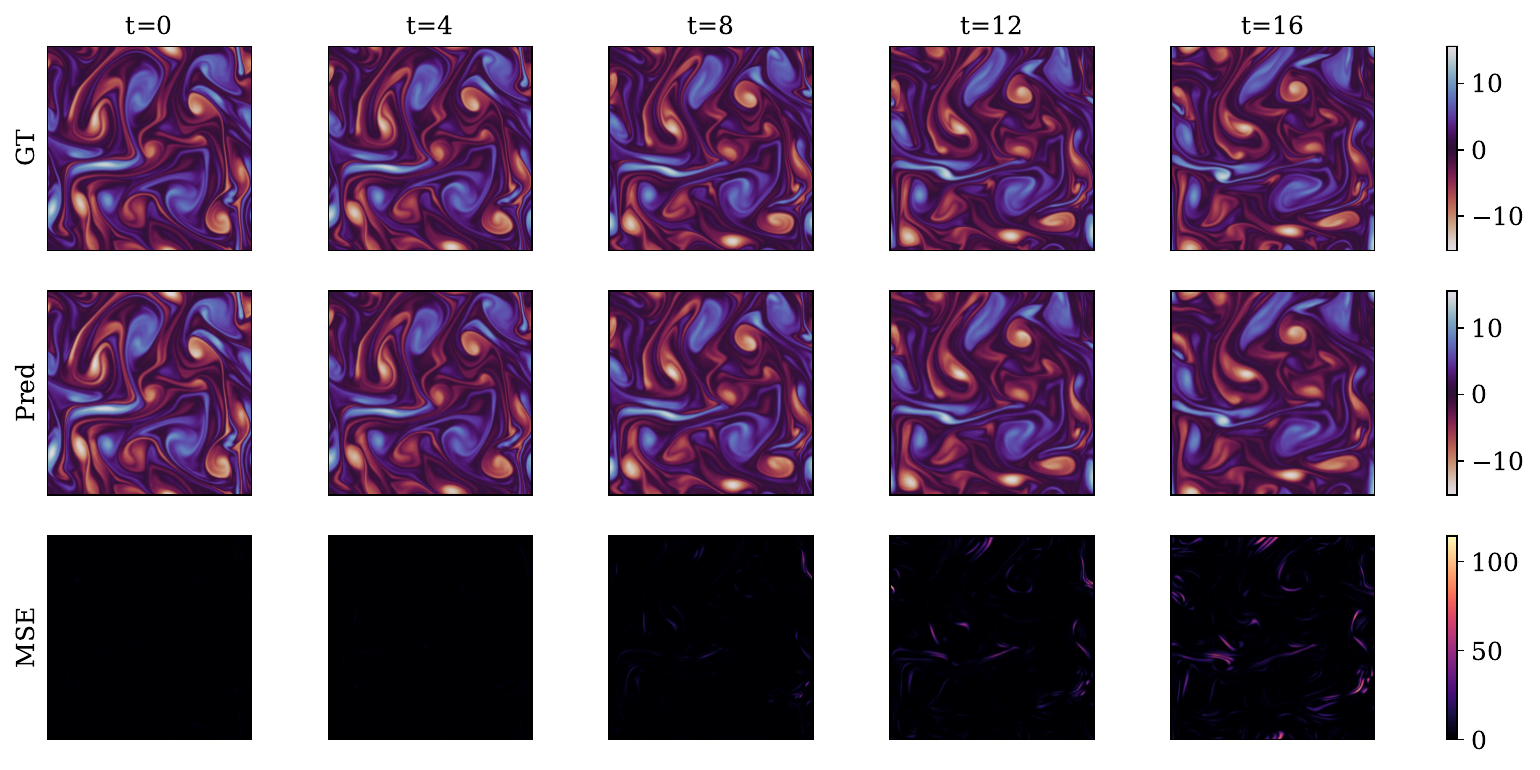}
    \caption{Generation using ECHO's framework on the Vorticity dataset. We generate $1024\times1024$ spatial points per frame. (Top) ground truth trajectory, (Middle) ECHO's prediction, (Bottom) Error.}
    \label{fig:echovorticity}
\end{figure}

\clearpage
\subsection{Gray-Scott}
\label{app:ssec:visu_gs}

\begin{figure}[htbp]
    \centering
    \includegraphics[width=\linewidth]{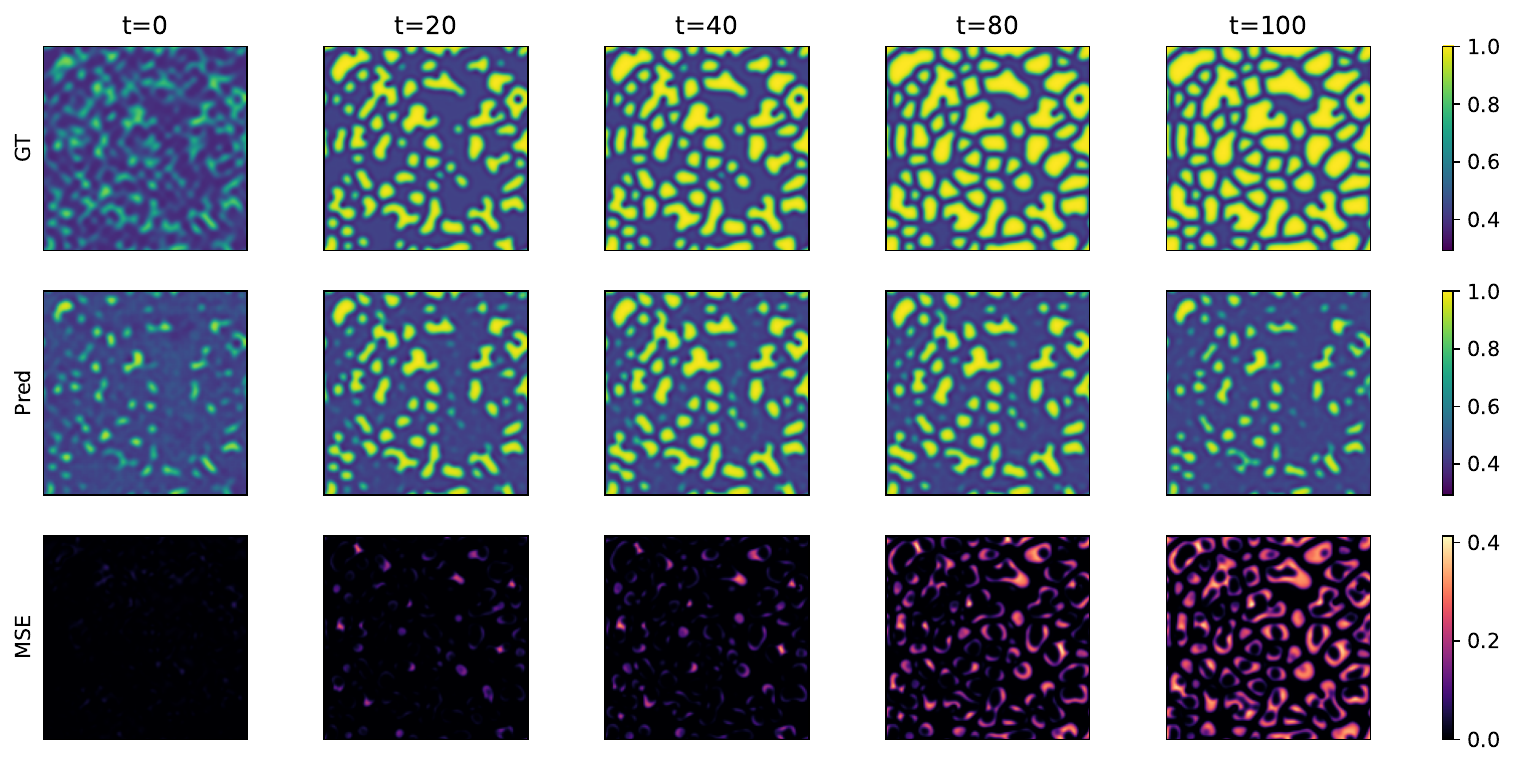}
    \caption{Generation using ECHO's framework on the Gray-Scott dataset. (Top) ground truth trajectory, (Middle) ECHO's prediction, (Bottom) Error.}
    \label{fig:echograyscott}
\end{figure}

\clearpage
\subsection{Rayleigh-Benard}
\label{app:ssec:visu_rb}

\begin{figure}[htbp]
    \centering
    \includegraphics[width=\linewidth]{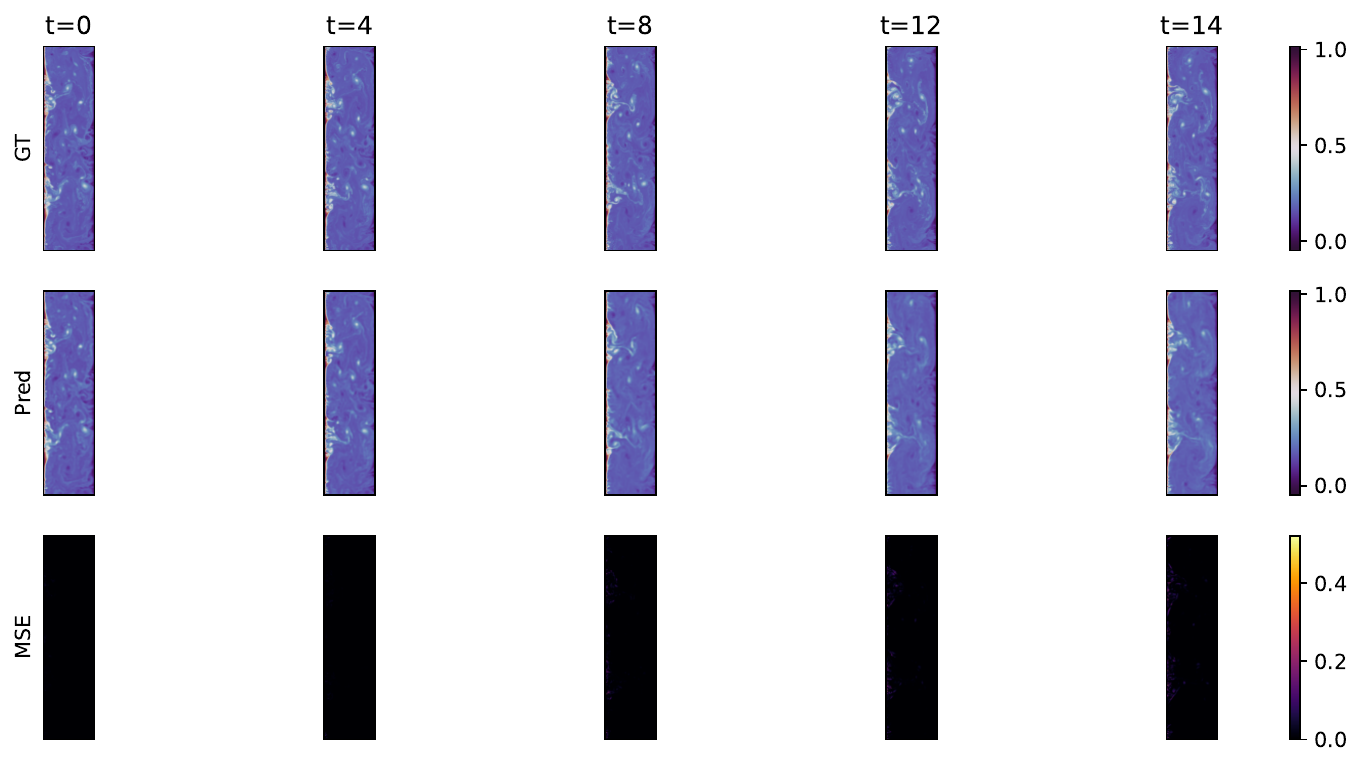}
    \caption{Generation using ECHO's framework on the Rayleigh-Benard dataset. (Top) ground truth trajectory, (Middle) ECHO's prediction, (Bottom) Error.}
    \label{fig:fluxrb}
\end{figure}

\clearpage
\subsection{Active-Matter}
\label{app:ssec:visu_am}

\begin{figure}[htbp]
    \centering
    \includegraphics[width=\linewidth]{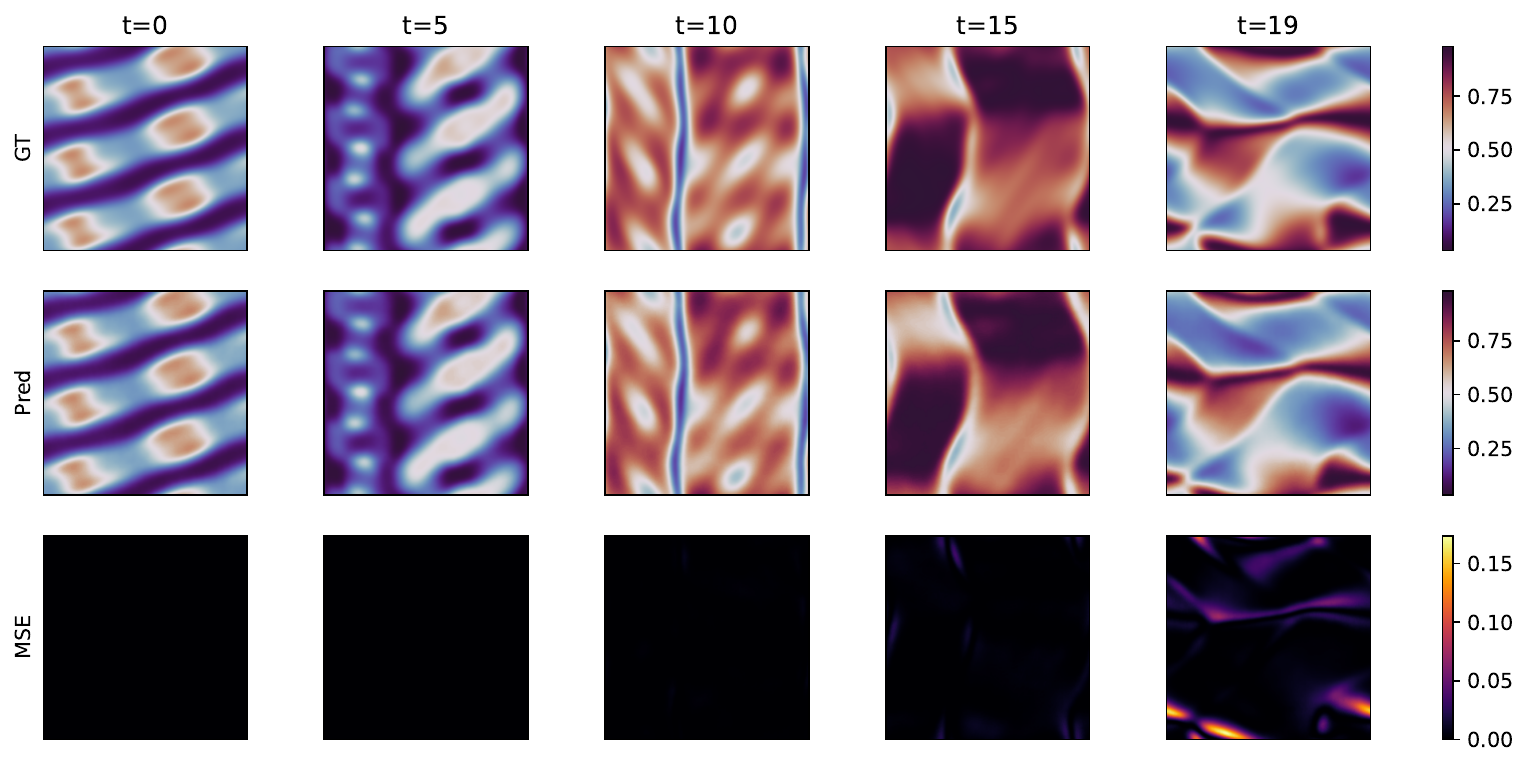}
    \caption{Generation using ECHO's framework on the Active Matter dataset. (Top) ground truth trajectory, (Middle) ECHO's prediction, (Bottom) Error.}
    \label{fig:fluxam}
\end{figure}
\clearpage

\subsection{Acoustic Scattering Maze}
\label{app:ssec:visu_asm}

\begin{figure}[htbp]
    \centering
    \includegraphics[width=\linewidth]{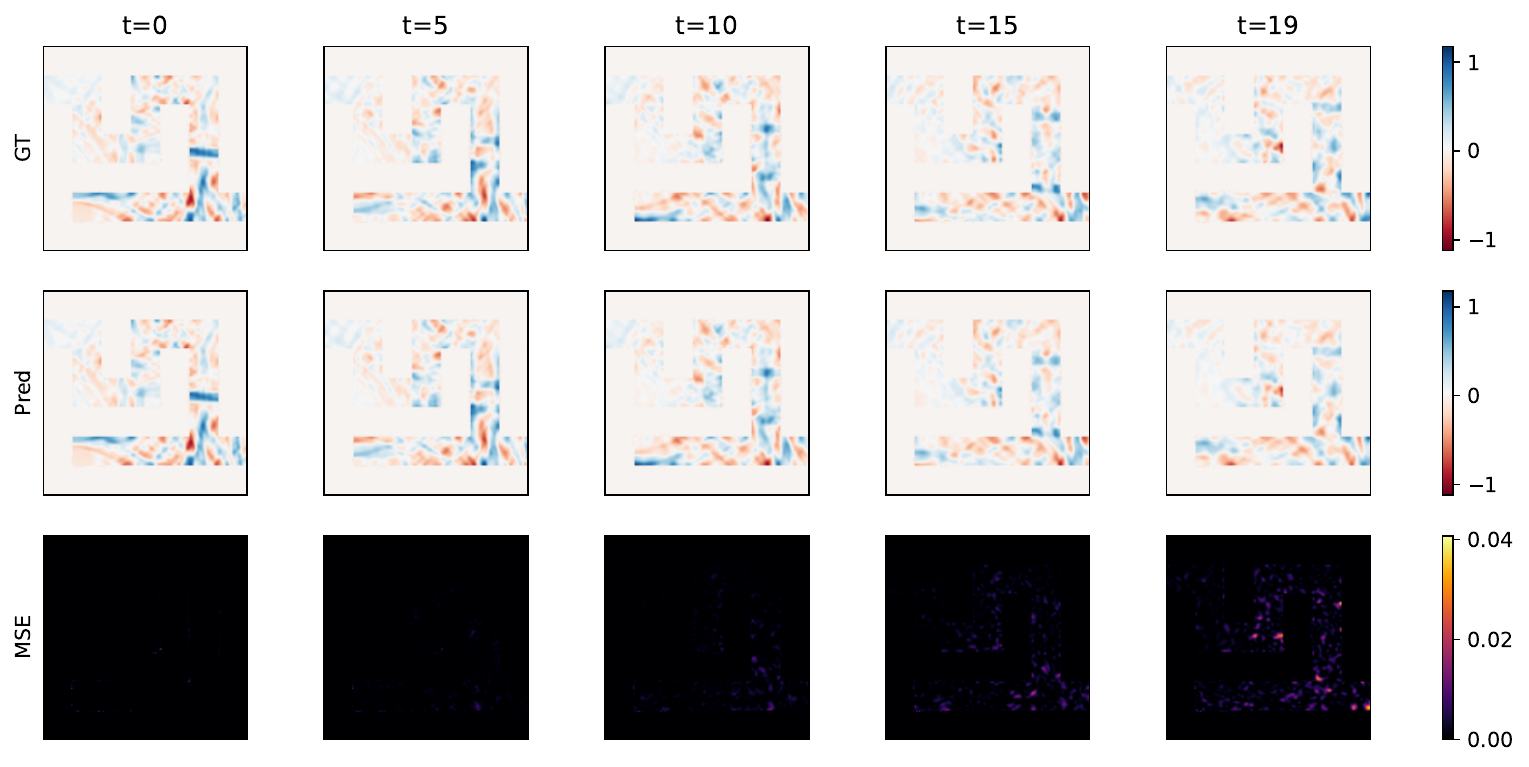}
    \caption{Generation using ECHO's framework on the Acoustic Scattering Maze dataset. (Top) ground truth trajectory, (Middle) ECHO's prediction, (Bottom) Error.}
    \label{fig:fluxasm}
\end{figure}

\clearpage
\subsection{Eagle}
\label{app:ssec:visu_eagle}

\begin{figure}[htbp]
    \centering
    \includegraphics[width=\linewidth]{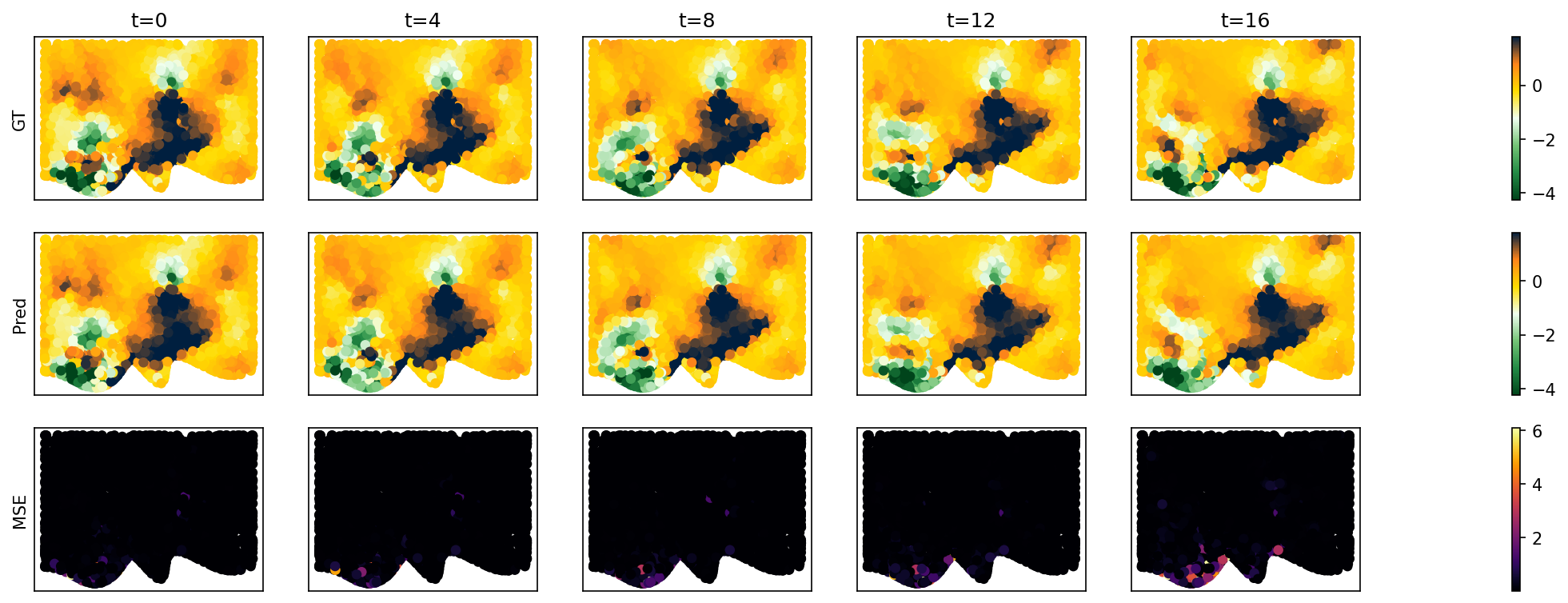}
    \caption{Generation using ECHO's framework on the Eagle dataset. (Top) ground truth trajectory, (Middle) ECHO's prediction, (Bottom) Error.}
    \label{fig:fluxeagle20}
\end{figure}

\begin{figure}[htbp]
    \centering
    \includegraphics[width=\linewidth]{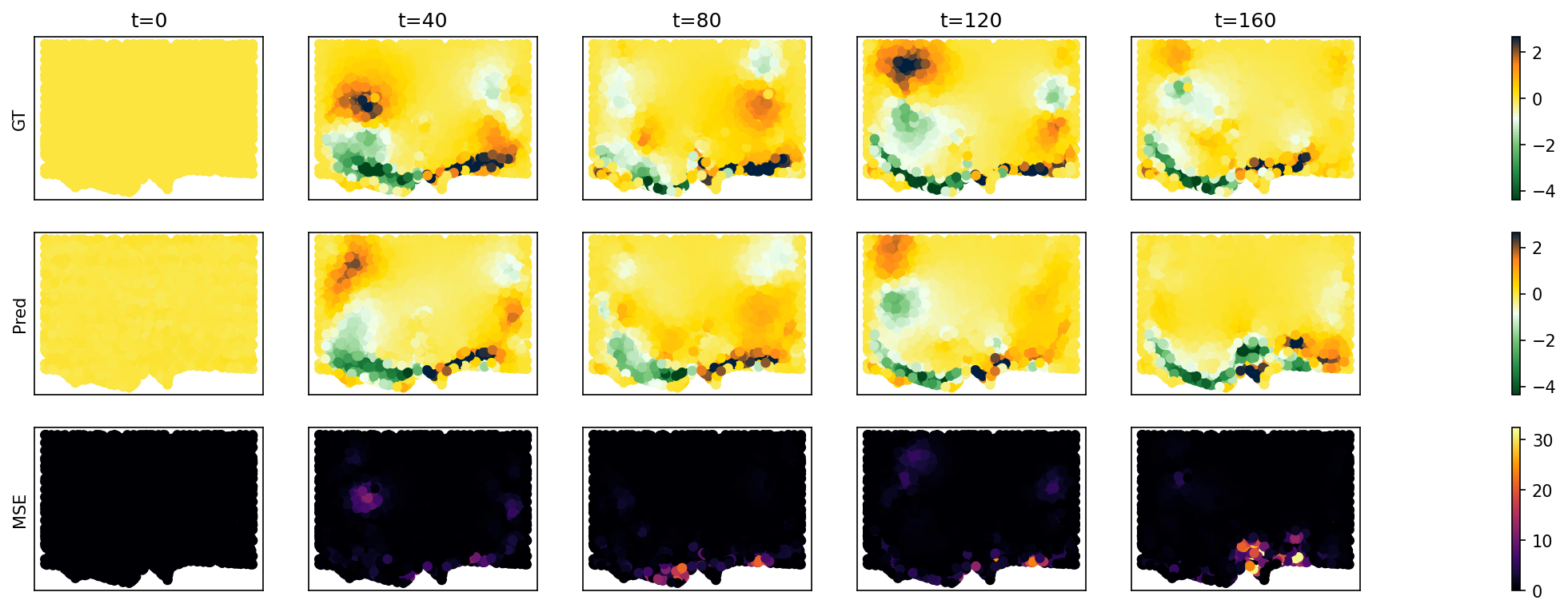}
    \caption{Generation using ECHO's framework on the Eagle dataset - long trajectory generation. (Top) ground truth trajectory, (Middle) ECHO's prediction, (Bottom) Error.}
    \label{fig:fluxeagle200}
\end{figure}